%% file: main_icml23.tex
\documentclass{article}
\input{math_commands_icml.tex}

\usepackage{microtype}
\usepackage{graphicx}
\usepackage{subfigure}
\usepackage{booktabs} 
\usepackage[shortlabels]{enumitem}
\usepackage{xcolor,colortbl}
\usepackage{titlesec}

\usepackage{listings}
\definecolor{dkgreen}{rgb}{0,0.6,0}
\definecolor{gray}{rgb}{0.5,0.5,0.5}
\definecolor{mauve}{rgb}{0.58,0,0.82}

\usepackage[accepted]{icml2023}

\usepackage{amsmath}
\usepackage{amssymb}
\usepackage{mathtools}
\usepackage{amsthm}

\usepackage{hyperref}
\usepackage[capitalize,noabbrev]{cleveref}

\theoremstyle{plain}
\newtheorem{theorem}{Theorem}[section]
\newtheorem{proposition}[theorem]{Proposition}
\newtheorem{lemma}[theorem]{Lemma}

\theoremstyle{definition}

\newtheorem{assumption}[theorem]{Assumption}
\theoremstyle{remark}
\newtheorem{remark}[theorem]{Remark}

\usepackage[textsize=tiny]{todonotes}

\icmltitlerunning{\hfill FP-Diffusion: Improving Score-based Diffusion Models by Enforcing the Underlying Score Fokker-Planck Equation \hfill}

\begin{document}

\twocolumn[
\icmltitle{FP-Diffusion: Improving Score-based Diffusion Models by Enforcing\\ the Underlying Score Fokker-Planck Equation
}



\icmlsetsymbol{equal}{*}

\begin{icmlauthorlist}
\icmlauthor{Chieh-Hsin Lai}{sonyai}
\icmlauthor{Yuhta Takida}{sonyai}
\icmlauthor{Naoki Murata}{sonyai}
\icmlauthor{Toshimitsu Uesaka}{sonyai}
\icmlauthor{Yuki Mitsufuji}{sonyai,sony}
\icmlauthor{Stefano Ermon}{sch}
\end{icmlauthorlist}

\icmlaffiliation{sonyai}{Sony AI, Tokyo, Japan}
\icmlaffiliation{sony}{Sony Group Corporation, Tokyo, Japan}
\icmlaffiliation{sch}{Department of Computer Science, Stanford University, Stanford, CA, USA}

\icmlcorrespondingauthor{Chieh-Hsin Lai}{Chieh-hsin.lai@sony.com}

\icmlkeywords{generative model, diffusion models, score-based diffusion models, fokker-planck equation, consistency}
\vskip 0.3in

]



\printAffiliationsAndNotice{}  

\begin{abstract}

Score-based generative models (SGMs) learn a family of noise-conditional score functions corresponding to the data density perturbed with increasingly large amounts of noise. These perturbed data densities are linked together by the \emph{Fokker-Planck equation} (FPE), a partial differential equation (PDE) governing the spatial-temporal evolution of a density undergoing a diffusion process. In this work, we derive a corresponding equation called the \emph{score FPE} that characterizes the noise-conditional scores of the perturbed data densities (i.e., their gradients). Surprisingly, despite the impressive empirical performance, we observe that scores learned through denoising score matching (DSM) fail to fulfill the underlying score FPE, which is an inherent self-consistency property of the ground truth score.
We prove that satisfying the score FPE is desirable as it improves the likelihood and the degree of conservativity. Hence, we propose to regularize the DSM objective to enforce satisfaction of the score FPE, and we show the effectiveness of this approach across various datasets.

\end{abstract}

\section{Introduction}\label{sec:intro}



Score-based generative models (SGMs), also referred to as diffusion models~\citep{sohl2015deep,song2019generative,ho2020denoising,song2020score,song2020sliced}, 
have led to major advances in the 
generation of synthetic images~\citep{dhariwal2021diffusion,saharia2022photorealistic,rombach2022high,kim2022refining} and audio~\citep{kong2020diffwave}. In addition,
SGMs have been applied to various downstream tasks such as media content editing~\citep{meng2021sdedit,cheuk2022diffroll}, or restoration~\citep{kawar2022denoising,saito2022unsupervised,murata2023gibbsddrm}. 
An SGM involves a stochastic forward and backward process. In the forward process, also known as the diffusion process,
noise with gradually increasing variances is added to each data point until the original structure is lost, transforming data into pure noise. 
The backward process attempts to reverse the diffusion process by using  
a neural network (called a noise-conditional score model) that is trained to 
gradually denoise the data, effectively transforming pure noise into clean data samples. 
The neural network is trained with 
a denoising score matching objective~\citep{hyvarinen2005estimation,vincent2011connection} 
to estimate the score (i.e., the gradient of the log-likelihood function) of the data density perturbed with various amounts of noise (as in forward process). 

The training can be interpreted as a joint estimation of  the scores of the original data density and all its perturbations. Crucially, all these densities are closely related to each other, as they correspond to the same data density perturbed with various amounts of noise.
With sufficiently small time steps, the forward 
process is a diffusion~\citep{song2020score} and 
the spatial-temporal evolution of the data
density is thus governed by the classic Fokker-Planck partial differential equation (PDE)~\citep{oksendal2003stochastic}. In principle, this implies that with knowledge of the density for a \emph{single} noise level, we could  recover all the densities by solving the Fokker-Planck equation (FPE) without any additional learning.

\textbf{Our contributions }
Building on the above notions, we derive an associated system of PDEs that characterizes the evolution of the 
\emph{scores} (i.e., gradients) of the perturbed data densities; we term it as \emph{score Fokker-Planck equation} (\emph{score FPE}). In theory, the ground truth scores of the perturbed data densities must satisfy the score FPE (\emph{self-consistency property}).
Hence, we mathematically study  the implications of satisfying the score FPE. 
We prove the following effects of reducing the score FPE error: (a) improvement in the log-likelihood of the probability flow ordinary differential equation (ODE)  diffusion mode~\citep{song2020score}, (Theorems~\ref{th:min_ode} and \ref{th:min_ode_add}); and (b) improvement in the degree of conservativity of the models (Proposition~\ref{th:conservativity}). 
In addition, we prove that (c) score FPE error reduction can be achieved by enforcing higher-order score matching~\citep{meng2021estimating,lu2022maximum} (Proposition~\ref{th:higher}). 
In practice, we observe  that many existing, pre-trained score models do not numerically satisfy the score FPE. Therefore, we propose a new loss function for training diffusion models by combining the traditional score matching objective with a regularization term derived from the underlying score FPE to enforce the consistency of models. Our proposed new method is called \emph{FP-Diffusion}.
We show that FP-Diffusion enables more accurate density estimation on synthetic data and improves the likelihood on the MNIST, Fashion MNIST, CIFAR-10 and ImageNet32 (ImageNet downsampled to $32\times32$)~\citep{chrabaszcz2017downsampled} datasets. 

\section{Background}\label{sec:background}
\citet{song2020score} 
unified denoising score matching~\citep{song2019generative} and diffusion probabilistic models~\citep{sohl2015deep,ho2020denoising} via a stochastic process $\bm{x}(t)$ with continuous time $t \in [0, T]$. The process is driven by the following forward  SDE
\begin{equation}\label{eq:sde_forward}
    d\bm{x}(t) = \bm{f}(\bm{x}(t), t) dt + g(t) d\bm{w}_t,
\end{equation}
where $\bm{f}(\cdot, t)\colon\mathbb{R}^D\rightarrow\mathbb{R}^D$, $g(\cdot)\colon\mathbb{R}\rightarrow\mathbb{R}$ are pre-assigned\footnote{With specific choices of $\bm{f}$ and $g$, there are two common instantiations of the stochastic differential equation (SDE): VE and VP. See Appendix~\ref{sec:instances} for details.} and $\bm{w}_t$ is a standard Wiener process. Under moderate conditions~\citep{anderson1982reverse}, a reverse time SDE from $T$ to $0$ can be obtained as
\begin{equation}\label{eq:sde_backward}
    d\bm{x}(t) = [\bm{f}(\bm{x}(t), t) - g^2(t) \grad{\bm{x}}\log q_t(\bm{x}(t)) ] dt + g(t) d\bar{\bm{w}}_t,
\end{equation}
where $\bar{\bm{w}}_t$ is a standard Wiener process in reverse time, and $q_t(\bm{x})$ denotes the ground truth marginal density of $\bm{x}(t)$ following Eq.~\eqref{eq:sde_forward}. 
We can train a time-conditional neural network    $\bm{s}_{\bm{\theta}}=\bm{s}_{\bm{\theta}}(\bm{x}, t)$ to approximate $\grad{\bm{x}}\log q_t(\bm{x})$ by minimizing 
a score matching objective~\citep{hyvarinen2005estimation} $\mathcal{J}_{\text{SM}}(\bm{\theta}; \lambda(\cdot)):=$
\begin{equation*}
\frac{1}{2}\int_{0}^{T} \lambda(t) \mathbb{E}_{\bm{x} \sim q_{t}(\bm{x})}\Big[\norm{\bm{s}_{\bm{\theta}}(\bm{x}, t) - \grad{\bm{x}} \log q_t(\bm{x})}_2^2 \Big]dt.
\end{equation*}
As $ q_t(\bm{x})$ is generally inaccessible, the denoising score matching (DSM) loss~\citep{vincent2011connection,song2020score} $\mathcal{J}_{\text{DSM}}(\bm{\theta}; \lambda(\cdot))$
is exploited in practice instead
\begin{equation}\label{eq:dsm}
\begin{aligned}
     \mathcal{J}_{\text{DSM}}&(\bm{\theta}; \lambda(\cdot)):=\frac{1}{2}\int_{0}^{T} \lambda(t) \mathbb{E}_{\bm{x}(0)} \mathbb{E}_{ q_{0t}(\bm{x}(t)|\bm{x}(0))} \\&\Big[\norm{\bm{s}_{\bm{\theta}}(\bm{x}(t), t) - \nabla_{\bm{x}} \log q_{0t}(\bm{x}(t)|\bm{x}(0))}_2^2 \Big]dt,
\end{aligned}
\end{equation}
where $q_{0t}(\bm{x}(t)|\bm{x}(0))$ is the forward transition probability from $\bm{x}(0)$ to $\bm{x}(t)$. After $\bm{s}_{\bm{\theta}}(\bm{x}, t) \approx \grad{\bm{x}}\log q_t(\bm{x})$ is learned, we replace $\grad{\bm{x}} \log q_t(\bm{x})$ in Eq.~\eqref{eq:sde_backward} with $\bm{s}_{\bm{\theta}}$ and obtain a parametrized reverse-time SDE for a stochastic process $\hat{\bm{x}}_{\bm{\theta}}(t)$
\begin{equation}\label{eq:sde_backward_sub}
    d\hat{\bm{x}}_{\bm{\theta}}(t) = [\bm{f}(\hat{\bm{x}}_{\bm{\theta}}(t), t) - g^2(t) \bm{s}_{\bm{\theta}}(\hat{\bm{x}}_{\bm{\theta}}(t), t)  ] dt + g(t) \bar{\bm{w}}_t,
\end{equation}
Let $p_{t,\bm{\theta}}^{\text{SDE}}$ denote the marginal distribution of $\hat{\bm{x}}_{\bm{\theta}}(t)$ with an initial distribution defined as the prior $\pi$, where we suppress the dependence on $\pi$ for compactness.
We can design $\bm{f}$ and $g$ in Eq.~\eqref{eq:sde_backward}, such that $q_T(\bm{x})$ approximates a simple prior $\pi$; samples $\hat{\bm{x}}_{\bm{\theta}}(0)\sim p_{0,\bm{\theta}}^{\text{SDE}}$ can be generated by numerically solving Eq.~\eqref{eq:sde_backward_sub} backward with an initial sample from the prior  $\hat{\bm{x}}_{\bm{\theta}}(T) \sim \pi$. Intuitively, $\hat{\bm{x}}_{\bm{\theta}}(0)$ should be close to a sample from the data distribution.  

\citet{song2020score} also introduced a deterministic process (with a zero diffusion term) that describes the evolution of samples whose trajectories share the
same marginal probability densities as the forward SDE (Eq.~\eqref{eq:sde_backward_sub}). Specifically, the process evolves through time according to the following 
probability flow ODE
\begin{equation}\label{eq:prob_ode_gt}
    \frac{d\bm{x}}{dt}(t) = \bm{f}(\bm{x}(t), t) -\frac{1}{2}g^2(t) \grad{\bm{x}} \log q_t (\bm{x}(t)).
\end{equation}
As in the SDE case, the ground truth score in Eq.~\eqref{eq:prob_ode_gt} is approximated with the learned score model $\bm{s}_{\bm{\theta}}(\bm{x}, t)\approx\grad{\bm{x}} \log q_t (\bm{x})$. This yields to the following parameterized probability flow ODE
\begin{equation}\label{eq:prob_ode_sub}
    \frac{d\tilde{\bm{x}}_{\bm{\theta}}}{dt}(t) = \bm{f}(\tilde{\bm{x}}_{\bm{\theta}}(t), t) -\frac{1}{2}g^2(t) \bm{s}_{\bm{\theta}}(\tilde{\bm{x}}_{\bm{\theta}}(t), t)
\end{equation}
We denote the marginal density of $\tilde{\bm{x}}_{\bm{\theta}}$ as $p_{t,\bm{\theta}}^{\text{ODE}}$ 
with an initial condition sampled from the prior $\pi$, For compactness, we omit the dependence on $\pi$ in the notation.
By solving Eq.~\eqref{eq:prob_ode_sub} numerically using an initial value $\tilde{\bm{x}}_{\bm{\theta}}(T)\sim\pi$, we can  generate a sample $\tilde{\bm{x}}_{\bm{\theta}}(0)\sim p_{0,\bm{\theta}}^{\text{ODE}}$ to approximate sampling from the data distribution.
Indeed, the deterministic dynamics in Eq.~\eqref{eq:prob_ode_sub} make it possible to compute exact likelihoods for this generative model. 
Let $\tilde{\bm{x}}_{\bm{\theta}}(t)\in\mathbb{R}^D$ evolve in reverse time via Eq.~\eqref{eq:prob_ode_sub}, starting with $\tilde{\bm{x}}_{\bm{\theta}}(T)\sim \pi$.
The ``instantaneous change of variables''~\citep{chen2018neural} characterizes the temporal changes in $\log p^{\textup{ODE}}_{t,\bm{\theta}}$ along the trajectory  $\big\{\tilde{\bm{x}}_{\bm{\theta}}(t): t\in[0,T] \big\}$ via the following ODE: 
\begin{align*}
    &\frac{d\log p^{\textup{ODE}}_{t,\bm{\theta}}(\tilde{\bm{x}}_{\bm{\theta}}(t))}{dt} 
    \\=& \frac{1}{2}g^2(t)\div{\bm{x}}\big(\bm{s}_{\bm{\theta}}(\tilde{\bm{x}}_{\bm{\theta}}(t), t)\big) -\div{\bm{x}}\big(\bm{f}(\tilde{\bm{x}}_{\bm{\theta}}(t), t)\big).
\end{align*}
Hence, the log-likelihood can be exactly calculated by numerically solving the concatenated ODEs backward from $T$ to $0$, after initialization with $\tilde{\bm{x}}_{\bm{\theta}}(0)\sim q_0(\bm{x})$
\begin{align*}
   &\frac{d}{dt} 
   \begin{bmatrix}
         \tilde{\bm{x}}_{\bm{\theta}}(t)   \\
         \log p^{\textup{ODE}}_{t,\bm{\theta}}(\tilde{\bm{x}}_{\bm{\theta}}(t))   \\
    \end{bmatrix}
    \\=&
   \begin{bmatrix}
        \bm{f}(\tilde{\bm{x}}_{\bm{\theta}}(t), t) -\frac{1}{2}g^2(t) \bm{s}_{\bm{\theta}}(\tilde{\bm{x}}_{\bm{\theta}}(t), t)    \\
        \frac{1}{2}g^2(t)\div{\bm{x}}\big(\bm{s}_{\bm{\theta}}(\tilde{\bm{x}}_{\bm{\theta}}(t), t)\big) -\div{\bm{x}}\big(\bm{f}(\tilde{\bm{x}}_{\bm{\theta}}(t), t)\big)    \\
    \end{bmatrix}.  
\end{align*}

\section{Score Fokker-Planck equation for diffusion}
\label{sec:FP_derivation}

It is well known that the evolution of the ground truth density $q_t(\bm{x})$ associated with Eq.~\eqref{eq:sde_forward} is governed by the Fokker-Planck equation (FPE)~\citep{oksendal2003stochastic} 
\begin{equation*}
    \partial_t q_t (\bm{x}) = -\sum_{j=1}^{D} \partial_{x_j}\big(\tilde{\bm{F}}_j(\bm{x}, t) q_t (\bm{x})\big),
\end{equation*}
where $\tilde{\bm{F}}(\bm{x}, t):= \bm{f}(\bm{x}, t) -\frac{1}{2}g^2(t)\nabla_{\bm{x}} \log q_t(\bm{x})$. As there is a one-to-one mapping (up to a constant) between densities and their scores, we derive (in Appendix~\ref{sec:proofs}) an equivalent system of PDEs for the ground truth \emph{scores}  $\grad{\bm{x}} \log q_t(\bm{x})$. 
We designate it as the \emph{score Fokker-Planck equation} or simply the \emph{score FPE}.

\begin{proposition}[Score FPE]\label{th:fp}
 Assume the \emph{ground truth} density $q_t(\bm{x})$ is sufficiently smooth on $\mathbb{R}^D\times[0,T]$ with its score denoted as $\bm{s}(\bm{x}, t) :=\nabla_{\bm{x}} \log q_t(\bm{x})$. Then for all $(\bm{x}, t) \in \mathbb{R}^D\times[0,T]$, its log-density satisfies  the PDE
\begin{equation}\label{eq:scafp_gt}
\begin{aligned}
    \partial_t \log q_t(\bm{x}) = &\frac{1}{2}g^2(t) \div{\bm{x}}(\bm{s}(\bm{x}, t)) + \frac{1}{2}g^2(t)\norm{\bm{s}(\bm{x}, t)}^2_2 \\
    &  -\inner{\bm{f}(\bm{x}, t)}{\bm{s}(\bm{x}, t)} - \div{\bm{x}}(\bm{f}(\bm{x}, t))
\end{aligned}
\end{equation}
and its score $\bm{s}$  satisfies the following system of PDEs 
\begin{equation}\label{eq:fp_gt}
\begin{aligned}
    \partial_t \bm{s}(\bm{x}, t) &= \grad{\bm{x}}\Big[\frac{1}{2}g^2(t) \div{\bm{x}}(\bm{s}(\bm{x}, t)) + \frac{1}{2}g^2(t)\norm{\bm{s}(\bm{x}, t)}^2_2\\
    & -\inner{\bm{f}(\bm{x}, t)}{\bm{s}(\bm{x}, t)} - \div{\bm{x}}(\bm{f}(\bm{x}, t)) \Big]. 
\end{aligned}
\end{equation}
\end{proposition}


For notational simplicity, let $\mathcal{L}[\cdot] :=\frac{1}{2}g^2 \div{\bm{x}}(\cdot) + \frac{1}{2}g^2\norm{\cdot}^2_2-\inner{\bm{f}}{\cdot} - \div{\bm{x}}(\bm{f})$ be the operator mapping vector fields to real-valued functions. Thus, Eq.~\eqref{eq:scafp_gt} and Eq.~\eqref{eq:fp_gt} can be expressed as $\partial_t \log q_t(\bm{x})=\mathcal{L}[\bm{s}](\bm{x}, t)$ and $\partial_t \bm{s}(\bm{x}, t)=\grad{\bm{x}}\mathcal{L}[\bm{s}](\bm{x}, t)$, respectively.

Proposition~\ref{th:fp} shows that the time-conditional  scores 
$\bm{s}_{\bm{\theta}}(\bm{x}, t)$ learned by score-based models (via Eq.~\eqref{eq:dsm}) are highly redundant. In principle, given a ground truth score at an initial time $t_0$, we can theoretically recover scores for all times $t\geq t_0$ by solving the score FPE. We explain it intuitively by considering the special case when $\bm{f} \equiv \bm{0}$ and $g \equiv 1$, i.e., when, $\bm{x}(t)$ is obtained by adding Gaussian noise.
It is well-known that the densities $q_t$ and $q_{t_0}$ are related in a convolutional way as $q_t = q_{t_0} * \mathcal{N}(0,t)$, and that $q_t$ can be analytically obtained from $q_{t_0}$~\citep{masry1992gaussian} (e.g., by applying a Fourier transform and dividing). Hence, all scores can in principle be obtained analytically from the score at a single time-step, without any further learning. 

We provide empirical evidence to substantiate Proposition~\ref{th:fp} from two distinct perspectives, as presented in Section~\ref{subsec:solve_pde} and Appendix~\ref{subsec:supp-sec-3}, respectively.

\subsection{Pre-trained scores fail to satisfy score FPEs}\label{subsec:fail_sfpe}
Theoretically, with sufficient data and model capacity, score matching ensures that the optimal solution to Eq.~\eqref{eq:dsm} should satisfy Eq.~\eqref{eq:fp_gt} as it should approximate the ground truth score well. However, in our experiments, we observe that pre-trained scores $\bm{s}_{\bm{\theta}}$ learned via Eq.~\eqref{eq:dsm} do not fulfill the score FPE. Therefore, we introduce an error term $\bm{\epsilon}[{\bm{s_{\theta}}}]:= \bm{\epsilon}[{\bm{s_{\theta}}}](\bm{x}, t)$ to quantify how much $\bm{s}_{\bm{\theta}}$ deviates from the score FPE
\begin{equation}\label{eq:fp_error}
  \begin{aligned}
    \bm{\epsilon[{\bm{s_{\theta}}}]}(\bm{x}, t):= \partial_t \bm{s_{\theta}}(\bm{x}, t) - \grad{\bm{x}}\mathcal{L}[\bm{s_{\theta}}](\bm{x}, t).
\end{aligned}  
\end{equation}

Set $T=1$, we define the average residual of the score FPE, computed over $\bm{x}$, as a function of $t\in[0, 1]$
\begin{align*}
    r_{\text{FP, trans.}}[\bm{s}_{\bm{\theta}}](t):=\frac{1}{D}\mathbb{E}_{\bm{x}(0)}\mathbb{E}_{\bm{x}(t)|\bm{x}(0)}\Big[\norm{\bm{\epsilon}[{\bm{s_{\theta}}}](\bm{x}, t)}_2\Big]. 
\end{align*}
We further consider the following averaged residual for DSM 
\begin{align*}
   r_{\text{DSM-like}}[\bm{s}_{\bm{\theta}}](t):= \frac{1}{D}&\mathbb{E}_{\bm{x}(0)} \mathbb{E}_{ \bm{x}(t)|\bm{x}(0)} \Big[\lVert\bm{s}_{\bm{\theta}}(\bm{x}(t), t) 
    \\&- \nabla_{\bm{x}(t)} \log q_{0t}(\bm{x}(t)|\bm{x}(0))\rVert_2 \Big].
\end{align*}
Compared to the integrand in the standard DSM loss in Eq.~\eqref{eq:dsm}, $r_{\text{DSM-like}}[\bm{s}_{\bm{\theta}}]$ uses the $\ell_2$-norm (instead of the MSE) and drops the time-weighting function $\lambda(t)$ to be consistent with the averaged residuals of the score FPE.
\begin{figure*}[th]
     \centering
     \subfigure[VE SDE; MNIST]{\includegraphics[width=0.24\textwidth]{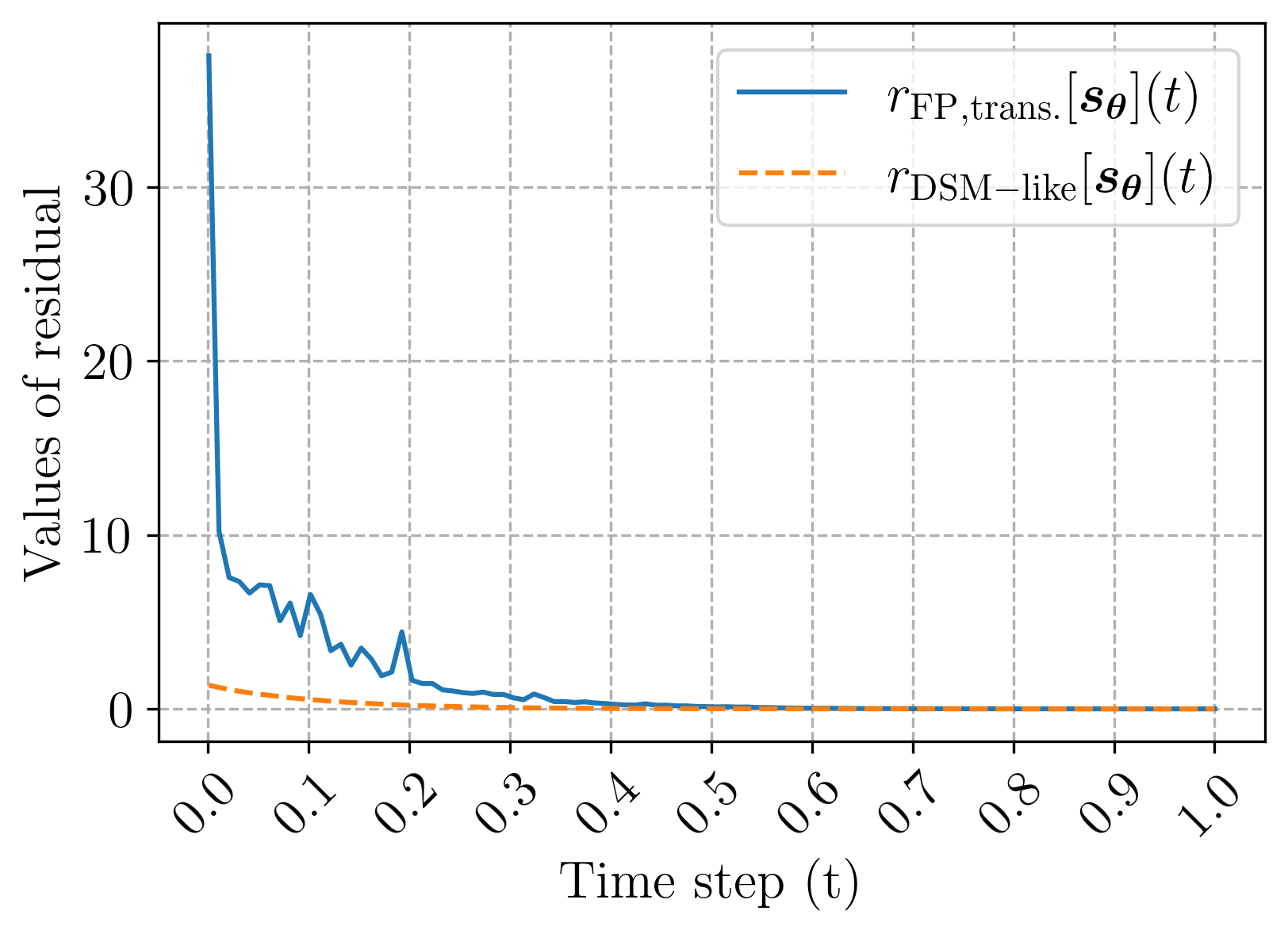}}
    \subfigure[VP SDE; MNIST]{\includegraphics[width=0.24\textwidth]{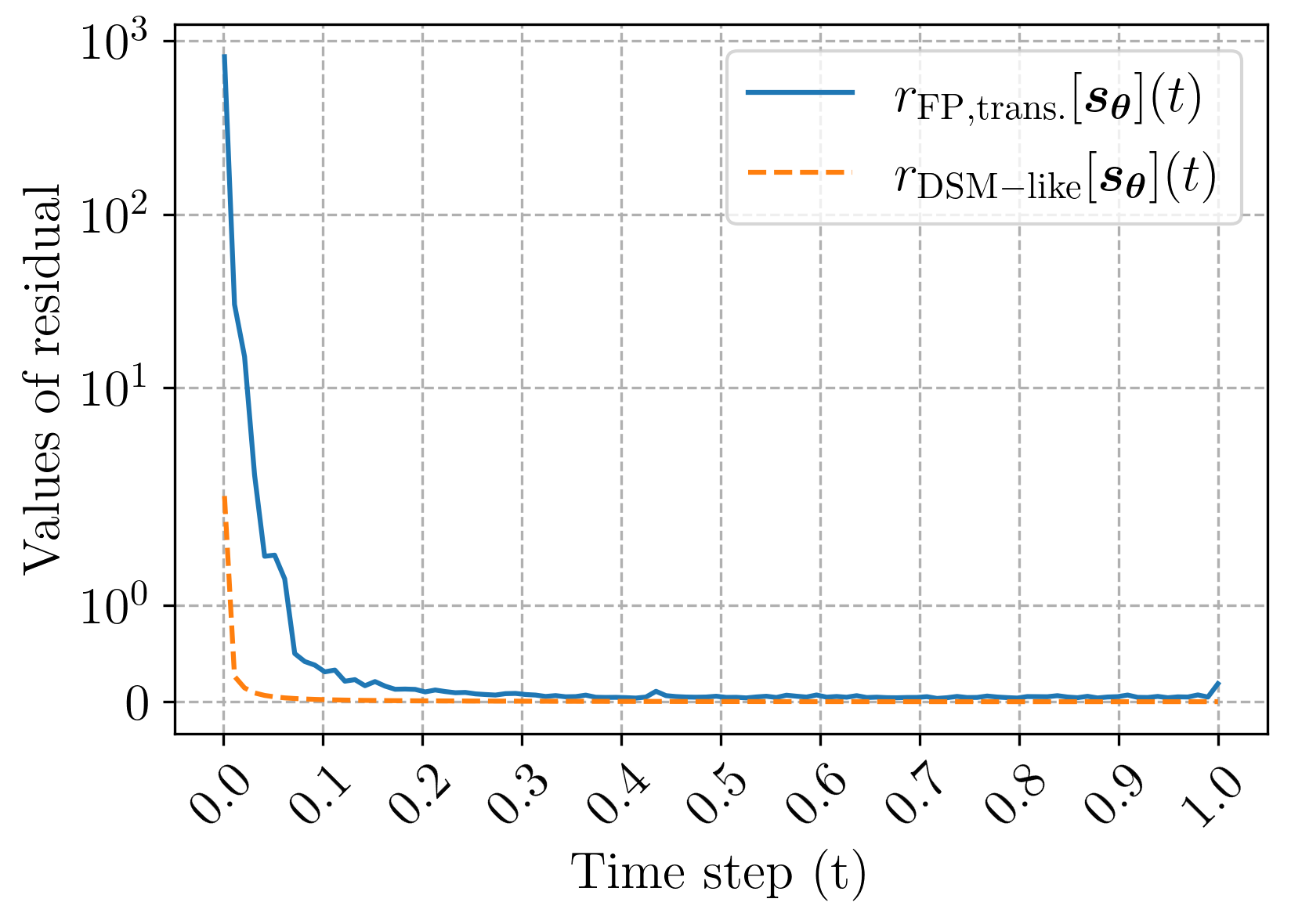}}
     \subfigure[VE SDE; CIFAR-10]{\includegraphics[width=0.24\textwidth]{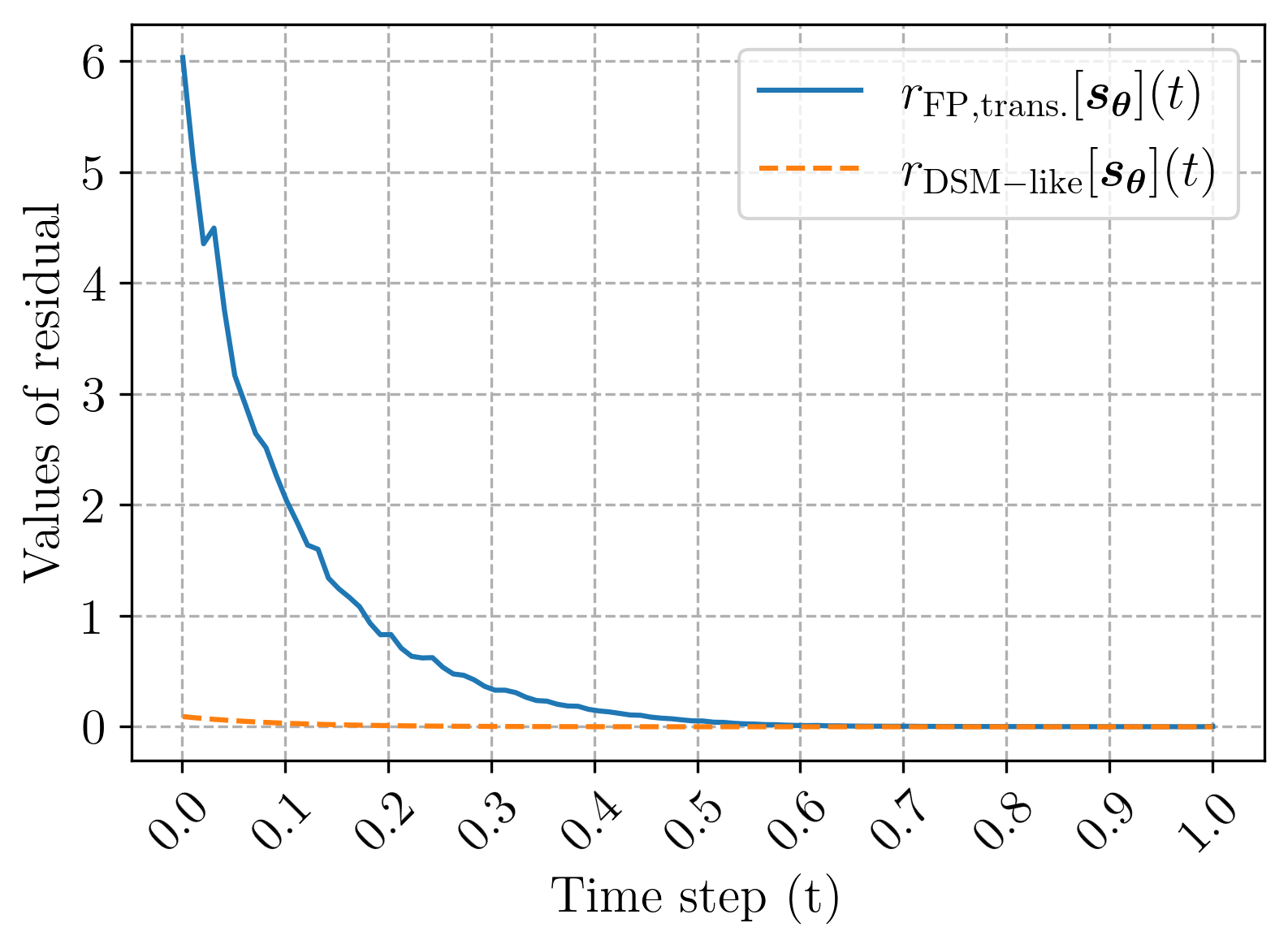}}
    \subfigure[VP SDE; CIFAR-10]{\includegraphics[width=0.24\textwidth]{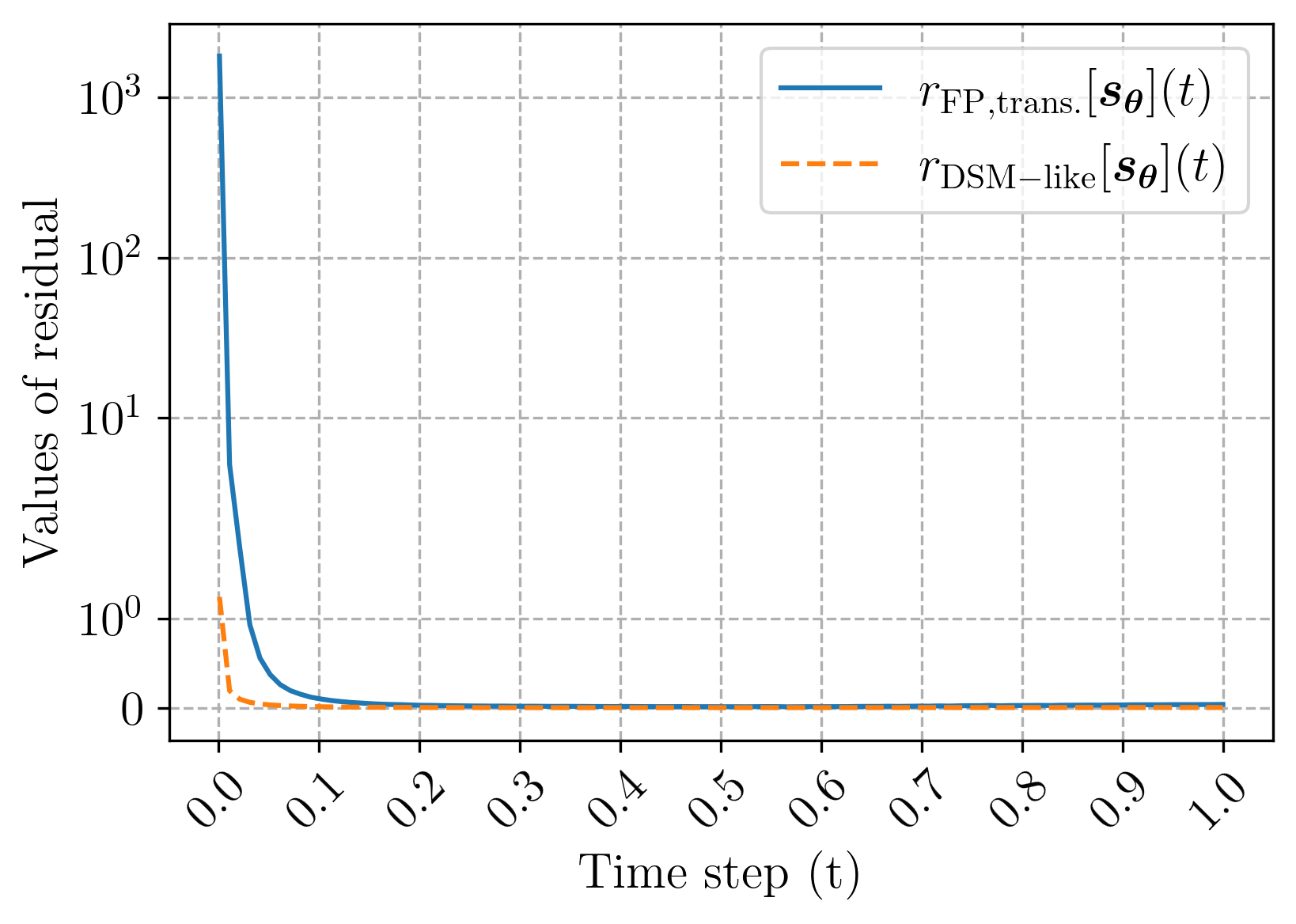}}    
        \caption{Comparison of the numerical scales of $r_{\text{DSM-like}}[\bm{s}_{\bm{\theta}}](t)$ and $r_{\text{FP, trans.}}[\bm{s}_{\bm{\theta}}](t)$ for pre-trained scores  $\bm{s}_{\bm{\theta}}$ on MNIST and CIFAR-10. We treat these errors as functions of time. The pre-trained models do not numerically satisfy the score FPE, in contrast to their DSM-like errors. We attempt to explain this phenomenon in Sections.~\ref{subsec:conservative} and \ref{sec:higher_sm}.}
        \label{fig:large_fp}
\end{figure*}

Figure~\ref{fig:large_fp} plots these residuals for score models that were pre-trained via DSM on the MNIST  and CIFAR-10 datasets.  
Despite achieving a low $r_{\text{DSM-like}}$ across all $t$ (orange curve), the pre-trained score models fail to satisfy the score FPE equation, especially for small $t$ (blue curve). 
This implies that models learned by DSM do not satisfy the score FPE.

\section{Theoretical implications of score FPE}
\label{sec:interpretation}

In this section, we first study three implications of satisfying the score FPE. Specifically, we show in Section~\ref{sec:kl_ode} that simultaneous minimization of  quantities related to the
score FPE  and the conventional score matching objective can reduce the KL divergence between the data density $q_0$ and the density $p^{\textup{ODE}}_{0,\bm{\theta}}$, determined by the parametrized probability flow ODE (Eq.~\eqref{eq:prob_ode_sub}). In Section~\ref{subsec:conservative} we prove that controlling of   $\bm{\epsilon}[{\bm{s_{\theta}}}]$ implicitly enforces the \emph{conservativity} of $\bm{s}_{\bm{\theta}}$. Moreover, in Section~\ref{sec:p_sde} we prove  that if the score FPE is satisfied, then under certain conditions,  $\bm{s}_{\bm{\theta}}$, ground truth score $\bm{s}$, $\grad{\bm{x}}\log p^{\textup{SDE}}_{t,\bm{\theta}}$, and $\grad{\bm{x}}\log p^{\textup{ODE}}_{t,\bm{\theta}}$
must match. Here $p^{\textup{SDE}}_{t,\bm{\theta}}$ and $p^{\textup{ODE}}_{t,\bm{\theta}}$ were defined in Section~\ref{sec:background} as the marginal density of the parametrized diffusion process and the probability flow ODE, respectively.
Finally, in Section~\ref{sec:higher_sm}, we investigate the connection between higher-order score matching~\citep{meng2021estimating,lu2022maximum} and the score FPE. We provide the proofs of all theorems in Appendix~\ref{sec:proofs}.

\subsection{Minimization \texorpdfstring{$\mathcal{D}_{\textmd{KL}}\big(q_0\big\Vert p^{\textmd{ODE}}_{0,\bm{\theta}}\big)$}{TEXT} }\label{sec:kl_ode}

In this section, we show that under certain regularity conditions (see Assumptions~\ref{cond:A} and \ref{cond_add:A}), simultaneous minimization of   $\mathcal{J}_{\text{SM}}(\bm{\theta})$ and certain score FPE related quantities (see Eqs.~\eqref{eq:sup_avg_fpe} and \eqref{eq:sup_avg_scafpe}) can decrease the KL divergence between $q_0$ and  $p^{\textup{ODE}}_{0,\bm{\theta}}$, denoted as $\mathcal{D}_{\textup{KL}}\big(q_0\big\Vert p^{\textup{ODE}}_{0,\bm{\theta}}\big)$. This is equivalent to improving the likelihood of  data under $p^{\textup{ODE}}_{0,\bm{\theta}}$.



First, we review an equation proposed by \citet{lu2022maximum} that quantifies the exact gap between $\mathcal{D}_{\textup{KL}}\big(q_0\big\Vert p^{\textup{ODE}}_{0,\bm{\theta}}\big)$ and the score matching loss $\mathcal{J}_{\text{SM}}(\bm{\theta})$. For compactness, we denote $\bm{s}_{\bm{\theta}}^{\textup{ODE}}(\bm{x}, t) := \grad{\bm{x}}\log p^{\textup{ODE}}_{t,\bm{\theta}}(\bm{x})$.

\begin{lemma}[\citet{lu2022maximum}]\label{th:lu_lemma} Set $\lambda(t)=g^2(t)$.  Let $q_0$ be the data distribution, and $q_t$ be the marginal density of $\bm{x}(t)$ following Eq.~\eqref{eq:sde_forward}. Assume that Assumption~\ref{cond:A}  is satisfied. Then, 
\begin{align*}
    \mathcal{D}_{\textup{KL}}\big(q_0\big\Vert p^{\textup{ODE}}_{0,\bm{\theta}}\big) 
    = \mathcal{D}_{\textup{KL}}\big(q_T\big\Vert p^{\textup{ODE}}_{T,\bm{\theta}}\big) + \mathcal{J}_{\text{SM}}(\bm{\theta}) + \mathcal{J}_{\text{Diff}}(\bm{\theta}), 
\end{align*}
where 
\begin{align*}
  \mathcal{J}_{\text{Diff}}(\bm{\theta}) &= \frac{1}{2}\int_{0}^{T} g^2(t) \mathbb{E}_{q_{t}(\bm{x})}\Big[\big(\bm{s}_{\bm{\theta}}(\bm{x}, t) - \bm{s}(\bm{x}, t) \big)^{\top} 
  \\ &\big( \bm{s}_{\bm{\theta}}^{\textup{ODE}}(\bm{x}, t) -\bm{s}_{\bm{\theta}}(\bm{x}, t)\big) \Big]dt. 
\end{align*}
\end{lemma}


We now introduce the main theoretical results in this section. First, we note that application of the Cauchy-Schwartz inequality to $\mathcal{J}_{\text{Diff}}(\bm{\theta})$ gives  $$\abs{\mathcal{J}_{\text{Diff}}(\bm{\theta})} \leq \sqrt{\mathcal{J}_{\text{SM}}(\bm{\theta})}\cdot \sqrt{\mathcal{J}_{\text{Fisher}}(\bm{\theta})}.$$
Here, $\mathcal{J}_{\text{Fisher}}(\bm{\theta})$ is a Fisher-like divergence in terms of the two scores $\bm{s}_{\bm{\theta}}(\bm{x}, t)$ and $\bm{s}_{\bm{\theta}}^{\textup{ODE}}(\bm{x}, t)$,  defined as $\mathcal{J}_{\text{Fisher}}(\bm{\theta}) :=$
\begin{equation*}
     \frac{1}{2}\int_{0}^{T} g^2(t) \mathbb{E}_{\bm{x} \sim q_{t}(\bm{x})}\norm{\bm{s}_{\bm{\theta}}(\bm{x}, t)-\bm{s}_{\bm{\theta}}^{\textup{ODE}}(\bm{x}, t)}_2^2dt.
\end{equation*}

 Next, in Theorem.~\ref{th:min_ode}, we show that under Assumption~\ref{cond:A},  $\mathcal{J}_{\text{Fisher}}(\bm{\theta})$ can be bounded from above by the averaged residual of the score FPE $M({\bm{\theta}})$: 
\begin{equation}\label{eq:J_fisher_demo}
     \mathcal{J}_{\text{Fisher}}(\bm{\theta}) \lesssim  M({\bm{\theta}}) + \sqrt{M({\bm{\theta}})} + C_1,
\end{equation}
where $C_1>0$ is a constant, $\lesssim$ denotes multiplicative constants independent of $\bm{\theta}$ are concealed,  
and $M({\bm{\theta}}):=$
 \begin{align}\label{eq:sup_avg_fpe}
      \sup_{t\in[0,T]}\mathbb{E}_{\bm{x} \sim q_{t}(\bm{x})}\left[\int_{0}^{T}\norm{\bm{\epsilon}[{\bm{s_{\theta}}}](\bm{x}, \tau)}_2^2 d\tau\right].
 \end{align}
 Furthermore, we can compute 
 \begin{align*}    M({\bm{\theta}})\leq\sup_{\bm{x}}\left[\int_{0}^{T}\norm{\bm{\epsilon}[{\bm{s_{\theta}}}](\bm{x}, \tau)}_2^2 d\tau\right],
 \end{align*}
meaning that this upper bound measures the worst time-averaged score FPE error. In Appendix~\ref{subsec:disc-th-min-ode}, we consider more interpretable quantities than $M({\bm{\theta}})$ by introducing 
the density weighting $p_{\tau}(\bm{x})$ in $\tau$-integrand and derive similar estimations as in Ineq.~\eqref{eq:J_fisher_demo}.
 
Moreover, we prove in Theorem.~\ref{th:min_ode_add} that with a different regularity condition (Assumption~\ref{cond_add:A}), $\mathcal{J}_{\text{Fisher}}(\bm{\theta})$ is upper bounded by $M({\bm{\theta}})$ and a ``time-derivative taming'' term that can be derived from Eq.~\eqref{eq:scafp_gt} which is defined as
\begin{equation}
\begin{aligned}\label{eq:sup_avg_scafpe}
m({\bm{\theta}}):=\sup_{\bm{x}}\int_{0}^{T}&\abs{\mathcal{L}[\bm{s}_{\bm{\theta}}](\bm{x}, \tau)} d\tau.
 \end{aligned}
\end{equation}

More specifically,
\begin{equation}\label{eq:J_fisher_demo_add}
     \mathcal{J}_{\text{Fisher}}(\bm{\theta}) \lesssim  M({\bm{\theta}}) + m({\bm{\theta}}) + C_2,
\end{equation}
where $C_2$ is another constant, distinct from $C_1$.

Hence, Lemma.~\ref{th:lu_lemma} together with Ineq.~\eqref{eq:J_fisher_demo} or \eqref{eq:J_fisher_demo_add} implies that $\mathcal{D}_{\textup{KL}}\big(q_0\big\Vert p^{\textup{ODE}}_{0,\bm{\theta}}\big)$  decreases when ``$ M({\bm{\theta}})$ and $\mathcal{J}_{\text{SM}}(\bm{\theta})$'' or ``$ M({\bm{\theta}})$, $ m({\bm{\theta}})$, and $\mathcal{J}_{\text{SM}}(\bm{\theta})$'' are reduced simultaneously. We now rigorously state these theorems.

\begin{theorem} \label{th:min_ode}
We have
\begin{equation}\label{eq:diff_sm_fisher_add}
    \Big(\mathcal{J}_{\text{Diff}}(\bm{\theta})\Big)^2 \leq \mathcal{J}_{\text{SM}}(\bm{\theta})\cdot \mathcal{J}_{\text{Fisher}}(\bm{\theta}).
\end{equation}
Moreover, if Assumption.~\ref{cond:A} is fulfilled, then there is another finite constant $C_1>0$ independent of $\bm{\theta}$ such that we can further bound Ineq.~\eqref{eq:diff_sm_fisher_add} above by 
\begin{align}\label{eq:diff_sm_M}
 \Big(\mathcal{J}_{\text{Diff}}(\bm{\theta})\Big)^2 \lesssim \mathcal{J}_{\text{SM}}(\bm{\theta}) \cdot \Big(M({\bm{\theta}}) + \sqrt{M({\bm{\theta}})} + C_1 \Big).
\end{align}
Thus, $\mathcal{D}_{\textup{KL}}\big(q_0\big\Vert p^{\textup{ODE}}_{0,\bm{\theta}}\big) \lesssim \mathcal{D}_{\textup{KL}}\big(q_T\big\Vert p^{\textup{ODE}}_{T,\bm{\theta}}\big)$
\begin{align*}
       + \mathcal{J}_{\text{SM}}(\bm{\theta}) +  \mathcal{J}_{\text{SM}}^{1/2}(\bm{\theta}) \Big(M({\bm{\theta}}) + \sqrt{M({\bm{\theta}})} + C_1 \Big)^{1/2}. 
\end{align*}
\end{theorem}

\begin{theorem}\label{th:min_ode_add}
If Assumption.~\ref{cond_add:A} is satisfied, then there is another finite constant $C_2>0$ independent of $\bm{\theta}$ such that
\begin{align}\label{eq:diff_sm_M_add}
 \Big(\mathcal{J}_{\text{Diff}}(\bm{\theta})\Big)^2\lesssim   \mathcal{J}_{\text{SM}}(\bm{\theta}) \cdot \Big(M({\bm{\theta}}) + m({\bm{\theta}}) + C_2 \Big).
\end{align}
\end{theorem}

It is noticed that constants $C_1$ and $C_2$ involve regularity bounds of the ground truth density and Lipschitz constants of networks.
Hence, the upper bounds in Ineq.~\eqref{eq:diff_sm_M} and \eqref{eq:diff_sm_M_add}
are difficult to compare.

As the ground truth score should follow the score FPE, it is intuitive that reduction of the score FPE residual encourages the network-parametrized score to approach the ground truth score (a special case is proved in Proposition~\ref{th:conti_strong}). Theorems~\ref{th:min_ode} and \ref{th:min_ode_add} support that reduction of these quantities related to the score FPE may also reduce the gap (in the KL divergence) of their corresponding densities. 
In Section~\ref{sec:exp}, we empirically support these claims.

\subsection{Conservativity}\label{subsec:conservative} 

The ground truth score $\bm{s}(\bm{x}, t) =\nabla_{\bm{x}} \log q_t(\bm{x})$ is a conservative vector field. That is, it can be expressed as a gradient of some real-valued function. However, scores learned in practice do not satisfy this property~\citep{salimans2021should}. Below, we prove that we can implicitly enforce conservativity by minimizing the time-averaged error $\bm{\epsilon}[{\bm{s_{\theta}}}](\bm{x}, \tau)$ of the score FPE.
  
\begin{proposition} \label{th:conservativity}
If  there is a $t_{\bm{\theta}}\in[0,T]$ 
 so that  $\bm{s}_{\bm{\theta}}(\bm{x}, t_{\bm{\theta}}) =\nabla_{\bm{x}} \log q_{t_{\bm{\theta}}}(\bm{x})$ for all $\bm{x}\in\mathbb{R}^D$, then there exists a real-valued function $\Psi_{\bm{\theta}}\colon \mathbb{R}^D \times [0, T] \rightarrow \mathbb{R}$ (with an explicit expression) that satisfies 
\begin{align}\label{eq:eq_conservativity}
    \bm{s}_{\bm{\theta}}(\bm{x}, t) - \grad{\bm{x}} \Psi_{\bm{\theta}}(\bm{x}, t) = \displaystyle\int_{t_{\bm{\theta}}}^{t}  \bm{\epsilon}[{\bm{s_{\theta}}}](\bm{x}, \tau)d\tau,
\end{align}
for all $(\bm{x}, t) \in \mathbb{R}^D\times[0,T]$. In particular, 
\begin{equation}\label{eq:conservativity}
    \norm{\bm{s}_{\bm{\theta}}(\bm{x}, t) - \grad{\bm{x}} \Psi_{\bm{\theta}}(\bm{x}, t)}_2 \leq \abs{\int_{t}^{t_{\bm{\theta}}} \norm{\bm{\epsilon}[{\bm{s_{\theta}}}](\bm{x}, \tau)}_2 d\tau}.
\end{equation}

\end{proposition}

Eq.~\eqref{eq:eq_conservativity} indicates that the error of the score FPE quantifies the degree of conservativity of $\bm{s}_{\bm{\theta}}$. We further explain this idea via Ineq.~\eqref{eq:conservativity}, from which we easily obtain
$\norm{\bm{s}_{\bm{\theta}}(\bm{x}, t) - \grad{\bm{x}} \Psi_{\bm{\theta}}(\bm{x}, t)}_2 \leq \abs{\int_{t}^{t_{\bm{\theta}}} \norm{\bm{\epsilon}[{\bm{s_{\theta}}}](\bm{x}, \tau)}_2 d\tau}\leq\int_{0}^{T} \norm{\bm{\epsilon}[{\bm{s_{\theta}}}](\bm{x}, \tau)}_2 d\tau$, for any $\bm{x}$ and $t$. Thus, if the $\bm{\theta}$-parametrized score approximately satisfies the score FPE, giving a small score FPE error $\int_{0}^{T} \norm{\bm{\epsilon}[{\bm{s_{\theta}}}](\bm{x}, \tau)}_2 d\tau$, then the estimated score should nearly be conservative, i.e., close to the gradient of a scalar function
$\Psi_{\bm{\theta}}(\bm{x}, t)$. We empirically support this fact in Section~\ref{subsec:conservative_exp}.

Proposition~\ref{th:conservativity} necessitates a precise alignment of scores at a given timestep. However, we propose a modification that allows for a small discrepancy by incorporating an error term into the score matching process. As a result, we present an expanded proposition, namely Proposition~\ref{th:revised_conservativity}, which is detailed in Appendix~\ref{subsec:proof-conservativity}.

\subsection{Equivalence of scores }
\label{sec:p_sde}




We now investigate another implication of satisfying the score FPE which connects the score $\bm{s}_{\bm{\theta}}$ with the ground truth $\bm{s}$, $\bm{s}_{\bm{\theta}}^{\textup{SDE}}$, and $\bm{s}_{\bm{\theta}}^{\textup{ODE}}$.
The following proposition provides conditions under which all of these  scores are identical if we train to reach a zero residual for the  score FPE for all $(\bm{x}, t)$.

\begin{proposition}\label{th:conti_strong}
(1) Suppose in some suitable function space, $\bm{0}$ is the unique strong solution to the PDEs 
 $\partial_{t}\bm{v} - \nabla_{\bm{x}} \big[ \frac{1}{2} g^2 \div{\bm{x}}(\bm{v}) + \frac{1}{2} g^2 \big(\norm{\bm{v}}_2^2 +2 \inner{\bm{v}}{\bm{s}} \big)- \inner{\bm{f}}{\bm{v}} \big] =0$ with a zero initial condition $\bm{v}(\bm{x}, 0) \equiv 0$ and a zero boundary condition. If there is some $\bm{\theta}_0$ so that for all $(\bm{x}, t)$ $\bm{\epsilon}[{\bm{s}_{\bm{\theta}_0}}](\bm{x}, t)=0$  and that $\bm{s}_{\bm{\theta}_0}(\bm{x}, 0)=\bm{s}(\bm{x}, 0)$, then $\bm{s}_{\bm{\theta}_0}(\bm{x}, t)=\bm{s}(\bm{x}, t)$, for all $(\bm{x}, t)$ . 
 
 (2) Moreover, suppose the PDEs $\partial_{t}\bm{v} + \nabla_{\bm{x}} \big[ \frac{1}{2} g^2 \div{\bm{x}}(\bm{v}) + \frac{1}{2} g^2 \norm{\bm{v}}_2^2 + \inner{\bm{f}}{\bm{v}} \big] =0$  with zero initial and boundary condition have $\bm{0}$ as the unique strong solution. Then $\bm{\epsilon}[{\bm{s}_{\bm{\theta}_0}}] \equiv 0$ and  $\bm{s}_{\bm{\theta}_0}(\bm{x}, 0)\equiv\bm{s}_{\bm{\theta}_0}^{\textup{SDE}}(\bm{x}, 0)$ implies $\bm{s}_{\bm{\theta}_0} \equiv \bm{s}_{\bm{\theta}_0}^{\textup{SDE}}$.

 (3) Lastly, if there is some $\bm{\theta}_0$ such that  $\partial_{t}\bm{v} - \nabla_{\bm{x}} \big[  \inner{\frac{1}{2} g^2\bm{s}_{\bm{\theta}_0} - \bm{f}}{\bm{v}}  \big] =0$ with zero initial and boundary conditions admit $\bm{0}$ as the unique strong solution, then $\bm{\epsilon}[{\bm{s}_{\bm{\theta}_0}}] \equiv 0$ and  $\bm{s}_{\bm{\theta}_0}(\bm{x}, 0)\equiv\bm{s}_{\bm{\theta}_0}^{\textup{ODE}}(\bm{x}, 0)$ implies $\bm{s}_{\bm{\theta}_0} \equiv \bm{s}_{\bm{\theta}_0}^{\textup{ODE}}$. 
\end{proposition}     
 
Proposition~\ref{th:conti_strong} implies that if the parametric scores match with the ground truth score at the initial time, the only global minimum is the ground truth score. Essentially,  the scores at any given time can be obtained solely by achieving a flawless alignment of scores at a single timestep through the dynamics of PDE. This indicates the score FPE residual is a proper quantity to measure the gaps between the ground truth and parametric scores. Indeed, this proposition is an extreme case of ``the continuous dependence of PDE solutions on parameters $\bm{\theta}$''~\citep{artstein1975continuous}. A more sophisticated analysis~\citep{lunardi2012analytic,papageorgiou1994solution} can be applied  to prove for instance, that as $\norm{\bm{\epsilon}[{\bm{s}_{\bm{\theta}}}]} \rightarrow 0$, $\norm{\bm{s}_{\bm{\theta}} - \bm{s}_{\bm{\theta}}^{\textup{SDE}}}\rightarrow 0$ if $\bm{f}\equiv 0$ (with a careful choice of norms). However, such technical generalization is outside this work's scope. 
 

\subsection{Higher-order score matching}\label{sec:higher_sm} Higher-order derivatives of the score can yield 
additional information about the data distribution~\citep{meng2021estimating,lu2022maximum}. 
We prove that bounding of the higher-order score matching loss can further control the FPE residual $\norm{\int_{0}^{t} \bm{\epsilon}[{\bm{s_{\theta}}}](\bm{x}, \tau)d\tau}_2$ for all $t\in[0,T]$. 
This partially explains why scores learned via $\mathcal{J}_{\text{DSM}}$ do not satisfy the score FPE, as DSM only matches gradients, while higher-order derivatives may still deviate from the ground truth. 
 
\begin{proposition}\label{th:higher}
    Assume that on $\mathbb{R}^D \times [0, T]$, higher-order score matchings admit the following error bounds: 
    $\norm{\bm{s} - \bm{s}_{\bm{\theta}}}_2 \leq \delta_0$, $\norm{\grad{\bm{x}}(\bm{s} - \bm{s}_{\bm{\theta}})}_F \leq \delta_1$, $\norm{\grad{\bm{x}}\div{\bm{x}}(\bm{s} - \bm{s}_{\bm{\theta}})}_2 \leq \delta_2$. 
    
    Then for all $(\bm{x}, t) \in\mathbb{R}^D\times[0,T]$, $\norm{\int_{0}^{t} \bm{\epsilon}[{\bm{s_{\theta}}}](\bm{x}, \tau)d\tau}_2 $
    \begin{align*}
        &\leq 2\delta_0  + \frac{1}{2} (\delta_2 + 2\delta_1 \delta_0)\int_{0}^{t} g^2(\tau)d\tau
        \\&+ \delta_1 \int_{0}^{t}  \big( g^2(\tau)  \norm{\bm{s}(\bm{x},\tau)}_2+\norm{\bm{f}(\bm{x},\tau)}_2\big) d\tau 
        \\ &+ \delta_0 \int_{0}^{t}\big( g^2(\tau)\norm{\grad{\bm{x}}\bm{s}(\bm{x},\tau)}_F + \norm{\grad{\bm{x}}\bm{f}(\bm{x},\tau)}_F \big) d\tau.
    \end{align*}
\end{proposition}

\section{Training with score FPE-regularizer}
\label{sec:sfpe_reg}

We showed in Section~\ref{subsec:fail_sfpe} that score models learned via $\mathcal{J}_{\text{DSM}}$ (Eq.~\eqref{eq:dsm}) do not satisfy the score FPE, a property that ground truth scores should satisfy \emph{a priori}. Motivated by this fact and Theorem~\ref{th:min_ode_add}, we hence devise a novel regularization term which is called \emph{score FPE-regularizer} and defined as $\mathcal{R}_{\text{FP}}(\bm{\theta})=\mathcal{R}_{\text{FP}}(\bm{\theta}; \alpha, \beta, \lambda_{\text{FP}}(\cdot), m) :=$ 
\begin{align*}
\mathbb{E}_{t\sim\mathcal{U}[0, T]}\mathbb{E}_{\bm{x}(0)} \mathbb{E}_{ \bm{x}(t)|\bm{x}(0)}\Big[& \alpha\cdot \frac{1}{D^m} \norm{\lambda_{\text{FP}}(t)\bm{\epsilon}[{\bm{s_{\theta}}}](\bm{x}, t)}_2^m 
\\& + \beta\cdot\abs{\mathcal{L}[\bm{s}_{\bm{\theta}}](\bm{x}, t)}  \Big].
\end{align*}
Here, $\alpha, \beta \geq 0 $ are parameters controlling the regularization strength, $\lambda_{\text{FP}}(\cdot)$ is the time-weighting function for the score FPE residual, and $m$ is an integer.  $\mathcal{R}_{\text{FP}}$ consists of the score FPE residual and time-derivative taming term, which respectively imitate Eq.~\eqref{eq:sup_avg_fpe} and Eq.~\eqref{eq:sup_avg_scafpe}. With the score FPE-regularizer, we propose a new loss $\mathcal{J}_{\text{FP}}$ which comprises  $\mathcal{J}_{\text{DSM}}$ and $\mathcal{R}_{\text{FP}}$ with $ \mathcal{J}_{\text{FP}}(\bm{\theta})=\mathcal{J}_{\text{FP}}(\bm{\theta}; \lambda(\cdot), \alpha, \beta, \lambda_{\text{FP}}(\cdot), m):=$
\begin{equation}\label{eq:DSM+FP}
   \mathcal{J}_{\text{DSM}}(\bm{\theta}; \lambda(\cdot)) +  \mathcal{R}_{\text{FP}}(\bm{\theta}; \alpha, \beta, \lambda_{\text{FP}}(\cdot), m) ,
\end{equation}
 We refer to a model trained with our proposed  $ \mathcal{J}_{\text{FP}}$ as \emph{FP-Diffusion}. 
We remark that Eq.~\eqref{eq:DSM+FP} returns the vanilla DSM loss (Eq.~\eqref{eq:dsm}) with $\alpha=\beta=0.0$. Hereafter, we take $\lambda(\cdot)=g^2(\cdot)$ in $\mathcal{J}_{\text{DSM}}$. 

Because $\bm{\epsilon}[{\bm{s_{\theta}}}]$ in $\mathcal{R}_{\text{FP}}$ is generally expensive to calculate for high dimensional data, we propose efficient approximations for $\partial_t \bm{s}_{\bm{\theta}}$ and $\div{\bm{x}}(\bm{s}_{\bm{\theta}})$. 

\textbf{Finite difference~\citep{fornberg1988generation} for $\partial_t\bm{s}_{\bm{\theta}}$ } $\partial_t\bm{s}_{\bm{\theta}}$ can be efficiently approximated by finite difference method as the derivative is one-dimensional. 
For high dimensional datasets, we set $(h_s, h_d) = (0.001, 0.0005)$ and approximate $\partial_t\bm{s}_{\bm{\theta}}(\bm{x}, t)$ by 
\begin{equation*}
    \frac{h_s^2 \bm{s}_{\bm{\theta}}(\bm{x}, t+h_d)  +(h_d^2-h_s^2)\bm{s}_{\bm{\theta}}(\bm{x}, t) -h_d^2\bm{s}_{\bm{\theta}}(\bm{x},t-h_s)}{h_s h_d (h_s + h_d)}.
\end{equation*}

\textbf{Hutchinson's estimator~\citep{hutchinson1989stochastic} for $\div{\bm{x}} (\bm{s}_{\bm{\theta}})$ } Hutchinson's trace estimator 
stochastically estimates the trace of any square matrix. As  $\div{\bm{x}} (\bm{s}_{\bm{\theta}}) = \textup{tr}\big(\grad{\bm{x}} \bm{s}_{\bm{\theta}}\big)$, we can apply Hutchinson's trick and replace the $\div{\bm{x}} (\bm{s}_{\bm{\theta}})$ term with an estimation 
$$ \frac{1}{M}\sum_{j=1}^{M}\bm{v}_j\grad{\bm{x}} \bm{s}_{\bm{\theta}}(\bm{x}, t)\bm{v}_j^T,$$ where $\bm{v}_j\sim\mathcal{N}(\bm{0}, \bm{I})$. We set $M=1$, following \citet{song2020sliced} which works well in practice. 

In Appendix~\ref{sec:technique}, we present supplementary information encompassing a theoretical analysis of the error involved in estimating the score FPE, as well as a potential technique that can enhance computational efficiency in this regard.
Moreover, we supplement with runtime comparison in Appendix~\ref{subsec:runtime}.

\section{Empirical implications of score FPE
}
\label{sec:add_exp}


In this section, we investigated two implications of the score FPE. First, we examined the solvability of scores through a Cauchy problem associated with the score FPE (as stated in Proposition~\ref{th:fp}). Second, we investigated how reducing the score FPE residual enhances the conservativity of a model (as described in Proposition~\ref{th:conservativity}).

\subsection{Scores learning  by solving Cauchy problems }\label{subsec:solve_pde}
\begin{figure}[th]
     \centering
     \resizebox{\linewidth}{!}{
     \subfigure[Generated samples by $\tilde{\bm{s}}_{\bm{\theta}}^{\text{GMM}}$]{\includegraphics[width=0.49\columnwidth]{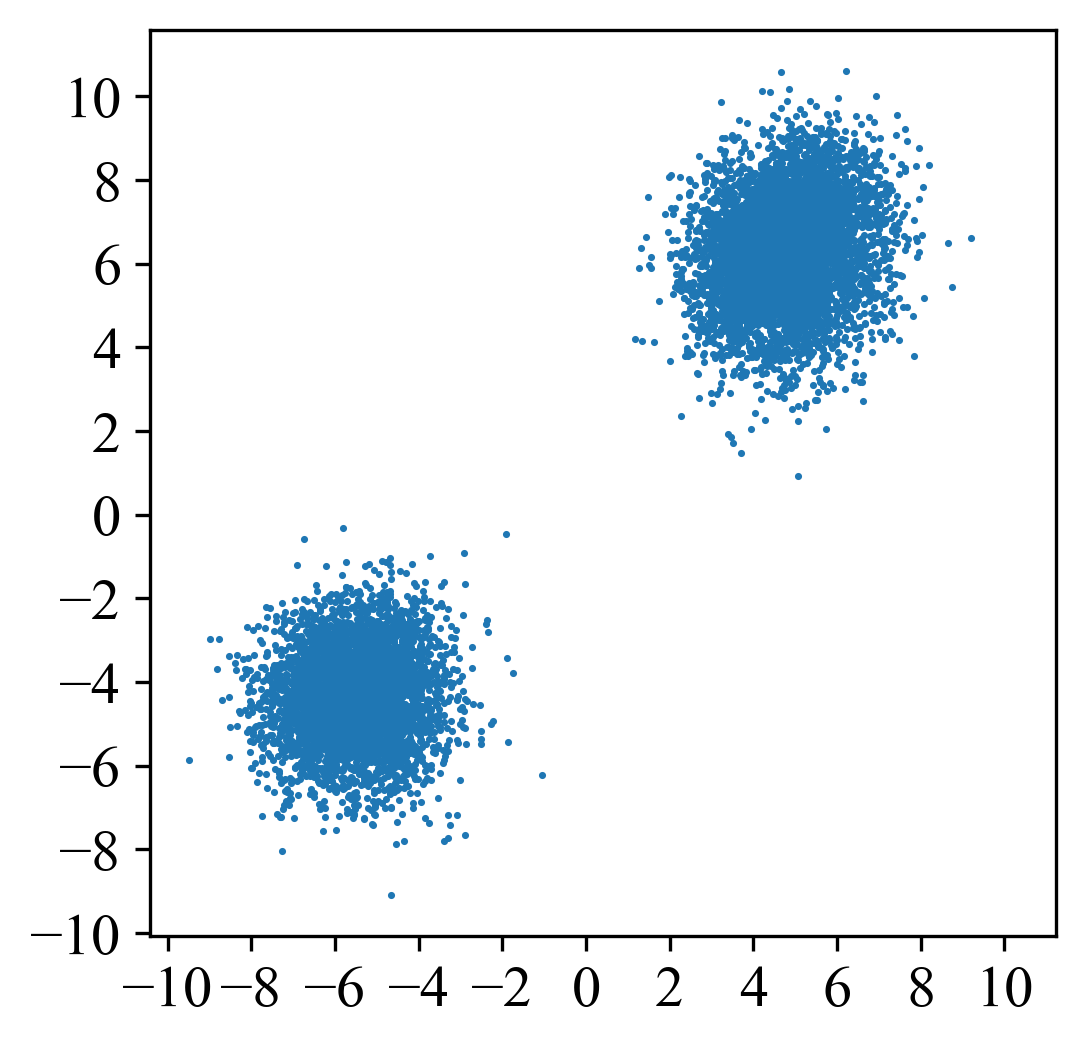}}
    \subfigure[Estimated density by $\tilde{\bm{s}}_{\bm{\theta}}^{\text{GMM}}$]{\includegraphics[width=0.49\columnwidth]{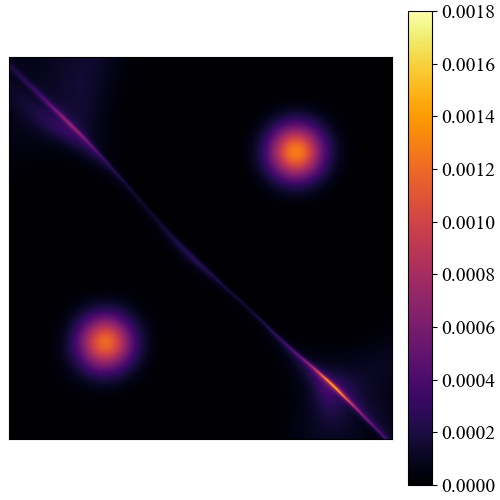} }
    }
        \caption{
        (a) visualizes instances generated by $\bm{\tilde{s}_{\theta}^{\textup{GMM}}}$. (b) shows the estimated density via the probability flow ODE of $\bm{\tilde{s}_{\theta}^{\textup{GMM}}}$. Scores at all times can be obtained by solving a Cauchy problem of the score FPE.}
         \label{fig:fpe_guided_gmm}
\end{figure}

 Here, we consider the data distribution as a 2D GMM $\frac{1}{5}\mathcal{N}\big((-5, -5), \bm{I} \big) + \frac{4}{5} \mathcal{N}\big((5, 5), \bm{I} \big)$. The diffusion process is taken as VE SDE (Eq.~\eqref{eq:vesde}). The  ground truth score of a 2D GMM, denoted as $\bm{s}^{\textup{GMM}}$, can be expressed explicitly in a closed form throughout the diffusion (as the diffusion process is linear in $\bm{x}$). In Section~\ref{sec:FP_derivation}, we explained that the score at all times can theoretically be solved, given the score at a single time step. That is, the score is a solution $\bm{\tilde{s}}$ to the following Cauchy problem on the system of PDEs: 
 \begin{equation}\label{eq:solv_fp}
    \begin{cases}
    \begin{aligned}
    &\partial_t \bm{\tilde{s}} (\bm{x}, t) =\grad{\bm{x}}\mathcal{L}[\bm{\tilde{s}}](\bm{x}, t), \quad  (\bm{x}, t) \in \mathbb{R}^D\times(0,T] \\ 
    &\bm{\tilde{s}}(\bm{x}, 0) =\bm{s}^{\textup{GMM}}(\bm{x}, 0), \quad  \bm{x} \in \mathbb{R}^D,   
    \end{aligned}
    \end{cases}
\end{equation}
where we recall $\mathcal{L}[\bm{\tilde{s}}]=\frac{1}{2}g^2 \div{\bm{x}}(\bm{\tilde{s}}) + \frac{1}{2}g^2\norm{\bm{\tilde{s}}}^2_2$.
We fulfill this idea by parametrizing solutions of Eq.~\eqref{eq:solv_fp} via neural networks $\bm{\tilde{s}_{\theta}^{\textup{GMM}}}$~\citep{raissi2019physics, blechschmidt2021three} and learning an optimal $\bm{\theta}$ to minimize:  
\begin{equation}\label{eq:fp_guided}
\begin{aligned}
    &\mathbb{E}_{t\sim\mathcal{U}[0, T]}\mathbb{E}_{\bm{x}(0)} \mathbb{E}_{ q_{0t}(\bm{x}(t)|\bm{x}(0))}\norm{\bm{\epsilon}[\bm{\tilde{s}_{\theta}^{\textup{GMM}}}](\bm{x}, t)}_2 
       \\&+ \mathbb{E}_{\bm{x}(0)} \norm{\bm{\tilde{s}_{\theta}^{\textup{GMM}}}(\bm{x}, 0) - \bm{s}^{\textup{GMM}} (\bm{x}, 0)}_2. 
\end{aligned}
\end{equation}
Interestingly, as shown in Figures~\ref{fig:fpe_guided_gmm}(a) and (b), respectively, $\bm{\tilde{s}_{\theta}^{\textup{GMM}}}$ generates satisfactory samples  and enables good density estimation. This supports our argument that all temporal score information can be obtained by solving the score FPE. Generally, an initial condition to match the ground truth score is impractical. Nevertheless, this opens up the possibility of learning diffusion models from noisy data by substituting the exact score matching (with the ground truth) at the initial time with a noise-contaminated score matching (i.e., denoising score matching trick).


\subsection{Reduction of score FPE residual implies conservativity }\label{subsec:conservative_exp}

\begin{figure}[th]
\centering\resizebox{0.85\linewidth}{!}{
\begin{minipage}{\columnwidth}
  \centering
  \includegraphics[width=\textwidth]{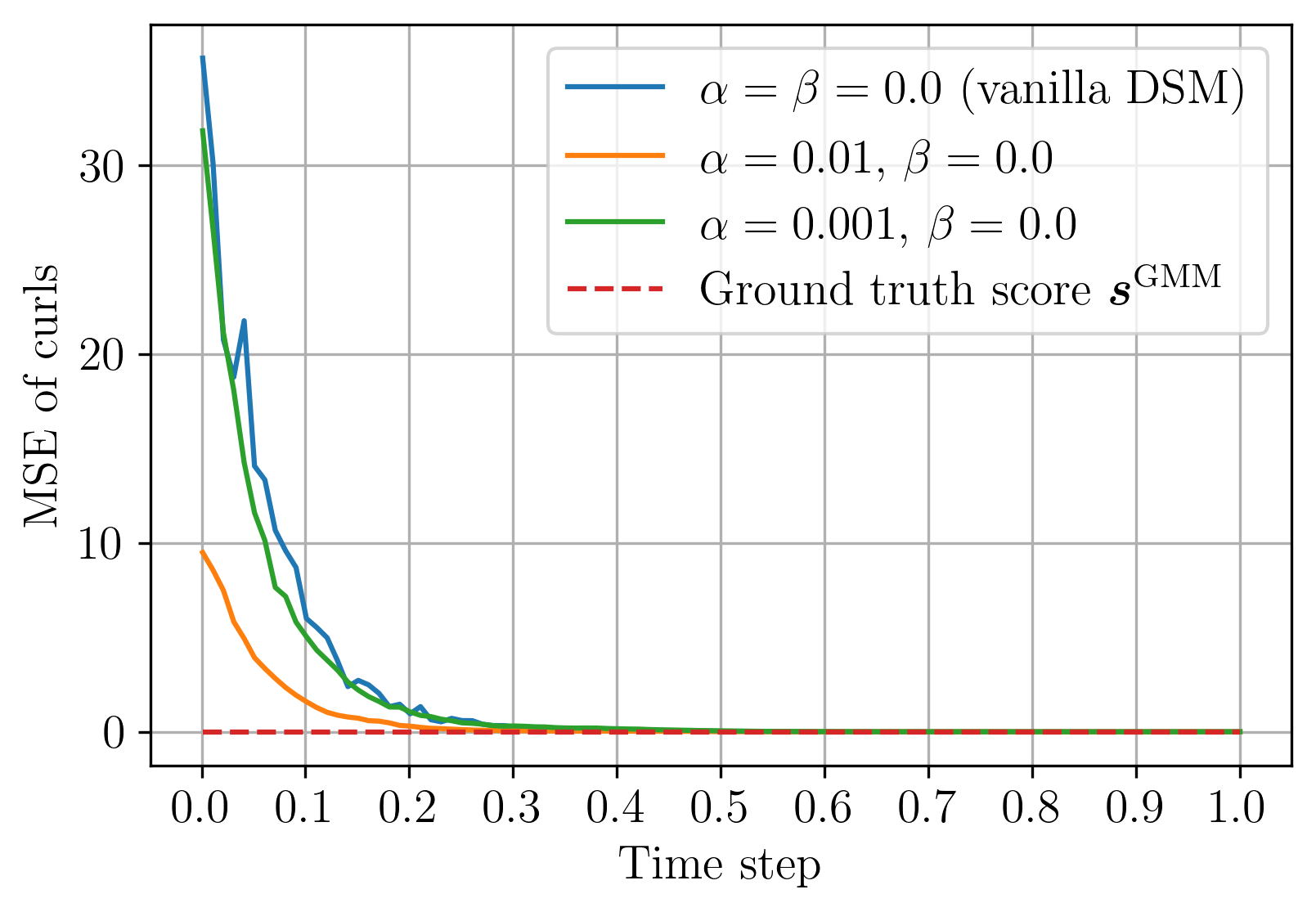}
\end{minipage}%
}
 \caption{Comparison of the MSEs of curls. Scores trained with the score FPE-regularizer tend to be conservative. }
 \label{fig:curl}
\end{figure}

We take the 2D GMM described in Section~\ref{subsec:solve_pde} as the data distribution. It is known that a vector field $\bm{F}:=(F_1, F_2, F_3)\colon\mathbb{R}^3\rightarrow\mathbb{R}^3$ is conservative if and only if its ``curl'', $\big(\partial_{\bm{x}_2}F_3 - \partial_{\bm{x}_3}F_2, \partial_{\bm{x}_3}F_1 - \partial_{\bm{x}_1}F_3, \partial_{\bm{x}_1}F_2 - \partial_{\bm{x}_2}F_1\big)$, is zero. Thus, considering the mean squared error (MSE) of their curls quantifies the degree of conservativity of the score. We compared these values for the following four cases: scores trained 
 (a) from Eq.~\eqref{eq:dsm}, (b) from Eq.~\eqref{eq:DSM+FP} with $(\alpha, \beta, \lambda_{\text{FP}}(\cdot), m) = (0.001, 0.0, 1.0, 1)$, (c) and from Eq.~\eqref{eq:DSM+FP} with $(\alpha, \beta, \lambda_{\text{FP}}(\cdot), m) = (0.01, 0.0, 1.0, 1)$, along with (d) the ground truth score. Figure~\ref{fig:curl} plots the MSEs of the curls of the trained and ground truth scores for each timestep. The time-averaged MSEs of curls of the four scores are $2.22$, $1.89$, $0.60$, and $3.73e-13$, respectively. We observed that the ground truth score is numerically conservative by its nature, and that scores trained with the score FPE-regularizer tend to be conservative, which empirically supports Proposition~\ref{th:conservativity}. 

\section{Density Estimation Experiments}\label{sec:exp}

We examined the effectiveness of $\mathcal{J}_{\text{FP}}$ on three synthetic datasets, MNIST, Fashion MNIST, CIFAR-10, and ImageNet32. Appendix~\ref{sec:implement} gives the implementation details and Appendix~\ref{subsec:ill-samples} visualizes randomly generated examples. We released our code at \url{https://github.com/sony/FP-Diffusion}.

\begin{figure*}[th]
\centering\resizebox{0.85\linewidth}{!}{
\begin{minipage}{.33\textwidth}
  \centering
  \includegraphics[width=\textwidth]{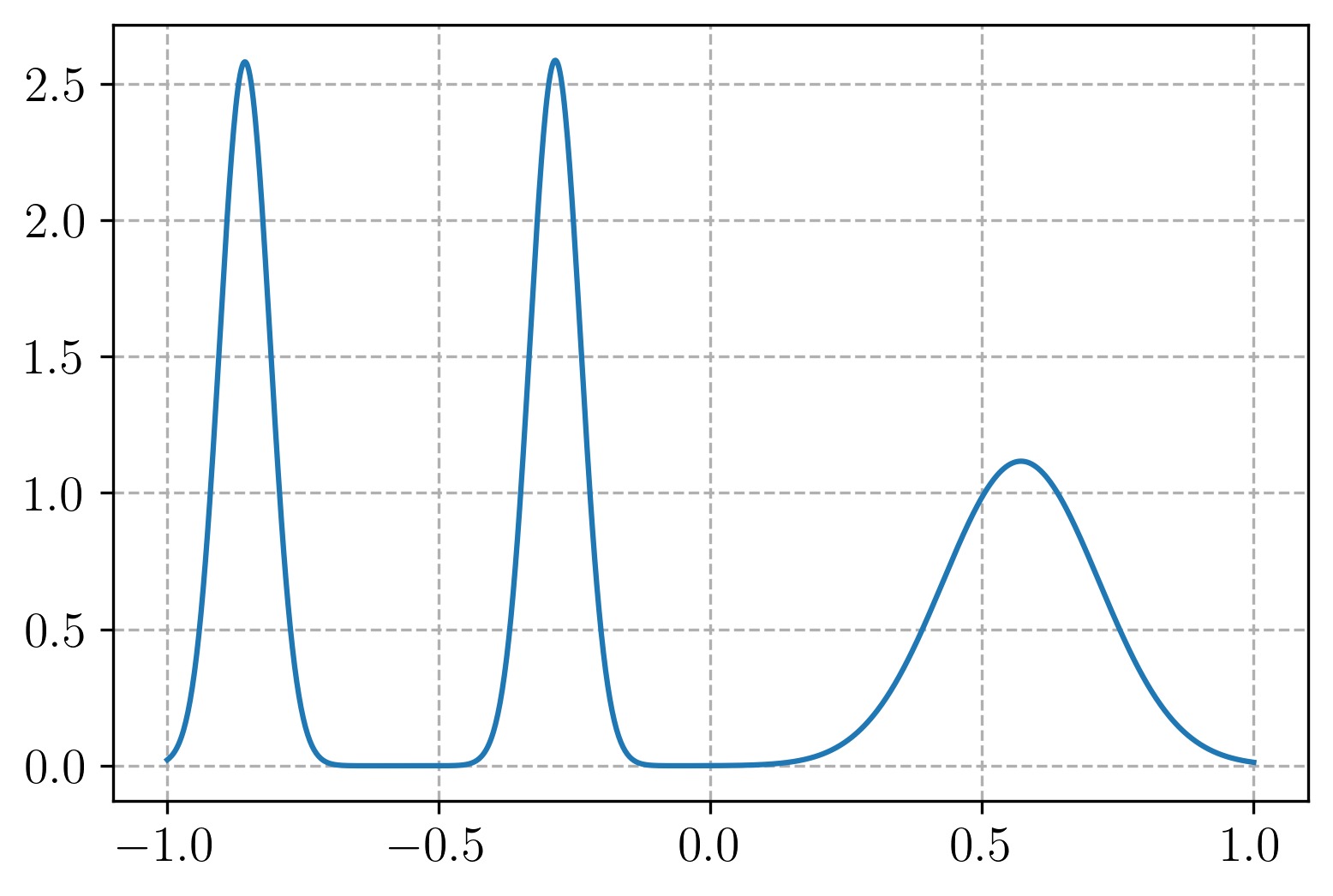}
  \\ (a) Data density
\end{minipage}%
\begin{minipage}{.33\textwidth}
  \centering
  \includegraphics[width=\textwidth]{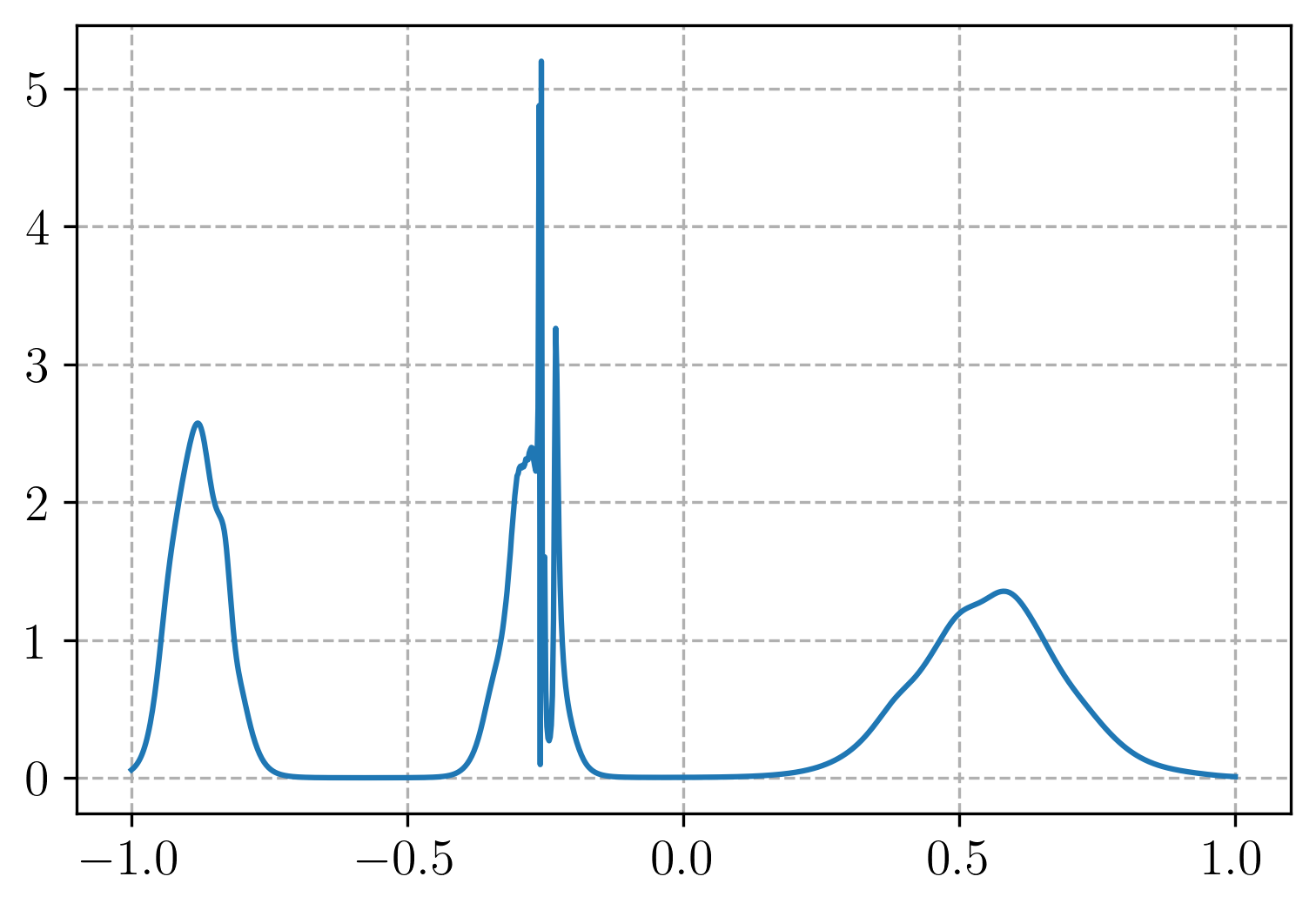}
  \\ (b) Vanilla $\text{DSM} $ ($\alpha=\beta=0.0$)
\end{minipage}%
\begin{minipage}{.33\textwidth}
  \centering
  \includegraphics[width=\textwidth]{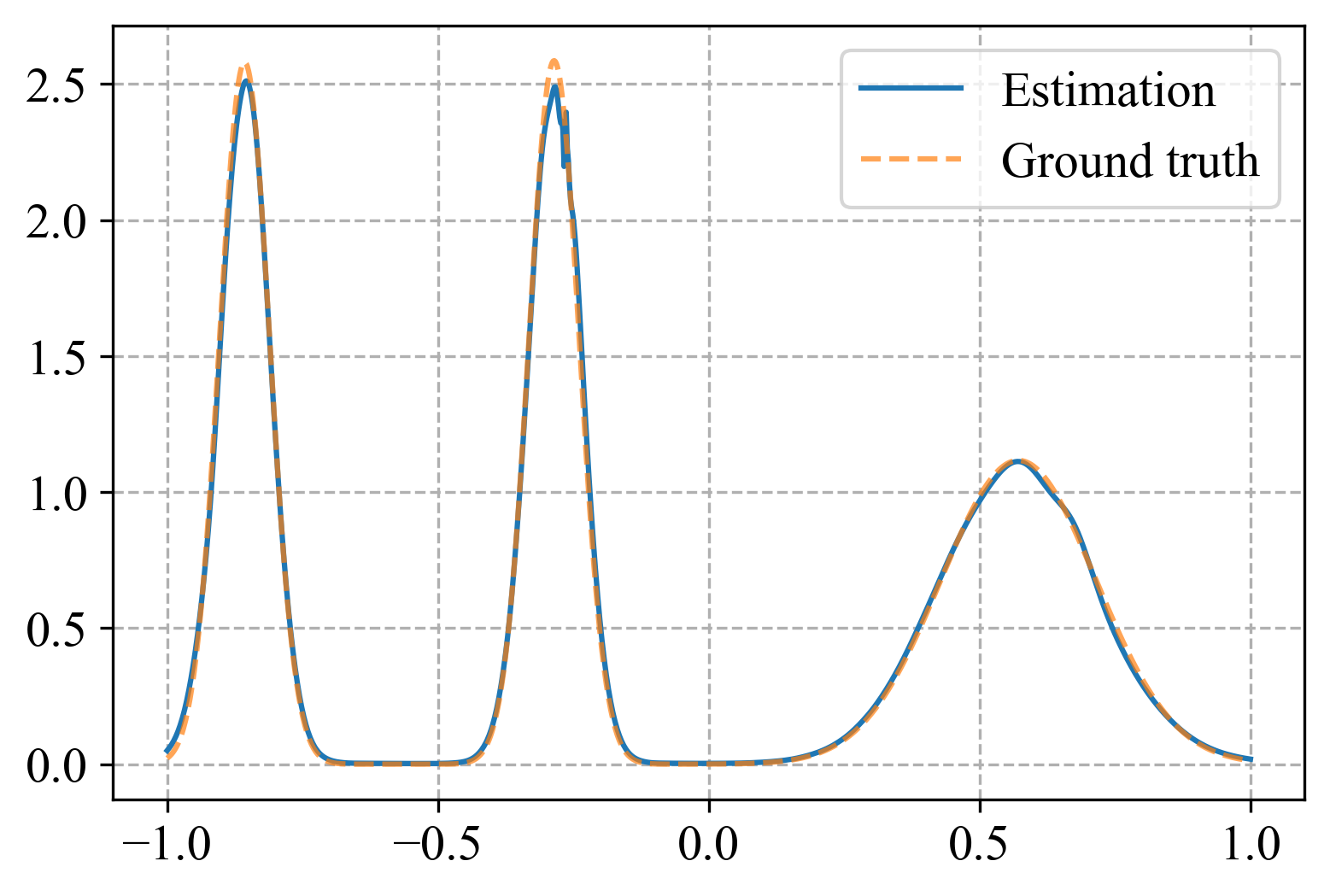}
  \\ (c) FP-Diffusion (with $\alpha = 0.0015$)
\end{minipage}}
 \caption{(a) demonstrates the ground truth data density. We compare (b) estimated density by probability flow ODE with $\bm{s}_{\bm{\theta}}$ trained with $\alpha=\beta=0.0$, and (c) with $\alpha = 0.0015$. Score FPE-regularizer improves density estimation.}
 \label{fig:comparison_fp_dsm}
\end{figure*}

\begin{figure*}[th]
\centering\resizebox{\linewidth}{!}{
\begin{minipage}{.44\textwidth}
  \centering
  \includegraphics[width=\textwidth]{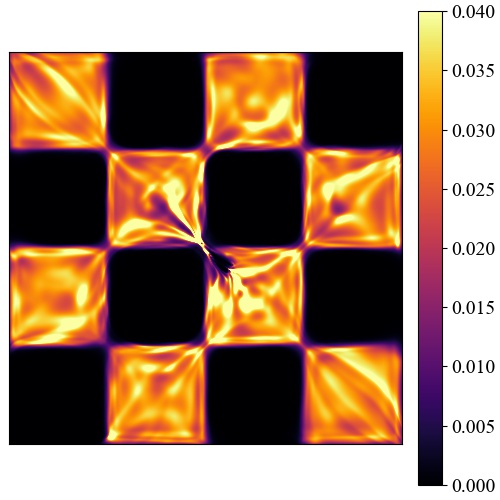}
  \\ \huge{(a) Vanilla DSM \\($\alpha=\beta=0.0$)}
\end{minipage}%
\begin{minipage}{.44\textwidth}
  \centering
  \includegraphics[width=\textwidth]{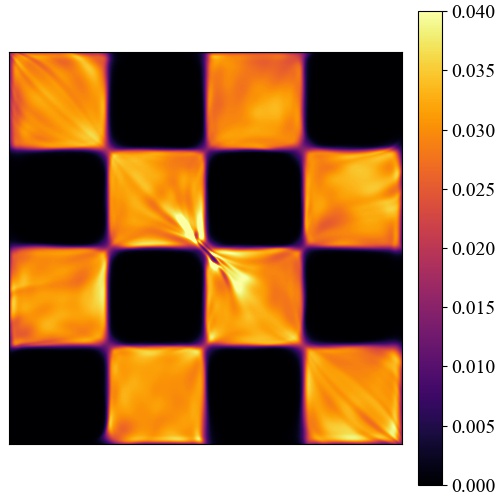}
  \\ \huge{(b) FP-Diffusion \\ (with $\alpha = 0.0015$)}
\end{minipage}
\begin{minipage}{.44\textwidth}
  \centering
  \includegraphics[width=\textwidth]{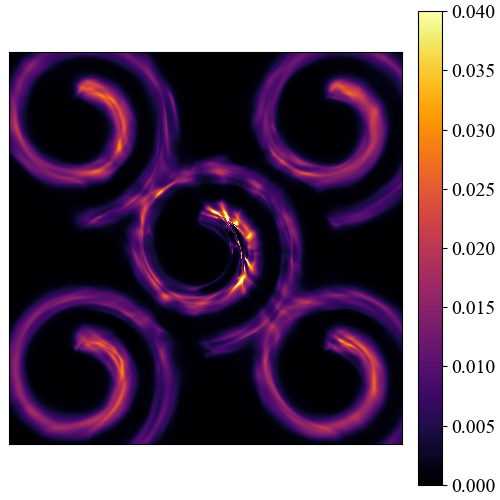}
  \\ \huge{(c) Vanilla DSM \\ ($\alpha=\beta=0.0$)}
\end{minipage}%
\begin{minipage}{.44\textwidth}
  \centering
  \includegraphics[width=\textwidth]{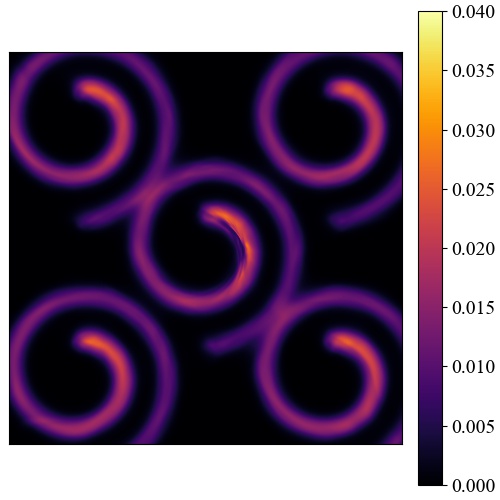}
  \\ \huge{(d) FP-Diffusion\\ (with $\alpha = 0.0015$)}
\end{minipage}
\begin{minipage}{.44\textwidth}
  \centering
  \includegraphics[width=\textwidth]{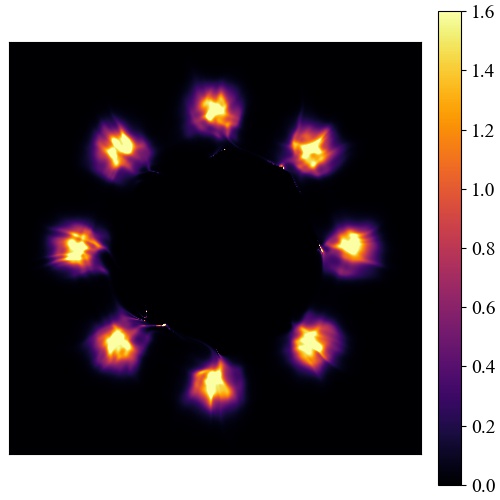}
  \\ \huge{(e) Vanilla DSM\\ ($\alpha=\beta=0.0$)}
\end{minipage}%
\begin{minipage}{.44\textwidth}
  \centering
  \includegraphics[width=\textwidth]{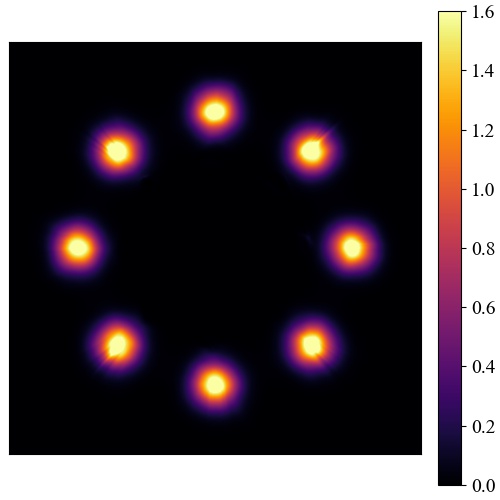}
  \\ \huge{(f) FP-Diffusion\\ (with $\alpha = 0.0015$)}
\end{minipage}
}
 \caption{Estimated densities on a 2D checkerboard, multiple Swiss rolls, and eight GMMs, respectively. (a), (c), and (e) show estimated densities via the probability flow ODE of $\bm{s}_{\bm{\theta}}$ trained with $\alpha=\beta=0.0$ (vanilla DSM). In contrast, (b), (d), and (e) show the densities via training with $\alpha = 0.0015$. The score FPE-regularizer estimated the data densities well.}
 \label{fig:comparison_ckb_sw}
\end{figure*}

\subsection{Synthetic datasets}\label{subsec:exp_synthetic} 

We compared and visualized density estimation via models trained with vanilla $\mathcal{J}_{\text{DSM}}$ (Eq.~\eqref{eq:dsm}) and the proposed $\mathcal{J}_{\text{FP}}$ (Eq.~\eqref{eq:DSM+FP}) with fixed $(\alpha, \beta, \lambda_{\text{FP}}(\cdot), m) = (0.0015, 0.0, 1.0, 1)$. Here, the forward SDE is taken as a VE type. We examined the models' performance across three synthetic datasets: a 1D GMM with three modes $\frac{3}{10}\mathcal{N}\big(-\frac{6}{7}, (\frac{1}{70})^2 \big) + \frac{3}{10} \mathcal{N}\big(-\frac{2}{7}, (\frac{1}{70})^2 \big) + \frac{4}{10} \mathcal{N}\big(\frac{4}{7}, (\frac{1}{7})^2\big)$, a 2D checkerboard, Swiss rolls, and a 2D Gaussian mixture models (GMM) with eight modes whose means are located equidistant on the unit circle and with a standard deviation 1. We refer to Appendix~\ref{subsec:implement-simple} for more details.

 For all datasets, scores trained with the score FPE-regularizer, as shown in Figure~\ref{fig:comparison_fp_dsm}(b) and Figures~\ref{fig:comparison_ckb_sw}(b), (d), and (f), can approximate the data density well, with improvement over  vanilla score matching, as shown in Figure~\ref{fig:comparison_fp_dsm}(c) and Figures~\ref{fig:comparison_ckb_sw}(a), (c), and (e). This reinforces the implication of Theorem~\ref{th:min_ode} that the score FPE-regularizer may improve density estimation of the probability flow ODE, as it enforces a known self-consistency property of the ground truth score.


\subsection{MNIST and Fashion MNIST}\label{subsec:exp_mnist}
We trained models with the proposed $\mathcal{J}_{\text{FP}}$ on MNIST and Fashion MNIST with different $\alpha$'s values from scratch, and we evaluated the test set negative log-likelihood (NLL) in terms of bits/dim (bpd). In FP-Diffusion, the rest of parameters were fixed as $(\beta, \lambda_{\text{FP}}(\cdot), m) = (0.0, 1.0, 1)$. Table~\ref{tb:nll_mnist_fmnist}  reports the averaged NLLs over five repeated runs of likelihood computations and three different initializations for training across two instantiations of the forward SDE (including VE, and VP) of various choices of $\alpha$. A lower NLL indicates a better performance. We observed a general improvement in the NLL with $\alpha = 0.1$. To better understand the choice of the hyper-parameter $\alpha$, ignoring dirt effects that result from different training initializations, we compare the NLLs via fine-tuning from pre-trained models in Appendix~\ref{subsec:sensitivity}. 



\begin{table}[th]
\caption{NLL comparisons on MNIST and Fashion MNIST}
\centering
\resizebox{\columnwidth}{!}{
\begin{tabular}{lccccc}
\hline
 & \multicolumn{2}{c}{\textbf{MNIST} } & \multicolumn{1}{c}{ }
 &\multicolumn{2}{c}{\textbf{Fashion MNIST} } \\ \cline{2-3} \cline{5-6}
\textbf{Method}  & VE     & VP        &  & VE & VP               \\ \hline
Vanilla~\citep{song2020score}   &  3.73 & 3.24   &  &  4.76 & 4.46  \\
FP-Diffusion ($\alpha=0.001$)    & 3.64  &  3.17  &  &  4.67 & 4.50 
\\
FP-Diffusion ($\alpha=0.01$)    & 3.58  &  3.12  &  &  4.61 & 4.36    
\\
FP-Diffusion ($\alpha=0.1$)   &  3.42 & \textbf{2.98}    &  & \textbf{4.40}  & \textbf{4.21}  
\\
FP-Diffusion ($\alpha=1.0$)    &  \textbf{3.30} & 3.11  &  & 4.44  & 4.36 
\\
FP-Diffusion ($\alpha=10.0$) & 3.31  & 3.21    &  &  4.46 & 4.67  
\\
 \hline
\end{tabular}
}
\label{tb:nll_mnist_fmnist}
\end{table}

\subsection{CIFAR-10 and ImageNet32}\label{subsec:exp_cifar}

We fine tuned the pre-trained VE models from the checkpoints of \citet{song2020score,lu2022maximum} by training them for $0.1$M additional iterations on CIFAR-10 and ImageNet32, respectively. Here, we set the hyper-parameters of FP-Diffusion as $(\alpha, \beta, \lambda_{\text{FP}}(\cdot), m) = (0.15, 0.01, g^2(\cdot), 2)$, but we also explored different choices as described  in Appendix~\ref{subsec:sensitivity}. Table~\ref{tb:nll_cifar_imagenet} reports the averaged NLLs of probability flow ODE on the test dataset over five repeated runs. Compared with vanilla DSM~\citep{song2020score}, FP-Diffusion significantly improved the NLL. Moreover, FP-Diffusion was competitive with higher-order DSM~\citep{lu2022maximum}, where we re-computed the NLL based on their checkpoints but also indicated their reported results in the parentheses. On CIFAR-10, we noticed that FP-Diffusion trained with VE and VE-deep architectures may obtain inferior FID scores: $10.83$ and $4.51$, respectively; compared with FID scores of vanilla models: $3.33$ and $2.44$. However, the difference is generally imperceptible (see Appendix~\ref{subsec:ill-samples} for illustration and quantitative measurements). 
Moreover, Appendix~\ref{subsec:fp-diff-enforce} provides empirical evidence demonstrating that FP-Diffusion exhibits superior adherence to the score FPE compared to the vanilla model.


\begin{table}[th]
\caption{NLL comparisons on CIFAR-10 and ImageNet32 }
\centering
\resizebox{\columnwidth}{!}{
\begin{tabular}{lcccc}
\hline
 & \multicolumn{2}{c}{\textbf{CIFAR-10} } & \multicolumn{1}{c}{ }
 &\multicolumn{1}{c}{\textbf{ImageNet32} } \\ \cline{2-3} \cline{5-5}
\textbf{Method}  & VE     & VE-deep     &       & VE              \\ \hline
FP-Diffusion                          & \textbf{3.36}    &   3.32  &  &  \textbf{3.77}                 
\\ \hline
Vanilla~\citep{song2020score}        &  3.61 (3.66)   &  3.42  &  &  4.01 (4.21)     
\\
2nd DSM~\citep{lu2022maximum}              &  3.44 & 3.35  &   &  3.82  (4.06)   
\\
3rd DSM~\citep{lu2022maximum}     &  3.38  & \textbf{3.31} (\textbf{3.27}) &   &  3.80  (4.02)  
\\
 \hline
\end{tabular}
}
\label{tb:nll_cifar_imagenet}
\end{table}



\section{Related work and discussions}\label{sec:literature}
Conservativity is a key property for understanding the consistency between the learned density and the ground truth density, as the latter is inherently conservative. Despite the dominance of diffusion models over energy-based methods (which are inherently conservative), exploring conservativity can offer valuable insights into their underlying mechanisms and potentially enhance both types of models.

\citet{salimans2021should} adopted a special parameterization to ensure conservativity and compared with energy-based models. In contrast, \citet{chao2022quasi} imposed a penalty to reach zero curl (i.e., conservativity), independently of the model architecture. 
In addition, they empirically showed that non-conservative scores may incur rotational vector fields tangent to the true score function, leading to inefficient updates during the sampling processes. However, how the non-conservativity affects the sampling theoretically and empirically is not well studied in the literature of diffusion models. Although enforcement of conservativity of diffusion models is not the main purpose of this work, conservativity is one of the outcomes by reducing the score FPE residual. Nevertheless, score FPE provides a different framework from the PDE perspective, which is theoretically solid and may stimulate further study. 

On the other hand, researchers have also attempted to theoretically explain the success of diffusion models by studying the gap between the data and learned densities. \citet{de2021diffusion} proved error bounds for these densities in terms of the total variation. \citet{song2021maximum} showed the likelihood of the diffusion model can be bounded by the score matching objective with a specific choice of temporal weighting. \citet{chen2022improved} provided a convergence analysis for any data distribution with second-order moment in KL divergence.  \citet{kwon2022score} found that minimization of the score matching loss may implicitly reduce the Wasserstein-2 distance between the data and learned density. 
\citet{meng2021estimating} introduced the concept of estimating higher-order gradients of a data distribution. Later, \citet{lu2022maximum} extended the idea and showed that the likelihood from the deterministic trajectory of a diffusion model may be improved by matching higher-order scores. 

\citet{shen2022self} showed that the asymptotic fixed point of the velocity field associated with the classic FPE \citep{fokker1914mittlere,planck1917satz} (governing the density evolution) can recover the solution of FPE in the Wasserstein-2 sense. However, its study was neither adapted to diffusion models nor generative models.

\section{Conclusion}\label{sec:conclude} We introduce the score FPE and theoretically study its relationship with likelihood improvements, conservativity, higher-order score matching, and scores induced by a parametric reverse diffusion.  Moreover, we propose to regularize models by enforcing  consistency properties of the ground truth score through the score FPE, and show this achieves better density estimation and likelihoods on various datasets. We empirically support our theory by finding that reduction of the score FPE residual improves the conservativity of a model. The Cauchy problem defined with the score FPE can be used to obtain time-conditioned scores directly by PDEs solving. Incorporating more advanced numerical methods for solving PDEs is an interesting avenue for future research.

\section*{Acknowledgements}
We would like to thank Lucas Mauch for their variable comments during the preparation of this manuscript. Additionally, we sincerely appreciate anonymous reviewers for their insightful feedback and suggestions.

\bibliography{sbm_bib}
\bibliographystyle{icml2023}

\clearpage
\appendix
\onecolumn

\newpage

\tableofcontents

\begin{itemize}
    \item[] \hyperref[sec:instances]{A. Instantiations of forward SDEs and corresponding score FPEs} 
    \item[] \hyperref[sec:fp-supp]{B. How scores satisfy score FPEs} 
        \begin{itemize}
            \item[] \hyperref[subsec:supp-sec-3]{B.1. Supportive experiments for Section~\ref{sec:FP_derivation}}
            \item[] \hyperref[subsec:fp-diff-enforce]{B.2. FP-Diffusion enforces satisfaction of score FPE}
        \end{itemize}
    \item[] \hyperref[sec:technique]{C. More details on techniques for efficient score FPE computation }
        \begin{itemize}
            \item[] \hyperref[sec:reduce1]{C.1. Trick to reduce computation cost of \texorpdfstring{$\partial_t\bm{s}_{\bm{\theta}}$}{TEXT}}
            \item[] \hyperref[sec:reduce2]{C.2. Trick to reduce computation cost of  \texorpdfstring{$\div{\bm{x}} (\bm{s}_{\bm{\theta}})$}{TEXT}}
            \item[] \hyperref[sec:reduce-error-bound]{C.3. Error analysis of score FPE estimation}
            \item[] \hyperref[sec:reduce3]{C.4. Potential technique to compute \texorpdfstring{$\bm{\epsilon}[{\bm{s_{\theta}}}]$}{TEXT} more efficiently}
        \end{itemize}
    
    \item[] \hyperref[sec:implement]{D. Implementation details} 
        \begin{itemize}
            \item[] \hyperref[subsec:implement-simple]{D.1. Synthetic dataset}
            \item[] \hyperref[subsec:implement-mnist-fmnist]{D.2. MNIST and Fashion MNIST}
            \item[] \hyperref[subsec:implement-complex]{D.3. CIFAR-10 and ImageNet32}
        \end{itemize}    
    
    \item[] \hyperref[sec:sensitivity]{E. Supplemental results} 
        \begin{itemize}
        \item[] \hyperref[subsec:sensitivity]{E.1. Sensitivity to hyper-parameters}
        \item[] \hyperref[subsec:runtime]{E.2. Runtime discussion}
        \item[] \hyperref[subsec:ill-samples]{E.3. Illustration and quality of generated samples}
    \end{itemize}
    \item[] \hyperref[sec:assumptions]{F. Theoretical assumptions}
    \item[] \hyperref[sec:proofs]{G. Proofs and discussions}
        \begin{itemize}
            \item[] \hyperref[subsec:pf-prop3.1]{G.1. Proof of Proposition~\ref{th:fp}}
            \item[] \hyperref[subsec:pf-thm4.2]{G.2. Proof of Theorem~\ref{th:min_ode}}
            \item[] \hyperref[subsec:disc-th-min-ode]{G.3. Discussions on Theorem~\ref{th:min_ode}}
            \item[] \hyperref[subsec:pf-thm4.3]{G.4. Proof of Theorem~\ref{th:min_ode_add}}
            \item[] \hyperref[subsec:proof-conservativity]{G.5. Proof and discussion of Proposition~\ref{th:conservativity}}
            \item[] \hyperref[subsec:pf-prop4.5]{G.6. Proof of Proposition~\ref{th:conti_strong}}
            \item[] \hyperref[subsec:pf-prop4.6]{G.7. Proof of Proposition~\ref{th:higher}}
            \item[] \hyperref[subsec:proof-error]{G.8. Proof of Proposition~\ref{th:error_analysis}}
        \end{itemize}
\end{itemize}

\section{Instantiations of forward SDEs and corresponding score FPEs}\label{sec:instances}

\citet{song2020score} categorizes the forward SDE into three types based on the behavior of the variance during evolution. Here, we focus on two types: the
\underline{V}ariance \underline{E}xplosion (VE) SDE and \underline{V}ariance \underline{P}reserving (VP) SDE. 
\paragraph{VE SDE} With a zero drift term $\bm{f}=0$ and a diffusion term $g(t)=\sqrt{\frac{d \sigma^2(t)}{dt}}$ for some function $\sigma(t)$, the forward SDE (Eq.~\eqref{eq:sde_forward}) becomes the following:
\begin{equation}\label{eq:vesde}
    d\bm{x}(t) =  \sqrt{\frac{d \sigma^2(t)}{dt}} d\bm{w}_t.
\end{equation} 
A typical instance of a VE SDE is Score Matching of Langevin Dynamics (SMLD)~\citep{song2019generative}, where $\sigma(t):=\sigma_{\textup{min}}\Big(\frac{\sigma_{\textup{max}}}{\sigma_{\textup{min}}} \Big)^t$ for $t\in(0,1]$. In our implementation, we follow the conventional setup of $(\sigma_{\textup{min}}, \sigma_{\textup{max}}):=(0.01, 50)$.


\paragraph{VP SDE}Let $\beta$ be a non-negative function of $t$. A VP SDE has a linear drift term $\bm{f}(\bm{x}, t)=-\frac{1}{2}\beta(t)\bm{x}$ and a diffusion term $g(t)=\sqrt{\beta(t)}$. Thus, the forward SDE is
$$d\bm{x}(t) = -\frac{1}{2}\beta(t) \bm{x}(t) dt + \sqrt{\beta(t)} d\bm{w}_t.$$
 A classic example of a VP SDE is Denoising Diffusion Probabilistic Modeling (DDPM)~\citep{sohl2015deep,ho2020denoising}, where $\beta(t):=\beta_{\textup{min}}+ t(\beta_{\textup{max}} - \beta_{\textup{min}})$ for $t\in[0,1]$. We adopt the common setup of $(\beta_{\textup{min}}, \beta_{\textup{max}}):=(0.1, 20)$ in our implementation.

Table~\ref{tb:sde_instances} summarizes the aforementioned SDE instantiations and their associated score FPEs.
    


\begin{table}[th]
  \caption{Summary of forward SDEs and their score FPEs}
  \small
  \centering
  \resizebox{0.8\textwidth}{!}{
  \begin{tabular}{cccc}
     \toprule
    
         &   \textbf{VE SDE}   & \textbf{VP SDE}  \\
        \midrule
    $\bm{f}(\bm{x}, t)$ & $\bm{0} $   &  $-\frac{1}{2}\beta(t)\bm{x}$  \\
      $g(t)$   & $\sigma_{\textup{min}}\Big(\frac{\sigma_{\textup{max}}}{\sigma_{\textup{min}}} \Big)^t\sqrt{2\log \big(\frac{\sigma_{\textup{max}}}{\sigma_{\textup{min}}}\big)}$
       &  $\sqrt{\beta(t)}$  \\

    SDE   &   $d\bm{x}(t) =  g(t) d\bm{w}_t$    &   $d\bm{x}(t) = -\frac{1}{2}\beta(t) \bm{x}(t) dt + \sqrt{\beta(t)} d\bm{w}_t$ \\
    \midrule
     Score FPE   & $\partial_t \bm{s} = \grad{\bm{x}}\big[\frac{1}{2}g^2(t) \div{\bm{x}}(\bm{s}) + \frac{1}{2}g^2(t)\norm{\bm{s}}^2_2  \big] $      &  $\partial_t \bm{s} = \frac{1}{2}\beta(t) \grad{\bm{x}}\big[ \div{\bm{x}}(\bm{s}) + \norm{\bm{s}}^2_2  + \inner{\bm{x}}{\bm{s}}\big] $  \\
 \bottomrule
  \end{tabular}
  }
  \label{tb:sde_instances}
\end{table}

\section{How scores satisfy score FPEs}\label{sec:fp-supp}

\subsection{Supportive experiments for Section~\ref{sec:FP_derivation}}\label{subsec:supp-sec-3}

In this section, we further demonstrate how score functions should satisfy the score FPE empirically. We treat the data distribution as a 2D GMM $\frac{1}{5}\mathcal{N}\big((-5, -5), \bm{I} \big) + \frac{4}{5} \mathcal{N}\big((5, 5), \bm{I} \big)$ as in Section~\ref{subsec:solve_pde} and use the same notations. The diffusion process is taken as a VE SDE (Eq.~\eqref{eq:vesde}).

\begin{figure}[th]
     \centering
     \subfigure[ FP residuals of the ground truth score and the score learned from Eq.~\eqref{eq:fp_guided}]{\label{fig:gmm_fp_gt}\includegraphics[width=0.43\textwidth]{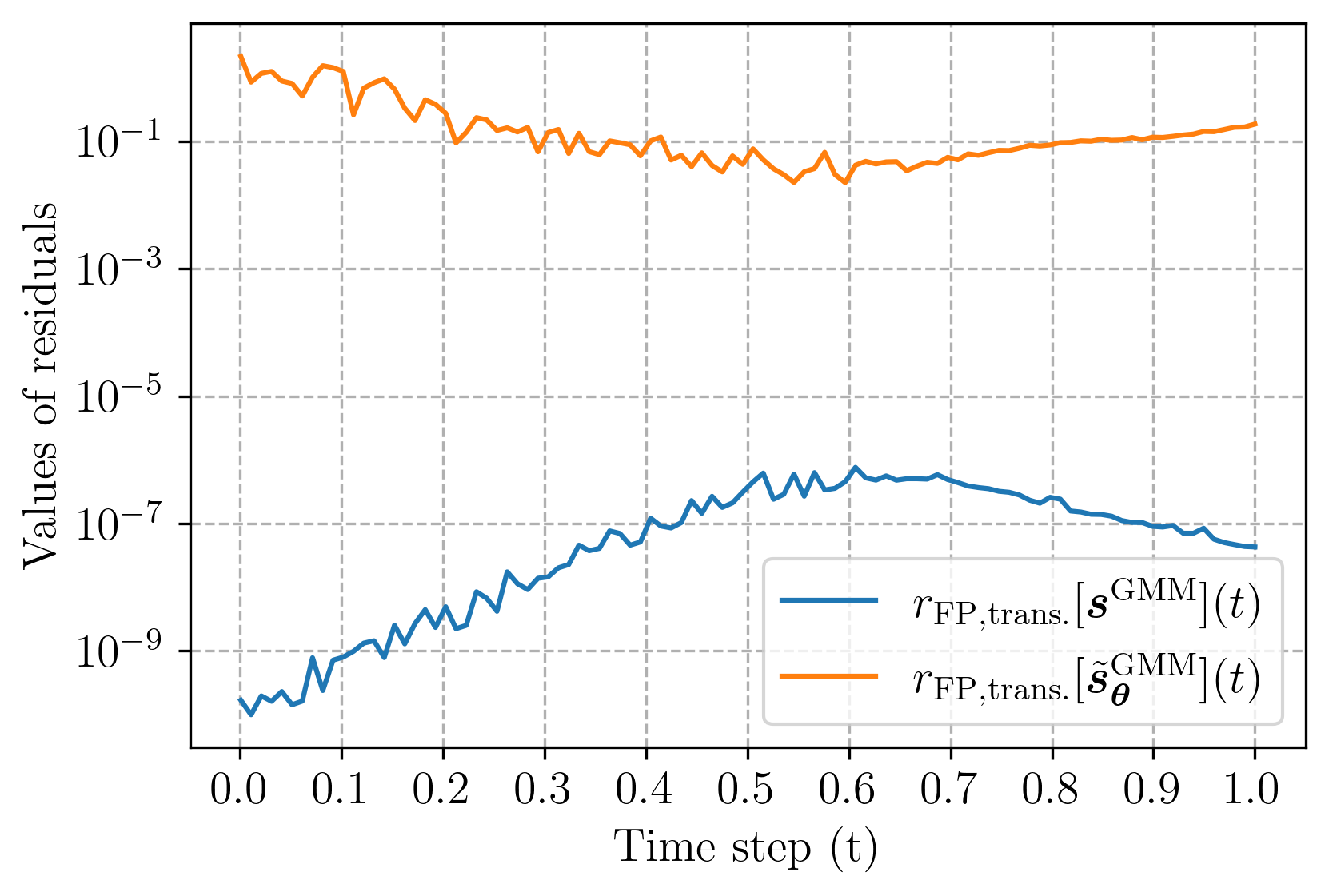}}
     \subfigure[FP residuals of the score learned from Eq.~\eqref{eq:dsm}]{\label{fig:gmm_fp_dsm}\includegraphics[width=0.43\textwidth]{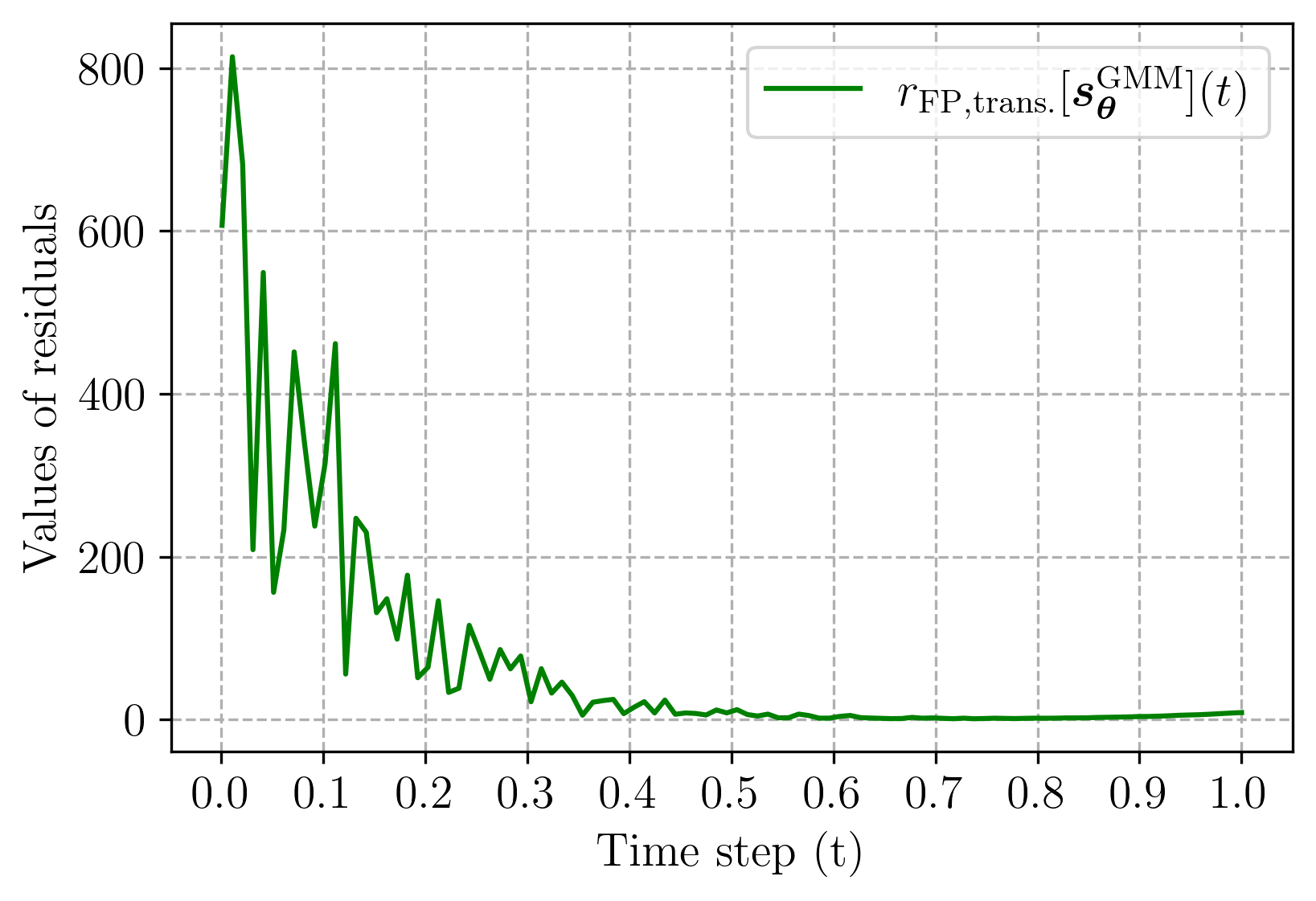}}
        \caption{Comparison of the score FPE residuals of $\bm{s}^{\textup{GMM}}$, $\tilde{\bm{s}}_{\bm{\theta}}^{\text{GMM}}$ and  $\bm{s}_{\bm{\theta}}^{\text{GMM}}$ for a 2D GMM. (a) shows that both the (closed-form) ground truth score $\bm{s}^{\textup{GMM}}$ and the score $\tilde{\bm{s}}_{\bm{\theta}}^{\text{GMM}}$ obtained by solving the score FPE (Eq.~\eqref{eq:solv_fp}) numerically satisfy the score FPE. On the other hand, (b) provides further evidence that $\bm{s}_{\bm{\theta}}^{\text{GMM}}$, which is learned from DSM, does not satisfy the score FPE.}
        \label{fig:gmm_fp}
\end{figure}

We examine whether $\bm{s}^{\textup{GMM}}$ satisfies the score FPE by computing $r_{\text{FP}}[\bm{s}^{\textup{GMM}}](t)$. Figure~\ref{fig:gmm_fp_gt} shows its residual as a function of time (blue curve) and supplements with the time residual of $\bm{\tilde{s}_{\theta}^{\textup{GMM}}}$, obtained by solving Eq.~\eqref{eq:fp_guided}. The score FPE residual of the ground truth is almost zero, which empirically supports Proposition~\ref{th:fp}. 

In addition, Figure~\ref{fig:gmm_fp_dsm} shows the computed residual of the score FPE as a function of time for a score $\bm{s}_{\theta}^{\textup{GMM}}$ learned by DSM (Eq.~\eqref{eq:dsm}). We observed that $\bm{s}_{\theta}^{\textup{GMM}}$ also does not satisfy the score FPE. This phenomenon matches with the results shown in Figure~\ref{fig:large_fp} for realistic datsets.

\subsection{FP-Diffusion enforces satisfaction of score FPE}\label{subsec:fp-diff-enforce}
\begin{figure}[th]
     \centering     \includegraphics[width=0.43\textwidth]{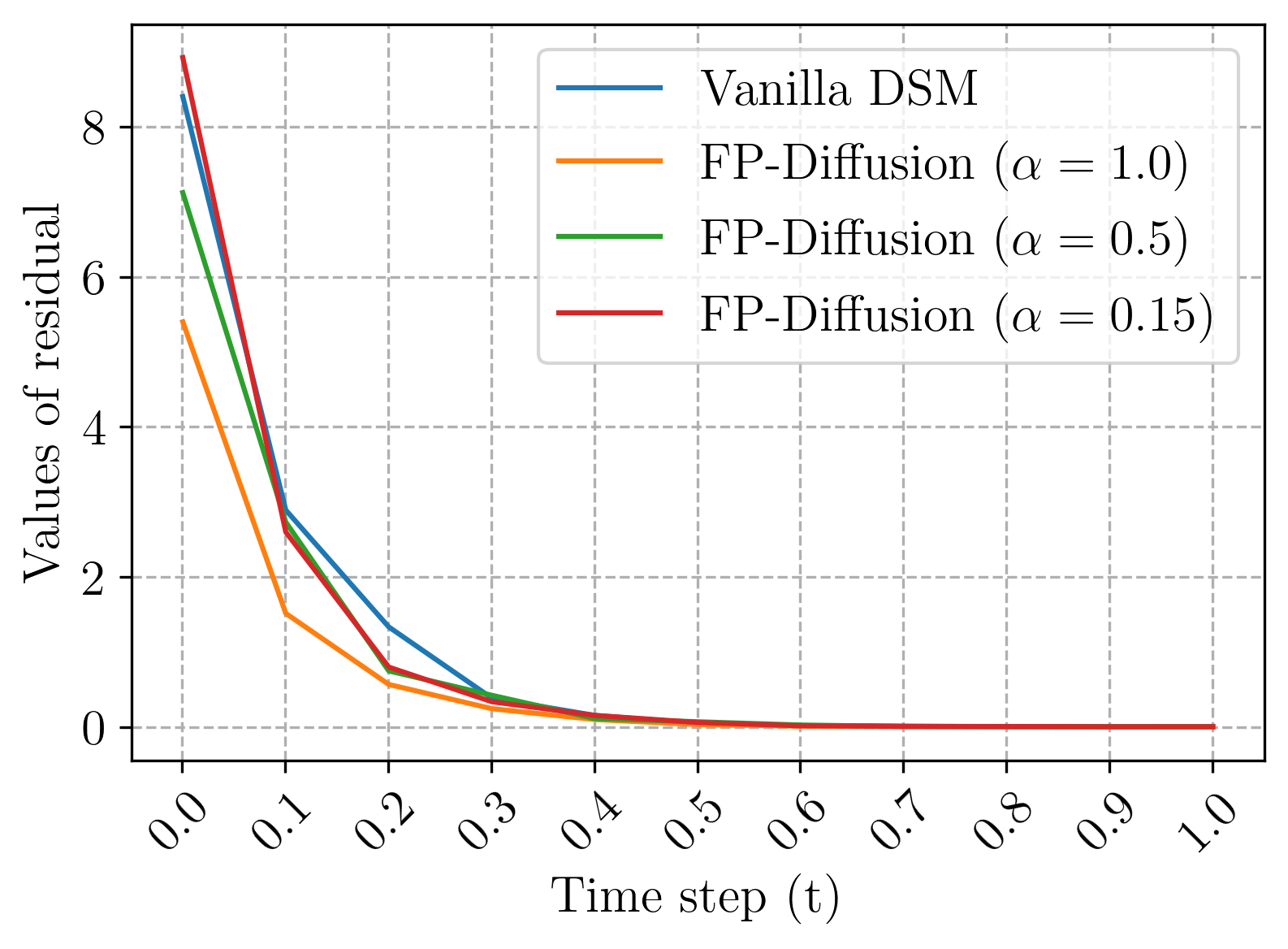}
        \caption{Comparison of the score FPE residuals as a function of time of vanilla DSM model and FP-Diffusion trained with $\alpha=1.0$, $0.5$, and $0.15$, respectively.}
        \label{fig:fpdiff_fp}
\end{figure}

In this section, we compared the score FPE residuals as a function of time (i.e., $r_{\text{FP, trans.}}[\bm{s}_{\bm{\theta}}](t)$) between a vanilla DSM model (trained for $0.1$M additional iterations) and FP-Diffusion respectively trained on CIFAR-10 with various $\alpha \in \{1.0, 0.5, 0.15\}$ and a fixed $(\beta, \lambda_{\text{FP}}(\cdot), m) = (0.01, g^2(\cdot), 2)$. The forward SDE was taken as the VE type. For demonstration purpose, we only plotted the residuals at representative timesteps $\{10^{-5}, 0.1, 0.2,\cdots,1.0  \}$ in Figure~\ref{fig:fpdiff_fp} and recorded their corresponding numerical values in Table~\ref{tb:res_numerical}. We remark that $\alpha=0.15$ achieves the best NLL reported in Table~\ref{tb:nll_cifar_imagenet}. Thanks to the score FPE-regularizer, FP-Diffusion generally obtains smaller residuals compared with the vanilla model (i.e., scores satisfy the score FPE better).

\begin{table}[th]
\caption{Values of score FPE residuals as a function of time of vanilla DSM model and FP-Diffusion trained with $\alpha=1.0$, $0.5$, and $0.15$, respectively. For each timestep, we mark the largest value in bold.}
\centering
\resizebox{0.9\columnwidth}{!}{
\begin{tabular}{lcccccccccccc}
\hline
\textbf{Model/Time}                                          & $10^{-5}$ & $0.1$ & $0.2$ & $0.3$ & $0.4$ & $0.5$ & $0.6$ & $0.7$ & $0.8$ & $0.9$ & $1.0$              
\\ \hline
Vanilla~\citep{song2020score}   & 8.401 & \textbf{2.889} & \textbf{1.331} & 0.383 & \textbf{0.152}& \textbf{0.067} & \textbf{0.023} & \textbf{0.008} & \textbf{0.005} & \textbf{0.002} &  \textbf{0.003}             \\ \hline
FP-Diffusion ($\alpha=1.0$)  & 5.399 & 1.512 & 0.565 & 0.242 & 0.097 & 0.030 & 0.009 & 0.003 & 0.001 & 0.000 & 0.001            \\ \hline    
FP-Diffusion ($\alpha=0.5$)  & 7.121 & 2.728 & 0.745 & \textbf{0.421} & 0.110 & 0.046 & 0.015 &0.005 &0.002 & 0.001& 0.000            \\ \hline    
FP-Diffusion ($\alpha=0.15$)  & \textbf{8.922} & 2.597 & 0.796 & 0.335 & 0.132 & 0.058 & 0.014 & 0.006 & 0.004 & 0.001&  0.001           \\ \hline    
\end{tabular}
}
\label{tb:res_numerical}
\end{table}

\section{More details on techniques for efficient score FPE computation }\label{sec:technique}
As explained in Section~\ref{sec:sfpe_reg}, the computation of $\bm{\epsilon}[{\bm{s_{\theta}}}](\bm{x}, t)$ in $\mathcal{R}_{\text{FP}}(\bm{\theta})$ is generally expensive; hence, we applied two techniques, the finite difference trick and Hutchinson's trace estimator, to replace the expensive computations of certain components in $\bm{\epsilon}[{\bm{s_{\theta}}}](\bm{x}, t)$.

\subsection{Trick to reduce computation cost of \texorpdfstring{$\partial_t\bm{s}_{\bm{\theta}}$}{TEXT}}\label{sec:reduce1}

Typically, $\partial_t\bm{s}_{\bm{\theta}}$ can be computed via automatic differentiation. However, it can be efficiently approximated by finite differences as the derivative is one-dimensional. We review the one-dimensional finite difference method and summarize its estimation error in the following lemma.
\begin{lemma}\label{th:finite_diff}\citep{fornberg1988generation}
Let $\alpha\colon[0,1] \rightarrow \mathbb{R}^D$ be a vector-valued function that is continuously differentiable up to third order derivatives. Let $h_s$ and $h_d$ be step-size hyper-parameters. Then, we have the following estimate of $\alpha'(t)$:
\begin{equation*}
    \frac{h_s^2 \alpha(t+h_d) +(h_d^2-h_s^2)\alpha(t)-h_d^2\alpha(t-h_s)}{h_s h_d (h_s + h_d)} + \mathcal{O}\Big(\frac{h_d h_s^2 + h_s h_d^2}{h_s + h_d} \Big).
\end{equation*}
In particular, if $h_s = h_d=:h$, then the estimate becomes
\begin{equation*}
    \frac{\alpha(t+h) -\alpha(t-h)}{2h} + \mathcal{O}(h^2).
\end{equation*}
\end{lemma}

In implementation for a high-dimensional dataset, we consider $\alpha(\cdot):= \bm{s}_{\bm{\theta}}(\bm{x},\cdot)$; hence, $\partial_t\bm{s}_{\bm{\theta}}(\bm{x}, t)$ is approximated as 
\begin{equation*}
    \frac{h_s^2 \bm{s}_{\bm{\theta}}(\bm{x}, t+h_d)  +(h_d^2-h_s^2)\bm{s}_{\bm{\theta}}(\bm{x}, t) -h_d^2\bm{s}_{\bm{\theta}}(\bm{x},t-h_s)}{h_s h_d (h_s + h_d)},
\end{equation*}
where we set $(h_s, h_d) = (0.001, 0.0005)$.

\subsection{Trick to reduce computation cost of  \texorpdfstring{$\div{\bm{x}} (\bm{s}_{\bm{\theta}})$}{TEXT}}\label{sec:reduce2}
Hutchinson's trace estimator~\citep{hutchinson1989stochastic}
stochastically estimates the trace $\textup{tr}(\bm{A})$ of any square matrix $\bm{A}$. The idea is to choose a distribution $p_{\bm{v}}$  so that $\mathbb{E}_{\bm{v}\sim p_{\bm{v}}}[\bm{v}] = \bm{0}$ and $\mathbb{E}_{\bm{v}\sim p_{\bm{v}}}[\bm{v}\bm{v}^T] = \bm{I}$. Hence,  $\textup{tr}(\bm{A}) = \textup{tr}(\bm{A}\mathbb{E}_{\bm{v}\sim p_{\bm{v}}}[\bm{v}\bm{v}^T]) = \mathbb{E}_{\bm{v}\sim p_{\bm{v}}}[\textup{tr}(\bm{A}\bm{v}\bm{v}^T)] = \mathbb{E}_{\bm{v}\sim p_{\bm{v}}}[\textup{tr}(\bm{v}\bm{A}\bm{v}^T)] = \mathbb{E}_{\bm{v}\sim p_{\bm{v}}}[\bm{v}\bm{A}\bm{v}^T]$. By i.i.d. sampling $\{\bm{v}_j\}_{j=1}^{M}$ from $p_{\bm{v}}$, we can use an unbiased estimator 
$$ \frac{1}{M}\sum_{j=1}^{M}\bm{v}_j\bm{A}\bm{v}_j^T$$
to estimate $\textup{tr}(\bm{A})$. Note that $\div{\bm{x}} (\bm{s}_{\bm{\theta}}(\bm{x}, t)) = \textup{tr}\big(\grad{\bm{x}} \bm{s}_{\bm{\theta}}\big)$. Thus, we can apply Hutchinson's trick and replace the $\div{\bm{x}} (\bm{s}_{\bm{\theta}})$ term with the following estimation: 
$$ \frac{1}{M}\sum_{j=1}^{M}\bm{v}_j\grad{\bm{x}} \bm{s}_{\bm{\theta}}(\bm{x}, t)\bm{v}_j^T.$$ In implementation, $p_{\bm{v}}$ is usually taken as a standard normal distribution or a Rademacher distribution.

We set $M=1$ in our implementation.

\subsection{Error analysis of score FPE estimation}\label{sec:reduce-error-bound}

In this section, we provide a theoretical error analysis of score FPE by using finite difference approximation and Hutchinson’s estimation.  

Here we consider the case of taking the Rademacher distribution for Hutchinson's estimation. The proof is differed to Appendix~\ref{subsec:proof-error}.
As for the case of normal distribution, we may utilize a statistical bound in \citep{roosta2015improved} and obtain a similar statistical estimate by following the identical argument.

\begin{proposition}\label{th:error_analysis} Let $\bm{s}(\bm{x}, t)$ be a vector field on $\mathbb{R}^D \times [t_0, T]$, where $t_0>0$ and $g$ be a continuous function defined on $[t_0, T]$ which achieves minimum at $t^*\in[t_0, T]$ and that $g^*:=g^2 (t^*)>0$.
Denote 
\begin{equation*}
    \mathscr{V}[\bm{s}]:=\partial_t\bm{s} -\frac{1}{2}g^2\textup{tr}(\nabla\bm{s})
\end{equation*}
and 
\begin{equation*}
    \hat{\mathscr{V}}[\bm{s}]:=\textup{FD}(\bm{s}) -\frac{1}{2}g^2\textup{tr}_{H^{(M)}}(\nabla\bm{s}).
\end{equation*} Here we denote a finite difference approximation in $t$ with parameters $h_s, h_d>0$ as $\textup{FD}(\bm{s})$, and a Hutchinson's estimator with $M$ samples from Rademacher distribution as $\textup{tr}_{H^{(M)}}(\nabla\bm{s})$ (see Section~\ref{sec:sfpe_reg} for their explicit definitions). Let $\mathscr{T}[\bm{s}]:=\partial_t\bm{s}-\textup{FD}(\bm{s})$, $\mathscr{H}[\bm{s}]:=\frac{1}{2}g^2 \big(\textup{tr}(\nabla\bm{s})-\textup{tr}_{H^{(M)}}(\nabla\bm{s})\big)$ and $\mathscr{E}[\bm{s}]:=\mathscr{V}[\bm{s}]-\hat{\mathscr{V}}[\bm{s}]$.
Notice from Lemma~\ref{th:finite_diff} that there is a constant $C>0$ so that $\norm{\mathscr{F}[\bm{s}]}_D < CD h$, where $h=\frac{h_s^2 h_d + h_d h_s^2}{h_s + h_d}$ and $\norm{\cdot}_D$ indicates the $\ell_D$-norm. For any $\epsilon \in (0, \frac{3}{8}g^*)$, if $h\in(0, \frac{\epsilon}{2CD})$, then 
    \begin{equation*}
        \mathbb{P}\big(\norm{\mathscr{E}[\bm{s}]}_D<\epsilon \big)\geq 1- \exp^{-\Big(\frac{M \epsilon^2}{2(g^* - \frac{8}{3}\epsilon)}  \Big)}.
    \end{equation*}
\end{proposition}

\subsection{Potential technique to compute \texorpdfstring{$\bm{\epsilon}[{\bm{s_{\theta}}}]$}{TEXT} more efficiently}\label{sec:reduce3}
In this section, we propose another potential trick to reduce the computation cost of differentiation. Recall that 
\begin{equation}\label{eq:residue_I_II}
    \bm{\epsilon}[{\bm{s_{\theta}}}](\bm{x}, t)= \underbrace{\partial_t \bm{s}_{\bm{\theta}}}_{\textup{(I)}} - \underbrace{\grad{\bm{x}}\Big[\frac{1}{2}g^2(t) \div{\bm{x}}(\bm{s}_{\bm{\theta}}) + \frac{1}{2}g^2(t)\norm{\bm{s}_{\bm{\theta}}}^2_2 -\inner{\bm{f}}{\bm{s}_{\bm{\theta}}} - \div{\bm{x}}(\bm{f})  \Big]}_{\textup{(II)}}
\end{equation}
The use of automatic differentiation to compute the gradient in $\bm{\epsilon}[{\bm{s_{\theta}}}](\bm{x}, t)$ (part (II) in Eq.~\eqref{eq:residue_I_II}) is generally cumbersome for high dimensional data. We thus propose to use random projection to replace the gradient computation (multi-dimensional) with a directional derivative (one-dimensional). Then, we can apply the finite difference trick introduced in above to further reduce the computation effort. We first recall a fundamental property before rigorously formulating the technique.

\begin{lemma}\label{th:directional} Let $M:=M(\bm{x},t)\colon\mathbb{R}^D\times[0,T]\rightarrow\mathbb{R}$ be a continuously differentiable function of $\bm{x}$. For any $\bm{v}\in\mathbb{R}^D$, 
\begin{equation*}
    D_{\bm{v}}M(\bm{x},t) = \inner{\grad{\bm{x}}M(\bm{x},t)}{\bm{v}},  
\end{equation*}
where $D_{\bm{v}}M(\bm{x},t)$ denotes the directional derivative of $M$ in $\bm{x}$ along the direction $\bm{v}$ and is defined as follows:
\begin{equation*}
    D_{\bm{v}}M(\bm{x},t):= \lim_{h\rightarrow0}\frac{M(\bm{x}+h\bm{v},t) - M(\bm{x},t)}{h} = \frac{d}{dh}M(\bm{x}+h\bm{v},t)\Big\vert_{h=0}.
\end{equation*}
\end{lemma}

For simplicity, we let $M(\bm{x},t):=\frac{1}{2}g^2(t) \div{\bm{x}}(\bm{s}_{\bm{\theta}}) + \frac{1}{2}g^2(t)\norm{\bm{s}_{\bm{\theta}}}^2_2 -\inner{\bm{f}}{\bm{s}_{\bm{\theta}}} - \div{\bm{x}}(\bm{f})$ and let $\bm{v}\in\mathbb{R}^D$ be an arbitrary vector. We project $\bm{\epsilon}[{\bm{s_{\theta}}}](\bm{x}, t)$ along direction $\bm{v}$ and apply Lemma~\ref{th:directional}:
\begin{align*}
    \inner{\bm{\epsilon}[{\bm{s_{\theta}}}](\bm{x}, t)}{\bm{v}}
    =\inner{\partial_t \bm{s}_{\bm{\theta}}-\grad{\bm{x}}M(\bm{x},t)}{\bm{v}} =\inner{\partial_t \bm{s}_{\bm{\theta}}}{\bm{v}}- \inner{\frac{d}{dh}M(\bm{x}+h\bm{v},t)\Big\vert_{h=0}}{\bm{v}}.
\end{align*}
Note that both $\partial_t \bm{s}_{\bm{\theta}}$ and $\frac{d}{dh}M(\bm{x}+h\bm{v},t)\big\vert_{h=0}$ entail one-dimensional differentiation and can be estimated via Lemma~\ref{th:finite_diff}, thus avoiding automatic differentiation. That is, we may have the estimation 
\begin{equation}\label{eq:proj_v}
    \bm{\epsilon}[{\bm{s_{\theta}}}](\bm{x}, t) \approx \mathbb{E}_{\bm{v}\sim p_{\bm{v}}} \inner{\bm{\epsilon}[{\bm{s_{\theta}}}](\bm{x}, t)}{\bm{v}},
\end{equation}
where $p_{\bm{v}}$ is the distribution of a random vector $\bm{v}\in\mathbb{R}^D$. However, the performance may be degraded by using the estimate in Eq.~\eqref{eq:proj_v} possibly because of the inaccurate approximation of the exact score FPE. Hence, we will need further study on lowering the computation costs while preventing performance degradation.

\section{Implementation details}\label{sec:implement}
In this section, we describe the details of our implementation on synthetic dataset, MNIST/Fashion MNIST, and CIFAR-10/ImageNet32. 

\subsection{Synthetic dataset}
\label{subsec:implement-simple}
We conducted our experiments on 4 NVIDIA GeForce RTX 3090 GPUs.

\textbf{2D GMM. }
For all experiments of the 2D GMM $\frac{1}{5}\mathcal{N}\big((-5, -5), \bm{I} \big) + \frac{4}{5} \mathcal{N}\big((5, 5), \bm{I} \big)$, we exploited a network structure similar to the one in a particular repository\footnote{ \url{https://colab.research.google.com/drive/120kYYBOVa1i0TD85RjlEkFjaWDxSFUx3?usp=sharing}} for all . In that repository, we modified the forward SDE modified to be a VE SDE or VP SDE (see Appendix~\ref{sec:instances}), but we simply replaced all convolutional layers with fully connected layers. We trained for $2,000$ epochs with a learning rate of $10^{-3}$ and a batch size of $500$.  

\textbf{Checkerboard, Swiss rolls, eight-mode 2D GMM, and 1D GMM.}
The neural network setups for the results shown in Figures~\ref{fig:comparison_ckb_sw} and \ref{fig:comparison_fp_dsm}  were the same as the toy model structures provided in the repository of \citet{lu2022maximum} \footnote{\url{https://github.com/LuChengTHU/mle_score_ode}}. We show our detailed data preparation below, as modified from the same repository.
We trained the models for 0.1M iterations with a learning rate of $10^{-3}$ and a batch size of $500$. For both training and inference, the start time was $10^{-3}$.

\begin{lstlisting}[language=Python, caption=Checkerboard dataset]
import numpy
import torch
x1 = np.random.rand(batch_size) * 4 - 2
x2_ = np.random.rand(batch_size) - np.random.randint(0, 2, batch_size) * 2
x2 = x2_ + (np.floor(x1) % 2)
checkerboard = torch.from_numpy(np.concatenate([x1[:, None], x2[:, None]], 1).float() * 2
\end{lstlisting}

\begin{lstlisting}[language=Python, caption=Swiss rolls dataset]
import numpy
import torch
import sklearn
data = sklearn.datasets.make_swiss_roll(n_samples=batch_size, noise=1.0)[0]
data = data.astype("float32")[:, [0, 2]]
data /= 4. 
data = torch.from_numpy(data).float()
r = 4.5
data1 = data.clone() + torch.tensor([-r, -r])
data2 = data.clone() + torch.tensor([-r, r])
data3 = data.clone() + torch.tensor([r, -r])
data4 = data.clone() + torch.tensor([r, r])
swiss_roll = torch.cat([data, data1, data2, data3, data4], axis=0)
\end{lstlisting}

\begin{lstlisting}[language=Python, caption=8 modes 2D GMM dataset]
import numpy
import torch
num_mixture = 8 
radius = 1.0
sigma = 0.1
mix_probs = [1/num_mixture] * num_mixture
std = torch.stack([torch.ones(dim) * sigma for i in range(len(mix_probs))], dim=0)
mix_probs = torch.tensor(mix_probs)
mix_idx = torch.multinomial(mix_probs, n, replacement=True)        
thetas = np.linspace(0, 2 * np.pi, num_mixture, endpoint=False)
xs = radius * np.sin(thetas, dtype=np.float32)
ys = radius * np.cos(thetas, dtype=np.float32)
center = np.vstack([xs, ys]).T
center = torch.tensor(centers)
centers = centers[mix_idx] 
stds = std[mix_idx]
eight_GMM = torch.randn_like(centers) * stds + centers  
\end{lstlisting}

\subsection{MNIST and Fashion MNIST}\label{subsec:implement-mnist-fmnist}
We conducted our experiments on 4 NVIDIA GeForce RTX 3090 GPUs. 

We trained score networks on MNIST and Fashion MNIST from scratch for 200 epochs with a learning rate of $10^{-3}$ and batch size of $32$ by using the setup as in the repository\footnote{ \url{https://colab.research.google.com/drive/120kYYBOVa1i0TD85RjlEkFjaWDxSFUx3?usp=sharing}}, with the forward SDE modified to be a VE SDE or VP SDE. For both training and inference, the start time was $10^{-3}$.

\subsection{CIFAR-10 and ImageNet32}
\label{subsec:implement-complex}

For CIFAR-10 and ImageNet32, we followed the same model architectures and experimental setups as in \citet{song2020score}\footnote{\url{https://github.com/yang-song/score_sde_pytorch}} and \citet{lu2022maximum}\footnote{\url{https://github.com/LuChengTHU/mle_score_ode/}}, respectively. More precisely, we used NCSN++ cont. for the VE and NCSN++ cont. deep for the VE-deep. We conducted our experiments on 4 NVIDIA A100 GPUs (40 GiB). The batch size was fixed as 48. Instead of training from scratch, we ued the pre-trained VE models provided by the two repositories and fine tuned them by training for  0.1M additional iterations.  As we have found that a smaller batch size may decrease the NLL of the probability flow ODE, for a fair comparison, we also trained the vanilla DSM models for 0.1M additional iterations. Table~\ref{tb:nll_cifar_imagenet} reports the results.  

We used uniform dequantization~\citep{ho2019flow++} for likelihood evaluation. To reduce the variance, we computed the NLL (in bpd) over five repeated runs and took their average. For both training and inference, we chose a start time of $10^{-5}$.

\section{Supplemental results}\label{sec:sensitivity}

\subsection{Sensitivity to hyper-parameters}\label{subsec:sensitivity}

\textbf{Fine-tuning pre-trained models on MNIST and Fashion MNIST. } In Section~\ref{subsec:exp_mnist}, we trained models from scratch on MNIST and Fashion MNIST, and compared vanilla DSM models with FP-Diffusion with various $\alpha$'s. However, different initializations and optimization dynamics may lead to variances of NLLs. To avoid this issue, we followed our training strategy for CIFAR-10 and ImageNet32 by fine tuning pre-trained models instead. More precisely, we first trained vanilla DSM models for 100 epochs and fine tuned with different $\alpha$'s for additional 100 epochs. Table~\ref{tb:nll_mnist_fmnist_finetune} shows the NLL comparisons, where we can observe general improvements with FP-Diffusion.

\begin{table}[th]
\caption{NLL comparisons on MNIST and Fashion MNIST (trained by fine-tuning)}
\centering
\resizebox{0.53\columnwidth}{!}{
\begin{tabular}{lccccc}
\hline
 & \multicolumn{2}{c}{\textbf{MNIST} } & \multicolumn{1}{c}{ }
 &\multicolumn{2}{c}{\textbf{Fashion MNIST} } \\ \cline{2-3} \cline{5-6}
\textbf{Method}  & VE     & VP        &  & VE & VP               \\ \hline
Vanilla~\citep{song2020score}   &  3.67 & 3.27   &  &  4.78 & 4.49  \\
FP-Diffusion ($\alpha=0.001$)    & 3.63  &  3.15  &  &  4.61 & 4.50 
\\
FP-Diffusion ($\alpha=0.01$)    & 3.60  &  3.10  &  &  4.58 & 4.46    
\\
FP-Diffusion ($\alpha=0.1$)   &  3.47 & \textbf{3.01}    &  & 4.42  & \textbf{4.23}  
\\
FP-Diffusion ($\alpha=1.0$)    &  \textbf{3.23} & 3.09  &  & \textbf{4.39}  & 4.27 
\\
FP-Diffusion ($\alpha=10.0$) & 3.30  & 3.17    &  &  4.41 & 4.43  
\\
 \hline
\end{tabular}
}
\label{tb:nll_mnist_fmnist_finetune}
\end{table}

\textbf{Fine-tuning pre-trained models on CIFAR-10. }
We compared the proposed FP-Diffusion with the VE SDE trained on CIFAR-10 with different hyper-parameter choices. Table~\ref{tb:sensitivity} reports the test set NLL results, which were computed by averaging over five repeated runs to reduce variances. We found that $(\alpha, \beta, \lambda_{\text{FP}}(\cdot), m) = (0.15, 0.01, g^2(\cdot), 2) $ generally works well on more complicated datasets such as CIFAR-10 and ImageNet32. We remark that the choice of $\beta$ makes the scale of $\abs{\mathcal{L}[\bm{s}_{\bm{\theta}}](\bm{x}, t)}$ in Eq.~\eqref{eq:DSM+FP} be comparable with $\mathcal{J}_{\text{DSM}}$, where $\beta\approx10^{-2}$ is generally a reasonable value for both CIFAR-10 and ImageNet32. Additionally, we motivate the choice of $\lambda_{\text{FP}}(\cdot)$ as $g^2(\cdot)$ in Appendix~\ref{subsec:disc-th-min-ode}.

\textbf{General recipe for searching hyper-parameters. }
A general recipe for the selection of hyper-parameters is (1) to set $\alpha$ 
 be at the scale of $10^{-1}$; (2) may need $\beta$
 (roughly at the scale of $10^{-2}$) for more complicated datasets.  However. hyper-parameters to obtain the best results may depend on the optimization and the structure of datasets as the landscapes of loss functions are totally different. We generally observed that a larger $\alpha$ 
 (roughly at the scale of $10^0$) is preferable for relatively sparse datasets such as MNIST and Fashion MNIST. In contrast, a smaller $\alpha$ 
 (roughly at the scale of $10^{-1}$) and $\beta$
 (roughly at the scale of $10^{-2}$) is preferable for more complicated datasets such as CIFAR-10 and ImageNet32. As for synthetic datasets, $(\alpha,\beta)\approx(10^{-2}, 0.0)$ works well. Nevertheless, FP-Diffusion overall improves NLLs against the vanilla training.

\begin{table}[th]
\caption{FP-Diffusion with different of hyper-parameter choices for the VE SDE trained on CIFAR-10. }
\centering
\resizebox{0.45\columnwidth}{!}{
\begin{tabular}{lc}
\hline
$(\alpha, \beta, \lambda_{\text{FP}}(\cdot), m) $ & \textbf{NLL (bpd) on CIFAR-10} \\ \hline
Vanilla DSM $\alpha=\beta=0.0$                       & 3.61                  \\ \hline

$(0.15, 0.01, g^2(\cdot), 2) $                    & \textbf{3.36}         \\ \hline
$(1.0, 0.01, g^2(\cdot), 2) $                     & 3.40                  \\
$(0.5, 0.01, g^2(\cdot), 2) $                     & 3.38                  \\
$(0.2, 0.01, g^2(\cdot), 2) $                     & 3.37                  \\
$(0.1, 0.01, g^2(\cdot), 2) $                     & 3.37                  \\
$(0.0, 0.01, *, *) $                                 & 3.37                  \\ \hline
$(1.0, 0.0, g^2(\cdot), 2) $                     & 3.57    
         \\ \hline
\end{tabular}
}
\label{tb:sensitivity}
\end{table}

\subsection{Runtime discussion}\label{subsec:runtime}
In this section, we compared the runtime of vanilla diffusion model (trained with $\mathcal{J}_{\textup{DSM}}$) and the proposed FP-Diffusion on CIFAR-10. We fixed the training batch size as 48 and examined their runtime on VE type model NCSN++ cont. with PyTorch.  The hardware was 4 NVIDIA A100 GPUs (40 GiB).  We believe the computation time of FP-Diffusion can be improved with a more optimized code and setup of the environment.
\begin{table}[th]
\caption{Runtime comparison of vanilla diffusion and FP-Diffusion trained on CIFAR-10. The forward SDE was taken as the VE type.  }
\centering
\resizebox{0.7\columnwidth}{!}{
\begin{tabular}{lccc}
\hline
\textbf{Method}                                                                  & \textbf{Time per iteration (sec)}               &  & \textbf{Memory (GiB) }                 \\ \hline
Vanilla~\citep{song2020score}                                                                          & 0.17                  &  & 23.48                  \\ \hline
FP-Diffusion                                                                     & \multirow{2}{*}{2.08} &  & \multirow{2}{*}{49.01} \\
$(\alpha, \beta, \lambda_{\text{FP}}(\cdot), m) = (0.15, 0.01, g^2(\cdot), 2)$ &                       &  &                       \\ \hline
\end{tabular}
}
\label{tb:runtime}
\end{table}

\subsection{Illustration and quality of generated samples}\label{subsec:ill-samples}
\begin{figure}[!th]
\centering
\begin{minipage}{.24\textwidth}
  \centering
  \includegraphics[width=\textwidth]{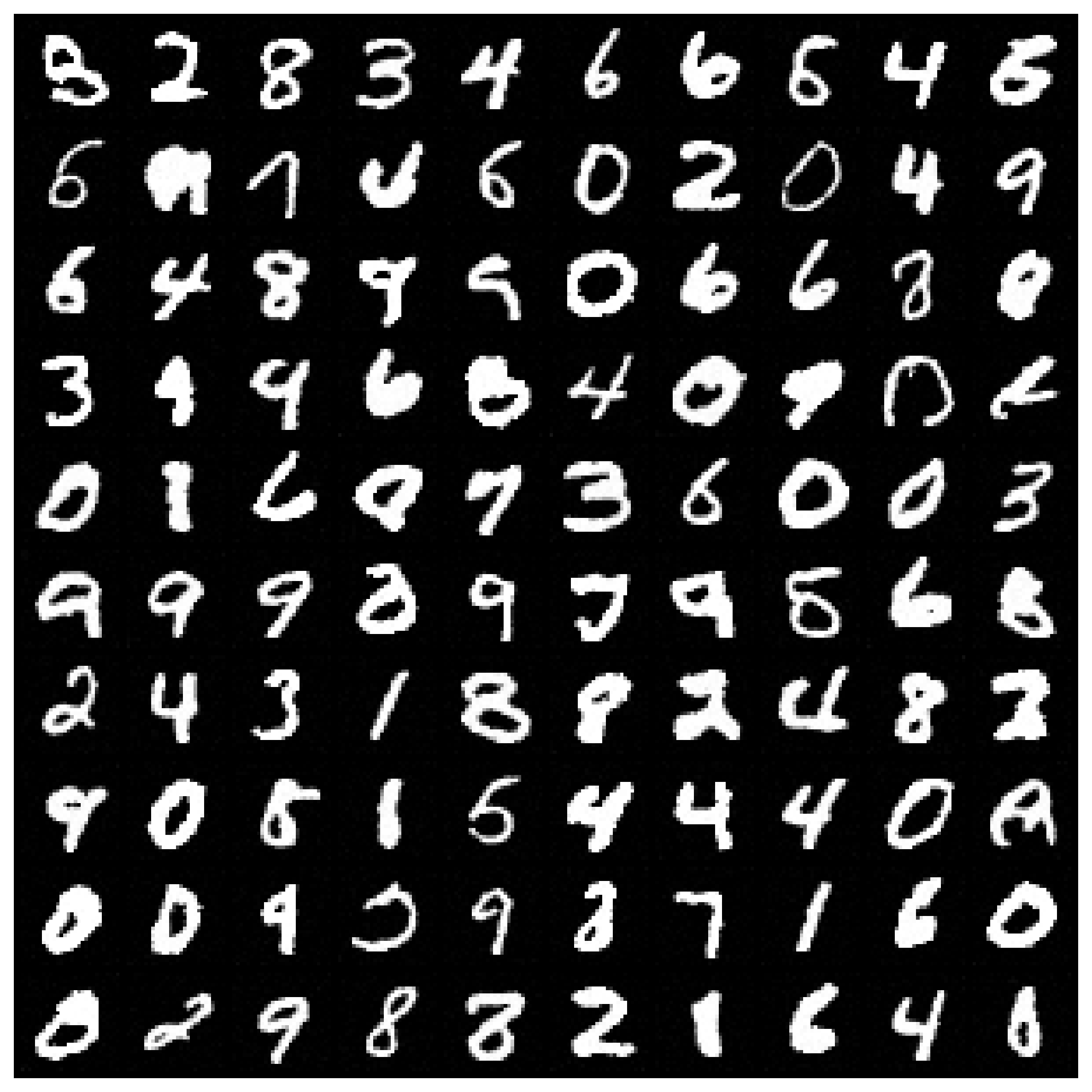}
  \\ (a) VE ($\alpha=\beta=0.0$)
\end{minipage}%
\begin{minipage}{.24\textwidth}
  \centering
  \includegraphics[width=\textwidth]{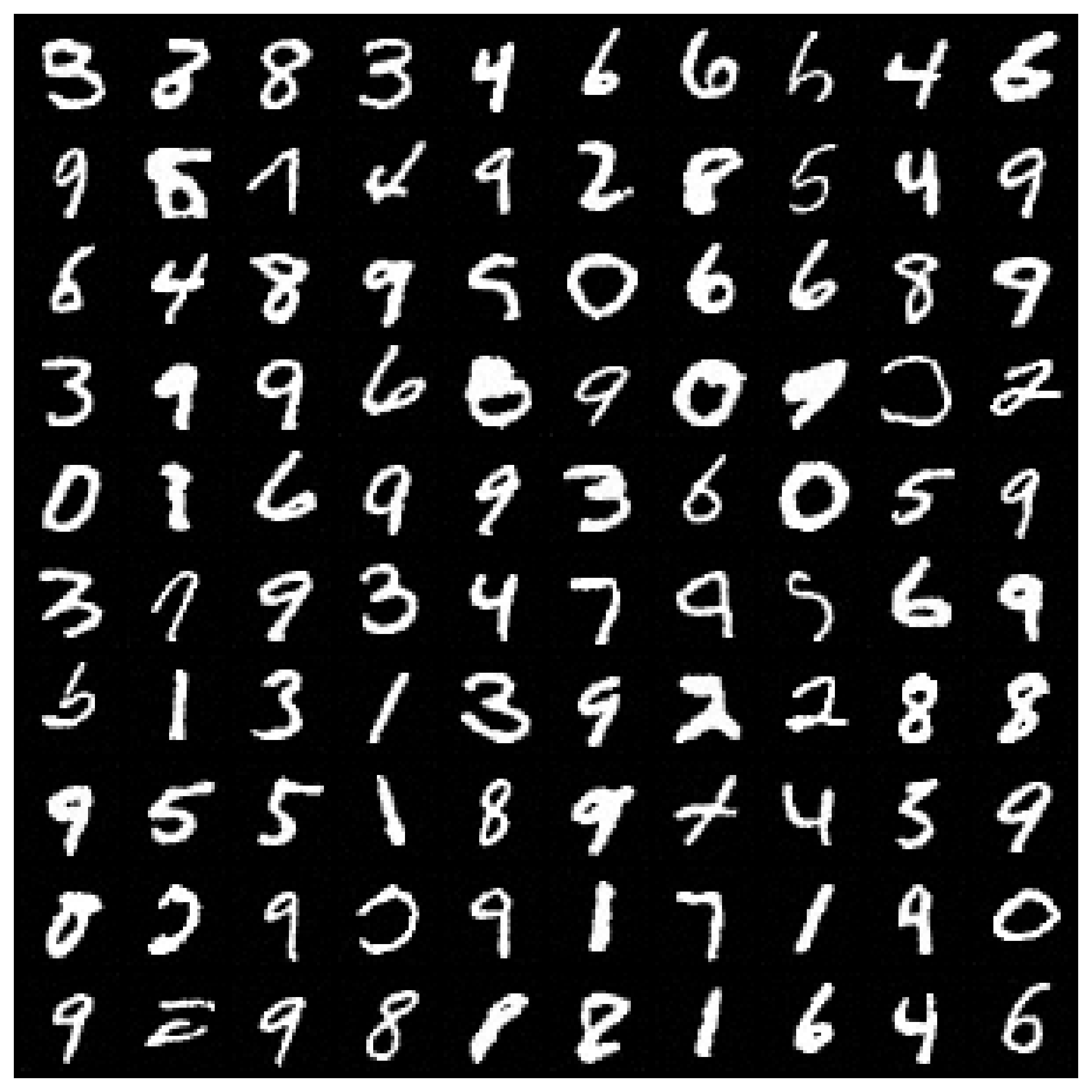}
  \\ (b) VE ($\alpha=1.0$) 
\end{minipage}
\begin{minipage}{.24\textwidth}
  \centering
  \includegraphics[width=\textwidth]{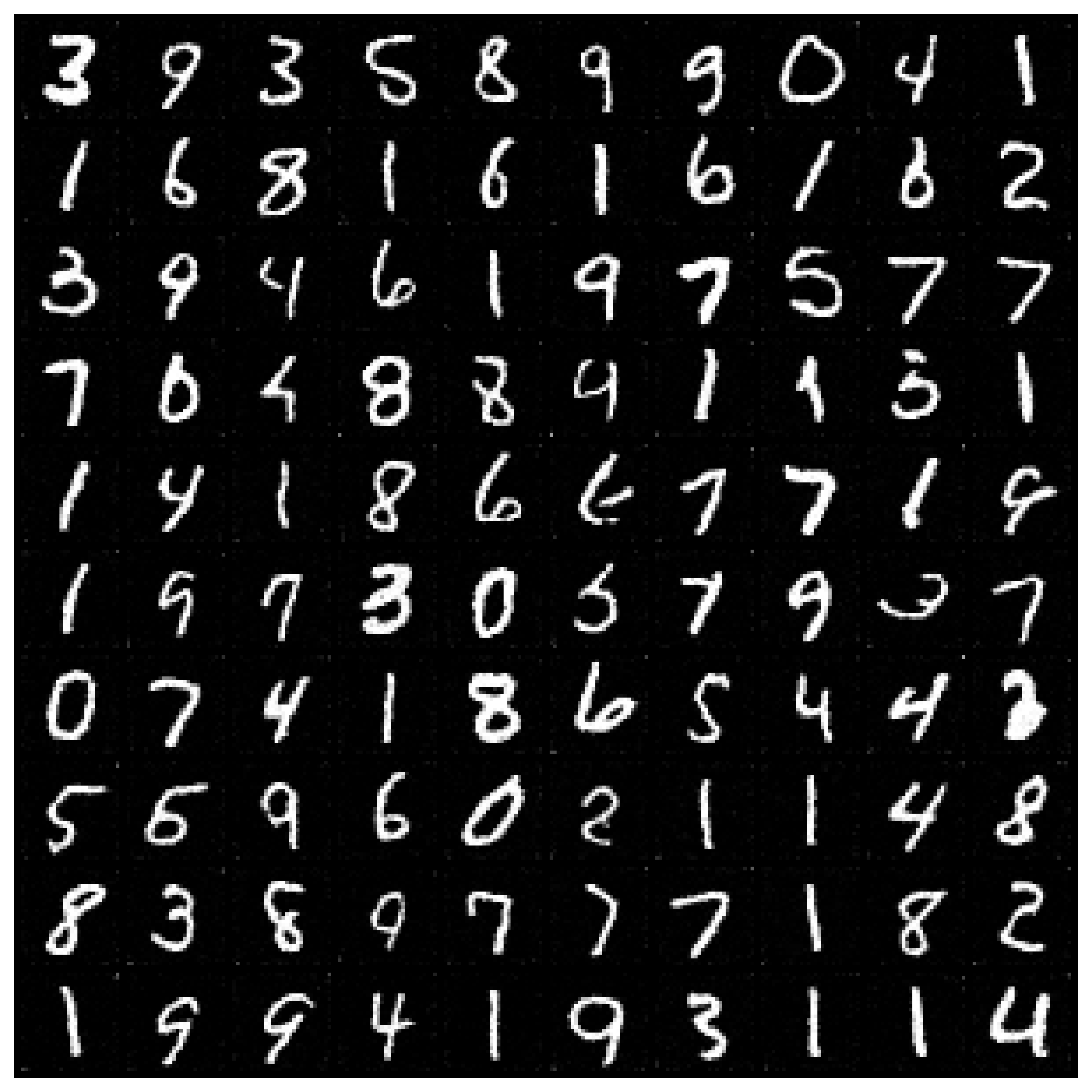}
  \\ (c) VP ($\alpha=\beta=0.0$)
\end{minipage}%
\begin{minipage}{.24\textwidth}
  \centering
  \includegraphics[width=\textwidth]{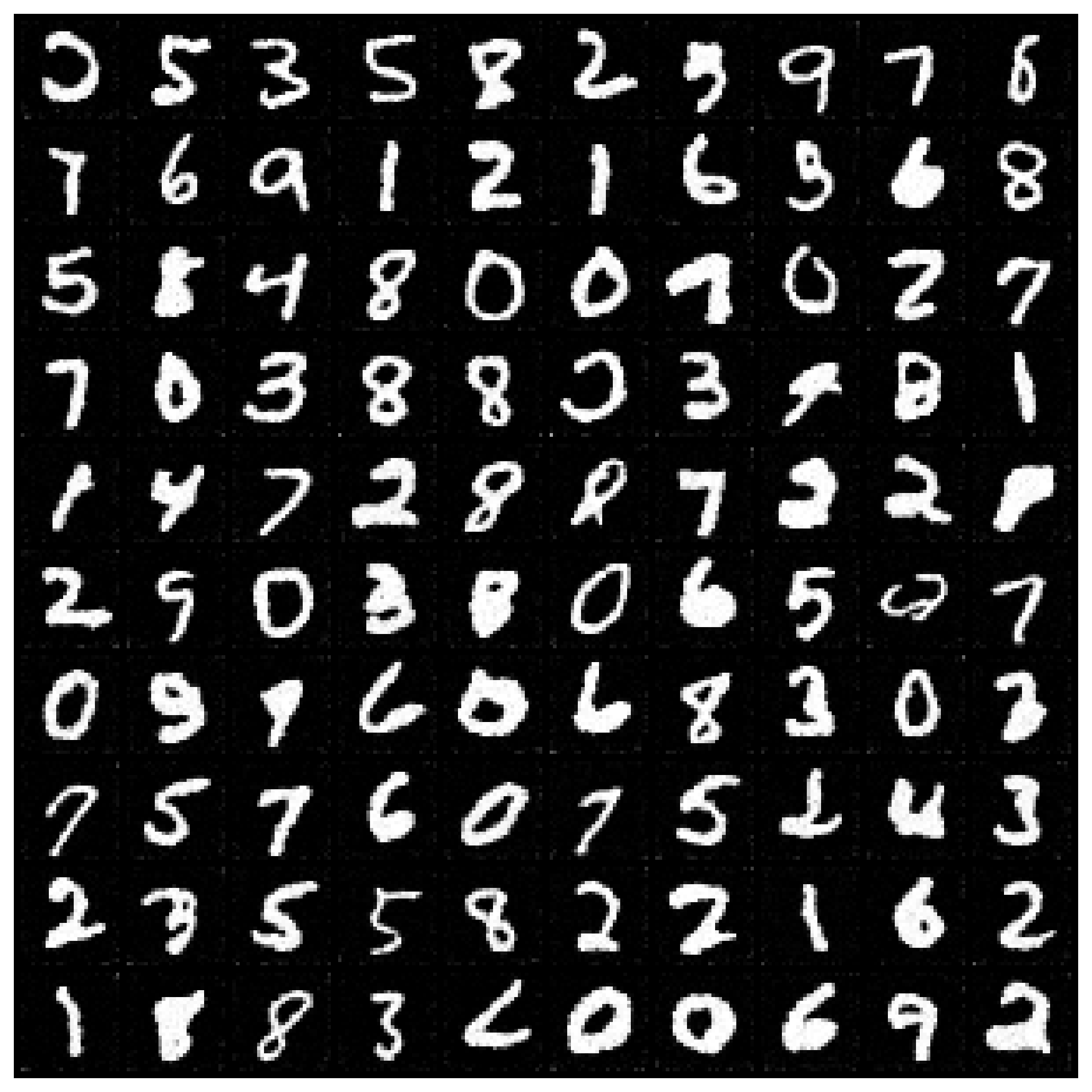}
  \\ (d) VP ($\alpha=1.0$) 
\end{minipage}
 \caption{Samples generated with models trained on MNIST by using (a, c) vanilla DSM and (b, d) FP-Diffusion with the  setup described in Section~\ref{subsec:exp_mnist}.}
 \label{fig:mnist}
\end{figure}

\begin{figure}[!th]
\centering
\begin{minipage}{.24\textwidth}
  \centering
  \includegraphics[width=\textwidth]{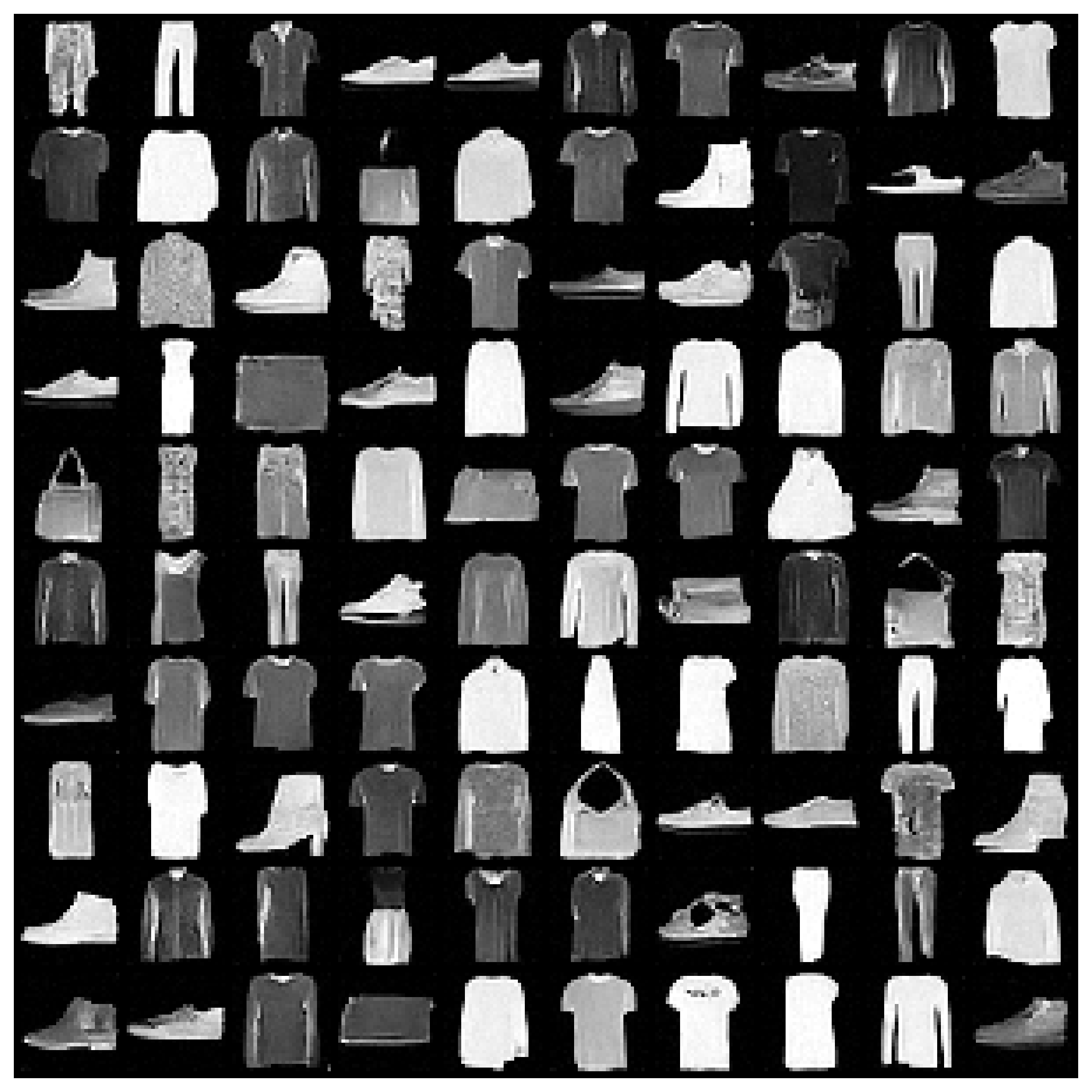}
  \\ (a) VE ($\alpha=\beta=0.0$)
\end{minipage}%
\begin{minipage}{.24\textwidth}
  \centering
  \includegraphics[width=\textwidth]{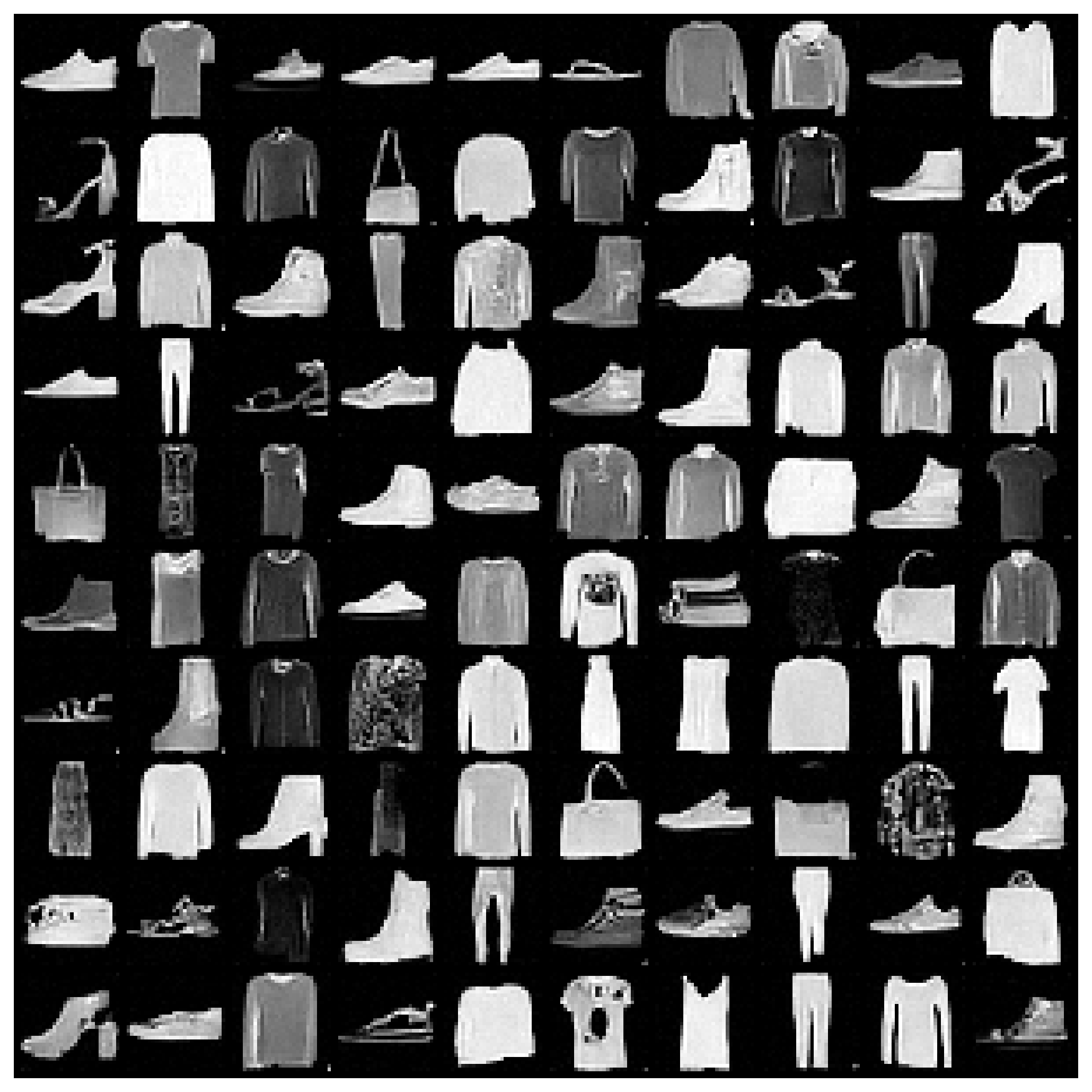}
  \\ (b) VE ($\alpha=1.0$) 
\end{minipage}
\begin{minipage}{.24\textwidth}
  \centering
  \includegraphics[width=\textwidth]{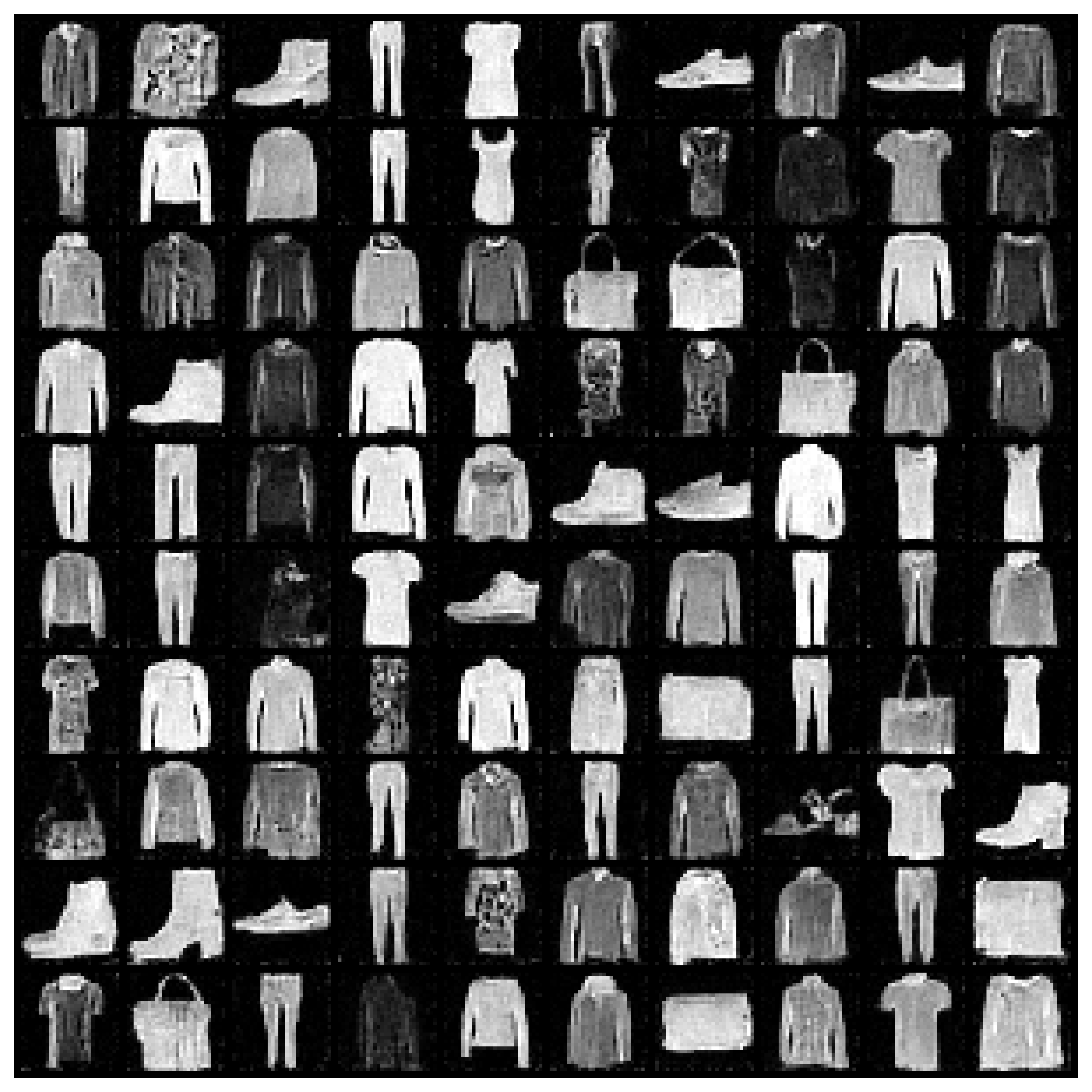}
  \\ (c) VP ($\alpha=\beta=0.0$)
\end{minipage}%
\begin{minipage}{.24\textwidth}
  \centering
  \includegraphics[width=\textwidth]{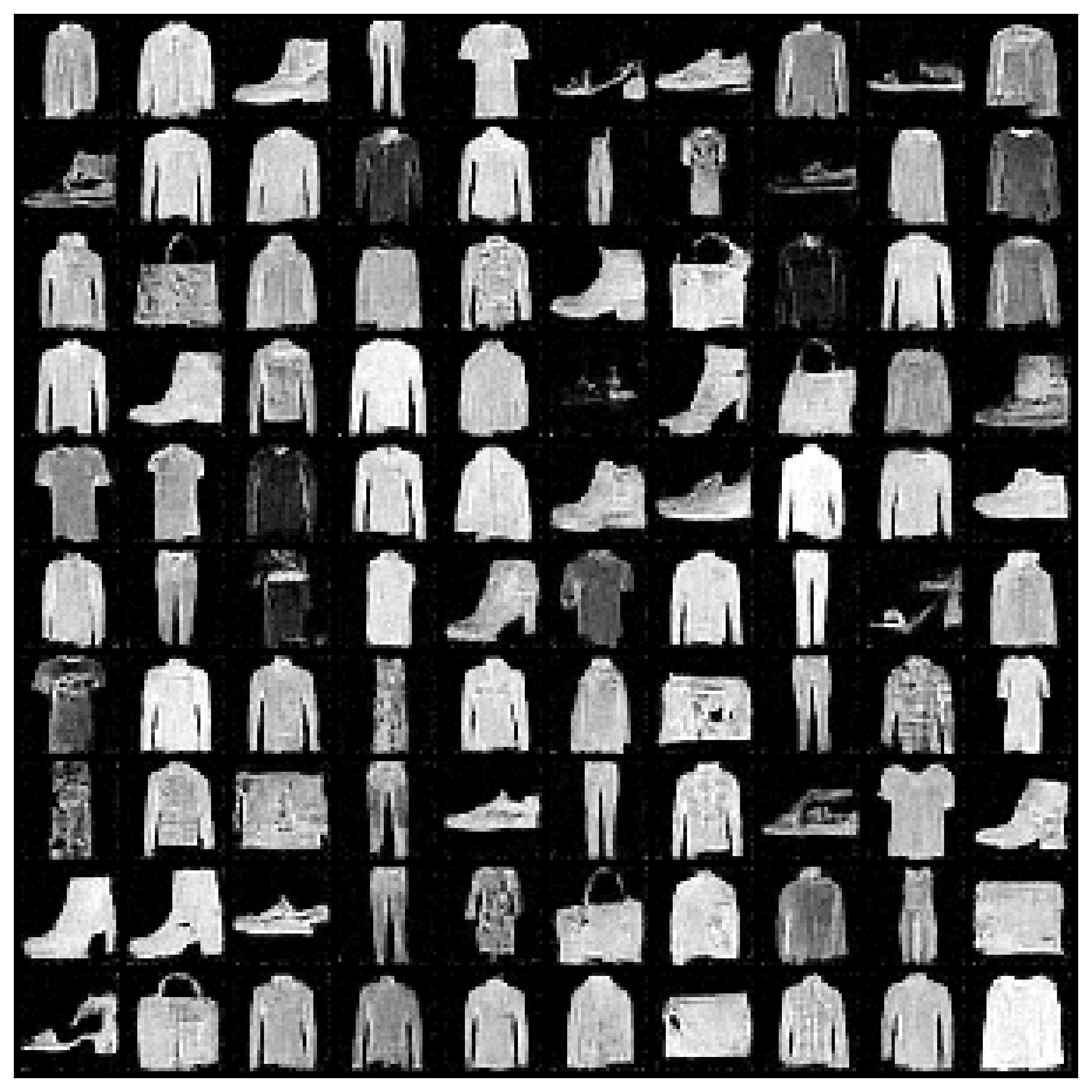}
  \\ (d) VP ($\alpha=1.0$) 
\end{minipage}
 \caption{Samples generated with models trained on Fashion MNIST by using (a, c) vanilla DSM and (b, d) FP-Diffusion with the  setup described in Section~\ref{subsec:exp_mnist}.}
 \label{fig:fmnist}
\end{figure}

\begin{figure}[!th]
\centering
\begin{minipage}{.24\textwidth}
  \centering
  \includegraphics[width=\textwidth]{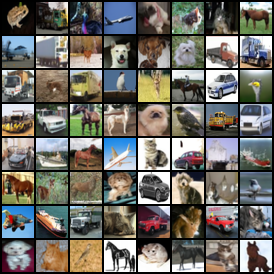}%
  \\ (a) VE (vanilla $\alpha=\beta=0.0$) on CIFAR-10
\end{minipage}%
\vspace{0.1cm}
\begin{minipage}{.24\textwidth}
  \centering
  \includegraphics[width=\textwidth]{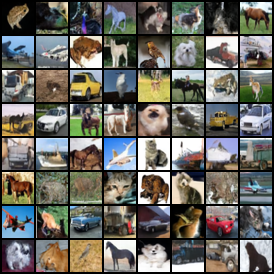}%
  \\ (b) VE (FP-Diffusion) on CIFAR-10
\end{minipage}%
\vspace{0.1cm}
\begin{minipage}{.24\textwidth}
  \centering
  \includegraphics[width=\textwidth]{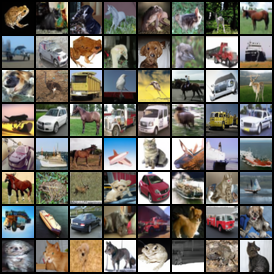}%
  \\ (c) VE-deep (vanilla  $\alpha=\beta=0.0$) on CIFAR-10
\end{minipage}%
\vspace{0.1cm}
\begin{minipage}{.24\textwidth}
  \centering
  \includegraphics[width=\textwidth]{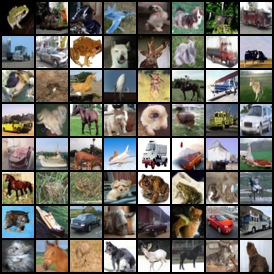}%
  \\ (d) VE-deep (FP-Diffusion) on CIFAR-10 
\end{minipage}
 \caption{Illustration of generated samples with VE and VE-deep models trained on CIFAR-10. (a) and (c) show samples generated by vanilla DSM. (b) and (d) show samples generated by FP-Diffusion with the setup described in Section~\ref{subsec:exp_cifar}.}
 \label{fig:cifar_imagenet}
\end{figure}

We visualized randomly generated samples with models trained on MNIST, Fashion MNIST, and CIFAR-10 in Figures~\ref{fig:mnist}, \ref{fig:fmnist}, and \ref{fig:cifar_imagenet}, respectively. In Table~\ref{tb:cifar_sample}, we reported numerical results of sample quality with models trained on CIFAR-10. Even though FP-Diffusion has inferior numerical measurements compared with vanilla models, the differences are imperceptible by comparing their generated samples.

\begin{table}[th]
\caption{Sample quality  on CIFAR-10. }
\centering
\resizebox{0.56\columnwidth}{!}{
\begin{tabular}{lccccc}
\hline
 & \multicolumn{2}{c}{\textbf{FID} $\downarrow$} & \multicolumn{1}{c}{ }
 &\multicolumn{2}{c}{\textbf{IS} $\uparrow$} \\ \cline{2-3} \cline{5-6}
\textbf{Method}  & VE     & VE-deep    &       & VE  & VE-deep           \\ \hline
Vanilla~\citep{song2020score}          &  \textbf{3.33}   &  \textbf{2.44}  &  &  \textbf{9.19}    & \textbf{9.80}
\\ \hline
FP-Diffusion             &  10.83   &   3.33  &  &  8.88   &   9.14     

\\
 \hline
\end{tabular}
}
\label{tb:cifar_sample}
\end{table}

\section{Theoretical assumptions}\label{sec:assumptions}
Here, we introduce some regularity conditions to establish Theorems~.\ref{th:min_ode} and \ref{th:min_ode_add} which are commonly used in theoretical studies of score-based models~\citep{song2021maximum,lu2022maximum,pidstrigach2022score,kwon2022score}.

\begin{assumption}\label{cond:A} We assume there are finite constants $L>0$, which is sufficiently large (may assume $L\geq 1$), and $\delta_T>0$ such that the following conditions hold for all $\bm{x}, \bm{y}\in\mathbb{R}^D$ and $t\in[0,T]$
\begin{enumerate}[(a)]
    \item\label{cond:expect_x} Bounded $2^\text{nd}$ non-central moment: $\mathbb{E}_{q_{0}(\bm{x})}[\norm{\bm{x}}_2^2] \leq L$, or $\sup_{t\in[0,T]}\big\{\mathbb{E}_{\bm{x} \sim q_{t}(\bm{x})}[\norm{\bm{x}}_2^2] \big\} \leq L$ to streamline the proof,
    \item\label{cond:s_linear} $\norm{\bm{s}_{\bm{\theta}}(\bm{x}, t)}_2 \leq L(1+ \norm{\bm{x}}_2)$,
    \item\label{cond:s_lip}  $\norm{\bm{s}_{\bm{\theta}}(\bm{x}, t) - \bm{s}_{\bm{\theta}}(\bm{y}, t)}_2 \leq L \norm{\bm{x}-\bm{y}}_2$,
    \item\label{cond:f_linear} $\norm{\bm{f}(\bm{x}, t)}_2 \leq L(1+ \norm{\bm{x}}_2)$,    
    \item\label{cond:f_lip}    
    $\norm{\bm{f}(\bm{x}, t) - \bm{f}(\bm{y}, t)}_2 \leq L \norm{\bm{x}-\bm{y}}_2$,
    \item\label{cond:ode_linear} $\norm{\bm{s}_{\bm{\theta}}^{\text{ODE}}(\bm{x}, t)}_2 \leq L(1+ \norm{\bm{x}}_2)$,    
    \item\label{cond:ode_lip}
    $\norm{\bm{s}_{\bm{\theta}}^{\textup{ODE}}(\bm{x}, t) - \bm{s}_{\bm{\theta}}^{\textup{ODE}}(\bm{y}, t)}_2 \leq L \norm{\bm{x}-\bm{y}}_2$,
    \item
    $\norm{\bm{s}(\bm{x}, t)}_2 \leq L(1+ \norm{\bm{x}}_2)$,
    \item
    $\norm{\bm{s}(\bm{x}, t) - \bm{s}(\bm{y}, t)}_2 \leq L \norm{\bm{x}-\bm{y}}_2$,
\end{enumerate}
 and that
\begin{enumerate} 
    \item[(j)]\label{cond:match_T} $\sup_{t\in[0,T]}\Big\{\mathbb{E}_{q_{t}(\bm{x})}\big[\norm{\bm{s}_{\bm{\theta}}(\bm{x}, T) - \bm{s}_{\bm{\theta}}^{\textup{ODE}}(\bm{x}, T)}_2^2\big]\Big\} \leq \delta_T^2$, or
$\sup_{\bm{x}\in\mathbb{R}^D}\norm{\bm{s}_{\bm{\theta}}(\bm{x}, T) - \bm{s}_{\bm{\theta}}^{\textup{ODE}}(\bm{x}, T)}_2^2\leq \delta_T^2$,
    \item[(k)]\label{cond:decay} For any $t\in[0,T]$, there is a $k>0$ so that as $\norm{x}_2 \to \infty$, $q_t(\bm{x})=\mathcal{O}\big(e^{-\norm{x}_2^k}\big)$ and $p^{\text{ODE}}_t(\bm{x})=\mathcal{O}\big(e^{-\norm{x}_2^k}\big)$. 
\end{enumerate}

\end{assumption}

\begin{assumption}\label{cond_add:A} We assume there is a finite constant $L>0$ such that for all $\bm{x}, \bm{y}\in\mathbb{R}^D$ and $t\in[0,T]$ the following conditions hold
\begin{enumerate}
    \item[(c')]\label{cond:grad_s_lip}
    $\norm{\nabla_{\bm{x}}\bm{s}_{\bm{\theta}}(\bm{x}, t) - \nabla_{\bm{x}}\bm{s}_{\bm{\theta}}(\bm{y}, t)}_2 \leq L \norm{\bm{x}-\bm{y}}_2$,  
    \item[(e')]\label{cond:grad_f_lip}   
    $\norm{\nabla_{\bm{x}}\bm{f}(\bm{x}, t) - \nabla_{\bm{x}}\bm{f}(\bm{y}, t)}_2 \leq L \norm{\bm{x}-\bm{y}}_2$,
    \item[(g')]\label{cond:grad_ode_lip} $\norm{\nabla_{\bm{x}}\bm{s}_{\bm{\theta}}^{\textup{ODE}}(\bm{x}, t) - \nabla_{\bm{x}}\bm{s}_{\bm{\theta}}^{\textup{ODE}}(\bm{y}, t)}_2 \leq L \norm{\bm{x}-\bm{y}}_2$.
\end{enumerate}
\end{assumption}

\section{Proofs and discussions}\label{sec:proofs}


\subsection{Proof of Proposition~\ref{th:fp}}\label{subsec:pf-prop3.1}
\begin{proof}

We prove the result with a more general forward SDE
\begin{equation}\label{eq:general_sde}
    d\bm{x} = \bm{F}(\bm{x}, t) dt + \bm{G}(\bm{x}, t) d\bm{w}_t,
\end{equation}
where $\bm{F}(\cdot,t)\colon\mathbb{R}^D\rightarrow\mathbb{R}^D$ and $\bm{G}(\cdot,t)\colon\mathbb{R}^D\rightarrow\mathbb{R}^{D\times D}$.

We know that the density $q_t(\bm{x})$ satisfies the Fokker-Planck equation~\citep{oksendal2003stochastic}
\begin{equation}\label{eq:fp_general}
    \partial_t q_t (\bm{x}) = -\sum_{j=1}^{D} \partial_{x_j}\big(\tilde{\bm{F}}_j(\bm{x}, t) q_t (\bm{x})\big),
\end{equation}
where $\tilde{\bm{F}}(\bm{x}, t):= \bm{F}(\bm{x}, t) - \frac{1}{2}\nabla \cdot [\bm{G}(\bm{x}, t)\bm{G}(\bm{x}, t)^T]-\frac{1}{2}\bm{G}(\bm{x}, t)\bm{G}(\bm{x}, t)^T\nabla_{\bm{x}} \log q_t(\bm{x})$. We further denote $\bm{A}(\bm{x},t):=\bm{F}(\bm{x}, t) - \frac{1}{2}\nabla \cdot [\bm{G}(\bm{x}, t)\bm{G}(\bm{x}, t)^T]$ and $\bm{B}(\bm{x},t):= -\frac{1}{2}\bm{G}(\bm{x}, t)\bm{G}(\bm{x}, t)^T$.

Now $\tilde{\bm{F}}(\bm{x}, t) = \bm{A}(\bm{x},t) + \bm{B}(\bm{x},t)\bm{s}(\bm{x}, t)$, and we have
\begin{align} 
    \partial_t \log q_t(\bm{x}) \nonumber
    &= \frac{1}{q_t(\bm{x})} \partial_t q_t(\bm{x}) \nonumber \\
    &= - \frac{1}{q_t(\bm{x})} \sum_{j=1}^{D} \partial_{x_j}\big(\tilde{\bm{F}}_j(\bm{x}, t) q_t (\bm{x})\big) \nonumber \\
    &= -\frac{1}{q_t(\bm{x})} \sum_{j=1}^{D} \big(\partial_{x_j} \tilde{\bm{F}}_j(\bm{x}, t) q_t (\bm{x})+\tilde{\bm{F}}_j(\bm{x}, t)  \partial_{x_j}q_t(\bm{x})  \big) \nonumber \\
    &= - \sum_{j=1}^{D}\big(\partial_{x_j} \tilde{\bm{F}}_j(\bm{x}, t) + \tilde{\bm{F}}_j(\bm{x}, t) \partial_{x_j}\log q_t(\bm{x})\big) \nonumber \\
    &= -\big(\div{\bm{x}}(\tilde{\bm{F}}) + \inner{\tilde{\bm{F}}}{\bm{s}}   \big)\nonumber \\
    &= -\Big[\div{\bm{x}}\big(\bm{B}\bm{s}\big) + \inner{\bm{B}\bm{s}}{\bm{s}}+\inner{\bm{A}}{\bm{s}}
     + \div{\bm{x}}(\bm{A}) \Big]\nonumber \\
    &= \frac{1}{2}\div{\bm{x}}\big(\bm{G}\bm{G}^T\bm{s}\big) + \frac{1}{2}\norm{\bm{G}^T\bm{s}}_2^2-\inner{\bm{A}}{\bm{s}}
    - \div{\bm{x}}(\bm{A}) \nonumber.
\end{align}
Since $\log q_t(\bm{x}) $ is sufficiently smooth, we can swap the order of differentiations and get $ \partial_t  \bm{s} = \partial_t  \grad{\bm{x}}\log q_t(\bm{x}) = \grad{\bm{x}} \partial_t \log q_t(\bm{x})$. Hence, the statement is proved. 

\end{proof}

\begin{remark} In Eq.~\eqref{eq:sde_forward} where $\bm{G}$ does not depend on $\bm{x}$, namely $\bm{G}(\bm{x}, t)\equiv g(t)\bm{I}$, then $\tilde{\bm{F}}(\bm{x}, t)= \bm{f}(\bm{x}, t) - \frac{1}{2}g^2(t)\nabla_{\bm{x}} \log q_t(\bm{x})$ and 
\begin{align*}
    \partial_t \log q_t(\bm{x})&= \frac{1}{2}g^2(t) \div{\bm{x}}(\bm{s}) + \frac{1}{2}g^2(t)\norm{\bm{s}}^2_2 -\inner{\bm{f}}{\bm{s}} - \div{\bm{x}}(\bm{f})  \\
    \partial_t \bm{s} &= \grad{\bm{x}}\Big[\frac{1}{2}g^2(t) \div{\bm{x}}(\bm{s}) + \frac{1}{2}g^2(t)\norm{\bm{s}}^2_2 -\inner{\bm{f}}{\bm{s}} - \div{\bm{x}}(\bm{f})  \Big]. 
\end{align*}

\end{remark}

\subsection{Proof of Theorem~\ref{th:min_ode}}\label{subsec:pf-thm4.2}

\begin{lemma}[Gr$\ddot{o}$nwall's inequality~\citep{gronwall1919note}]\label{th:gronwall} Assume that $\alpha$, $\beta$, and $u$ are continuous functions on $[0,T]$.  If $\beta$ is non-negative on $[0,T]$ and if $u$ satisfies the integral inequality
\begin{equation*}
    u(t) \leq \alpha(t) + \int_{t}^{T}\beta(\tau)u(\tau)d\tau,\qquad \text{for all } t\in[0,T]
\end{equation*}
then
\begin{equation*}
    u(t) \leq \alpha(t) + \int_{t}^{T}\alpha(\tau)\beta(\tau)\exp\big(\int_{t}^{\tau}\beta(r)dr \big)d\tau,\qquad \text{for all } t\in[0,T]
\end{equation*}
In particularly, if $\alpha$ is non-decreasing (especially, a constant independent of $t$), then
\begin{equation*}
    u(t) \leq \alpha(t) \exp\big(\int_{t}^{T}\beta(\tau)d\tau \big),\qquad \text{for all } t\in[0,T].
\end{equation*}
\end{lemma}
\begin{proof}{Gr$\ddot{o}$nwall's inequality}{}

Consider the function
\begin{equation*}
    v(\tau):=\exp\Big(-\int_{\tau}^{T} \beta(r) dr \Big)\int_{\tau}^{T} \beta(r)u(r) dr. 
\end{equation*}
Taking the derivative by the product rule leads to
\begin{align*}
    v'(\tau)&=\Big(-u(\tau) + \int_{\tau}^{T} \beta(r)u(r) dr \Big)\beta(\tau)\exp\Big(-\int_{\tau}^{T} \beta(r) dr \Big)
    \\&\geq -\alpha(\tau)\beta(\tau)\exp\Big(-\int_{\tau}^{T} \beta(r) dr \Big). 
\end{align*}
Integrating the above inequality from $\tau=t$ to $\tau=T$ proves the statement. 
\end{proof}


    
    

\begin{proof}{Theorem}~{\ref{th:min_ode}} 

We first prove the Ineq.~\eqref{eq:diff_sm_fisher_add}. Notice that we can rearrange $\mathcal{J}_{\text{Diff}}$ as
\begin{align*}
 \mathcal{J}_{\text{Diff}}(\bm{\theta}) &= \frac{1}{2}\displaystyle\int_{0}^{T} g^2(t) \mathbb{E}_{\bm{x} \sim q_{t}(\bm{x})}\Big[\big(\bm{s}_{\bm{\theta}}(\bm{x}, t) - \grad{\bm{x}} \log q_t(\bm{x})\big)^{\top} \big( \bm{s}^{\textup{ODE}}_{\bm{\theta}}(\bm{x}, t) -\bm{s}_{\bm{\theta}}(\bm{x}, t)\big) \Big] dt
 \\ &= \displaystyle\int_{0}^{T} \displaystyle\int_{\mathbb{R}^D }  \Big[g(t)\sqrt{\frac{q_t(\bm{x})}{2}} \big(\bm{s}_{\bm{\theta}}(\bm{x}, t) - \grad{\bm{x}} \log q_t(\bm{x})\big)\Big]^{\top} \Big[g(t)\sqrt{\frac{q_t(\bm{x})}{2}}\big( \bm{s}^{\textup{ODE}}_{\bm{\theta}}(\bm{x}, t) -\bm{s}_{\bm{\theta}}(\bm{x}, t)\big) \Big]dtd\bm{x}.
\end{align*}
The claim is established by applying Cauchy-Schwartz inequality to functions $g(t)\sqrt{\frac{q_t(\bm{x})}{2}} \big(\bm{s}_{\bm{\theta}}(\bm{x}, t) - \grad{\bm{x}} \log q_t(\bm{x})\big)$ and $g(t)\sqrt{\frac{q_t(\bm{x})}{2}}\big( \bm{s}^{\textup{ODE}}_{\bm{\theta}}(\bm{x}, t) -\bm{s}_{\bm{\theta}}(\bm{x}, t)\big)$.

Now, we prove the Ineq.~\eqref{eq:diff_sm_M}, in which we just need to consider the case when $M({\bm{\theta}}):=\sup_{t\in[0,T]}\mathbb{E}_{\bm{x} \sim q_{t}(\bm{x})}\big[\int_{0}^{T}\norm{\bm{\epsilon}[{\bm{s_{\theta}}}](\bm{x}, \tau)}_2 d\tau\big]<\infty$; otherwise, the result holds obviously. 
Throughout the proof, we simply use the notation $\lesssim$ to express $\lesssim_{T, \delta_T, g, L}$, which indicates the estimation depends only on $T, \delta_T, g, L$.

Recall that the probability flow ODE~\citep{song2020score} associated to Eq.~\eqref{eq:sde_backward_sub} is defined as  
\begin{equation*}
    \frac{d\bm{x}}{dt}(t) = \bm{f}(\bm{x}(t), t) -\frac{1}{2}g^2(t) \bm{s}_{\bm{\theta}}(\bm{x}(t), t).
\end{equation*}
By the special case of FPE (Eq.~\eqref{eq:fp_general}) with zero diffusion term, we obtain the PDE characterizes the evolution of $p^{\textup{ODE}}_{t,\bm{\theta}}$
\begin{equation*}
    \frac{\partial p^{\textup{ODE}}_{t,\bm{\theta}}}{\partial t} (\bm{x}, t)=  \div{\bm{x}}\Big( \big(\frac{1}{2}g^2 (t)\bm{s}_{\bm{\theta}}(\bm{x}, t)-\bm{f}(\bm{x}, t)  \big)   p^{\textup{ODE}}_{t,\bm{\theta}}(\bm{x}) \Big) 
\end{equation*}
Hence, 
\begin{align*}
    \frac{\partial \log p^{\textup{ODE}}_{t,\bm{\theta}}}{\partial t} &= \frac{1}{p^{\textup{ODE}}_{t,\bm{\theta}}} \frac{\partial p^{\textup{ODE}}_{t,\bm{\theta}}}{\partial t}
    \\&=\frac{1}{2}g^2 (t)\div{\bm{x}}(\bm{s}_{\bm{\theta}}) - \div{\bm{x}}(\bm{f})+\inner{\bm{s}^{\textup{ODE}}_{\bm{\theta}}}{\frac{1}{2}g^2 (t)\bm{s}_{\bm{\theta}}-\bm{f}},
\end{align*}
where we apply the product rule of divergence in the last equality. After taking the gradient from the both sides, we obtain\footnote{Indeed, Eq.~\eqref{eq:s_ode} can also be derived from Proposition~\ref{th:fp}.}
\begin{equation}\label{eq:s_ode}
\begin{aligned}
    \frac{\partial\bm{s}^{\textup{ODE}}_{\bm{\theta}}}{\partial t}(\bm{x}, t)
    &= \grad{\bm{x}} \frac{\partial \log p^{\textup{ODE}}_{t,\bm{\theta}}}{\partial t}
    \\ & = \grad{\bm{x}}\Big[\frac{1}{2}g^2 (t)\div{\bm{x}}(\bm{s}_{\bm{\theta}}) - \div{\bm{x}}(\bm{f}) \Big] + \grad{\bm{x}}\Big[\inner{\bm{s}^{\textup{ODE}}_{\bm{\theta}}}{\frac{1}{2}g^2 (t)\bm{s}_{\bm{\theta}}-\bm{f}}  \Big]
    \\ & = \grad{\bm{x}}\Big[\frac{1}{2}g^2 (t)\div{\bm{x}}(\bm{s}_{\bm{\theta}}) - \div{\bm{x}}(\bm{f}) \Big] + \grad{\bm{x}}\Big[\frac{1}{2}g^2 (t)\inner{\bm{s}^{\textup{ODE}}_{\bm{\theta}}}{\bm{s}_{\bm{\theta}}}-\inner{\bm{f}}{\bm{s}^{\textup{ODE}}_{\bm{\theta}}}  \Big]
\end{aligned}
\end{equation}
By rearranging Eq.~\eqref{eq:fp_error} and combining with Eq.~\eqref{eq:s_ode}, it results in 
\begin{align*}
    \bm{\epsilon}[{\bm{s_{\theta}}}](\bm{x}, t)
    &=\partial_t \bm{s}_{\bm{\theta}} - \grad{\bm{x}}\Big[\frac{1}{2}g^2(t) \div{\bm{x}}(\bm{s}_{\bm{\theta}}) - \div{\bm{x}}(\bm{f})  \Big] - \grad{\bm{x}}\Big[ \frac{1}{2}g^2(t)\norm{\bm{s}_{\bm{\theta}}}^2_2 -\inner{\bm{f}}{\bm{s}_{\bm{\theta}}}   \Big]
    \\&= \partial_t \bm{s}_{\bm{\theta}} - \partial_t \bm{s}^{\textup{ODE}}_{\bm{\theta}}
    - \grad{\bm{x}}\Big[ \frac{1}{2}g^2 (t)\inner{\bm{s}_{\bm{\theta}} - \bm{s}^{\textup{ODE}}_{\bm{\theta}}}{\bm{s}_{\bm{\theta}}}-\inner{\bm{f}}{\bm{s}_{\bm{\theta}} - \bm{s}^{\textup{ODE}}_{\bm{\theta}}}   \Big]
\end{align*}
That is, 
\begin{align}\label{eq:diff_ode_nnscore}
    \partial_t \big(\bm{s}_{\bm{\theta}} (\bm{x}, t)- \bm{s}^{\textup{ODE}}_{\bm{\theta}}(\bm{x}, t)\big)
    = \bm{\epsilon}[{\bm{s_{\theta}}}](\bm{x}, t) +  \grad{\bm{x}}\Big[ \frac{1}{2}g^2 (t)\inner{\bm{s}_{\bm{\theta}} - \bm{s}^{\textup{ODE}}_{\bm{\theta}}}{\bm{s}_{\bm{\theta}}}-\inner{\bm{f}}{\bm{s}_{\bm{\theta}} - \bm{s}^{\textup{ODE}}_{\bm{\theta}}}   \Big]
\end{align}
Fix a $t\in[0,T]$. We integrate both sides of the above equation from $\tau=T$ to $\tau=t$  
\begin{align*}
    \bm{s}_{\bm{\theta}} (\bm{x}, t)- \bm{s}^{\textup{ODE}}_{\bm{\theta}}(\bm{x}, t)
    &=
    \bm{s}_{\bm{\theta}} (\bm{x}, T)- \bm{s}^{\textup{ODE}}_{\bm{\theta}}(\bm{x}, T)
    \\&+\displaystyle\int_{T}^{t}\bm{\epsilon}[{\bm{s_{\theta}}}](\bm{x}, \tau) d\tau + \displaystyle\int_{T}^{t} \grad{\bm{x}}\Big[ \frac{1}{2}g^2 (t)\inner{\bm{s}_{\bm{\theta}} - \bm{s}^{\textup{ODE}}_{\bm{\theta}}}{\bm{s}_{\bm{\theta}}}-\inner{\bm{f}}{\bm{s}_{\bm{\theta}} - \bm{s}^{\textup{ODE}}_{\bm{\theta}}}   \Big] d\tau.
\end{align*}

Applying the $\ell_2$-norm
\begin{align}\label{eq:fisher_s_sode}
    \norm{\bm{s}_{\bm{\theta}} (\bm{x}, t)- \bm{s}^{\textup{ODE}}_{\bm{\theta}}(\bm{x}, t)}_2
    &\leq \nonumber
    \norm{\bm{s}_{\bm{\theta}} (\bm{x}, T)- \bm{s}^{\textup{ODE}}_{\bm{\theta}}(\bm{x}, T)}_2
    \\&+ \displaystyle\int_{t}^{T}\norm{\bm{\epsilon}[{\bm{s_{\theta}}}](\bm{x}, \tau)}_2 d\tau  
    \\&+ \displaystyle\int_{t}^{T} \norm{\grad{\bm{x}}\Big[ \frac{1}{2}g^2 (t)\inner{\bm{s}_{\bm{\theta}} - \bm{s}^{\textup{ODE}}_{\bm{\theta}}}{\bm{s}_{\bm{\theta}}}-\inner{\bm{f}}{\bm{s}_{\bm{\theta}} - \bm{s}^{\textup{ODE}}_{\bm{\theta}}}   \Big] }_2 d\tau . \nonumber
\end{align}

In the last term, we may compute $\grad{\bm{x}}\Big[ \frac{1}{2}g^2 (t)\inner{\bm{s}_{\bm{\theta}} - \bm{s}^{\textup{ODE}}_{\bm{\theta}}}{\bm{s}_{\bm{\theta}}}-\inner{\bm{f}}{\bm{s}_{\bm{\theta}} - \bm{s}^{\textup{ODE}}_{\bm{\theta}}}   \Big]$ as 
\begin{equation}
\begin{aligned}\label{eq:last_rearragne}
\frac{1}{2}g^2(\tau)\Big( \grad{\bm{x}}\bm{s}_{\bm{\theta}}\cdot\bm{s}_{\bm{\theta}} - \grad{\bm{x}}\bm{s}^{\textup{ODE}}_{\bm{\theta}}\cdot\bm{s}_{\bm{\theta}}\Big)
    &+\frac{1}{2}g^2(\tau)\Big( \grad{\bm{x}}\bm{s}_{\bm{\theta}}\cdot \big(\bm{s}_{\bm{\theta}} - \bm{s}^{\textup{ODE}}_{\bm{\theta}}\big) \Big)
    \\&~-\grad{\bm{x}}\bm{f}\cdot\big(\bm{s}_{\bm{\theta}} - \bm{s}^{\textup{ODE}}_{\bm{\theta}}\big) - \grad{\bm{x}}\bm{s}_{\bm{\theta}}\cdot\bm{f} + \grad{\bm{x}}\bm{s}^{\textup{ODE}}_{\bm{\theta}}\cdot\bm{f} 
\end{aligned}
\end{equation}

Hence, we can further estimate the last term of Ineq.~\eqref{eq:fisher_s_sode} as 
\begin{align*}
    &~\displaystyle\int_{t}^{T} \norm{\grad{\bm{x}}\Big[ \frac{1}{2}g^2 (t)\inner{\bm{s}_{\bm{\theta}} - \bm{s}^{\textup{ODE}}_{\bm{\theta}}}{\bm{s}_{\bm{\theta}}}-\inner{\bm{f}}{\bm{s}_{\bm{\theta}} - \bm{s}^{\textup{ODE}}_{\bm{\theta}}}   \Big] }_2 d\tau
    \\ \leq&~
    \displaystyle\int_{t}^{T}\frac{1}{2}g^2(\tau)\norm{\grad{\bm{x}}\bm{s}_{\bm{\theta}}\cdot\bm{s}_{\bm{\theta}}  }_2d\tau + \displaystyle\int_{t}^{T}\frac{1}{2}g^2(\tau)\norm{\grad{\bm{x}}\bm{s}^{\textup{ODE}}_{\bm{\theta}}\cdot\bm{s}_{\bm{\theta}}  }_2d\tau 
    \\+&~
    \displaystyle\int_{t}^{T}\frac{1}{2}g^2(\tau)\norm{ \grad{\bm{x}}\bm{s}_{\bm{\theta}}\cdot \big(\bm{s}_{\bm{\theta}} - \bm{s}^{\textup{ODE}}_{\bm{\theta}}\big) }_2d\tau 
    +\displaystyle\int_{t}^{T}\norm{ \grad{\bm{x}}\bm{f}\cdot\big(\bm{s}_{\bm{\theta}} - \bm{s}^{\textup{ODE}}_{\bm{\theta}}\big) }_2d\tau 
    \\+&~
    \displaystyle\int_{t}^{T}\norm{ \grad{\bm{x}}\bm{s}_{\bm{\theta}}\cdot\bm{f} }_2d\tau + 
    \displaystyle\int_{t}^{T}\norm{ \grad{\bm{x}}\bm{s}^{\textup{ODE}}_{\bm{\theta}}\cdot\bm{f} }_2d\tau
    \\   \leq&~ 
     \displaystyle\int_{t}^{T}\frac{1}{2}g^2(\tau)\norm{\grad{\bm{x}}\bm{s}_{\bm{\theta}}}_{\text{op}}\norm{\bm{s}_{\bm{\theta}}  }_2d\tau +  \displaystyle\int_{t}^{T}\frac{1}{2}g^2(\tau)\norm{\grad{\bm{x}}\bm{s}^{\textup{ODE}}_{\bm{\theta}}}_{\text{op}}\norm{\bm{s}_{\bm{\theta}}  }_2d\tau
    \\+&~
    \displaystyle\int_{t}^{T}\frac{1}{2}g^2(\tau)\norm{ \grad{\bm{x}}\bm{s}_{\bm{\theta}}}_{\text{op}}\norm{ \bm{s}_{\bm{\theta}} - \bm{s}^{\textup{ODE}}_{\bm{\theta}} }_2d\tau 
    +\displaystyle\int_{t}^{T}\norm{ \grad{\bm{x}}\bm{f}}_{\text{op}}\norm{\bm{s}_{\bm{\theta}} - \bm{s}^{\textup{ODE}}_{\bm{\theta}} }_2d\tau 
    \\+&~
        \displaystyle\int_{t}^{T}\norm{ \grad{\bm{x}}\bm{s}_{\bm{\theta}}}_{\text{op}}\norm{\bm{f} }_2d\tau
        + \displaystyle\int_{t}^{T}\norm{ \grad{\bm{x}}\bm{s}^{\textup{ODE}}_{\bm{\theta}}}_{\text{op}}\norm{\bm{f} }_2d\tau
     \\ \leq&~ \Bigg[L^2  (\int_{0}^{T} g^2(\tau)  d\tau) (1+\norm{\bm{x}}_2)\Bigg] + \Bigg[\displaystyle\int_{t}^{T}\big(\frac{L}{2}g^2(\tau) + L \big)\norm{ \bm{s}_{\bm{\theta}} - \bm{s}^{\textup{ODE}}_{\bm{\theta}} }_2d\tau \Bigg] 
     + \Bigg[2L^2 T (1+\norm{\bm{x}}_2) \Bigg]
     \\ \leq&~ C_1(L, T, g) (1+\norm{\bm{x}}_2) + \displaystyle\int_{t}^{T}\big(\frac{L}{2}g^2(\tau) + L \big)\norm{ \bm{s}_{\bm{\theta}} - \bm{s}^{\textup{ODE}}_{\bm{\theta}} }_2d\tau
\end{align*}

where $\norm{\bm{A}}_{\textup{op}}:=\max_{\bm{x}\neq \bm{0} }\frac{\norm{\bm{A}\bm{x}}_2}{\norm{\bm{x}}_2}$ denotes the operator norm of the matrix $\bm{A}$. In the second-to-last inequality, we apply Assumption~\ref{cond:A} together with the Rademacher’s theorem~\citep{evans2018measure} which bounds the total differentiations of $\bm{s}_{\bm{\theta}}$, $\bm{s}^{\textup{ODE}}_{\bm{\theta}}$, and $\bm{f}$ by their Lipschitz constants. Moreover, we summarize constant terms into $C_1:=C_1(L, T, g)$, which depends on $L$, $T$, and the function $g$. 

Combining this estimation with Ineq.~\eqref{eq:fisher_s_sode}, we have
\begin{align*}
    \norm{\bm{s}_{\bm{\theta}} (\bm{x}, t)- \bm{s}^{\textup{ODE}}_{\bm{\theta}}(\bm{x}, t)}_2
    &\leq \nonumber
    \norm{\bm{s}_{\bm{\theta}} (\bm{x}, T)- \bm{s}^{\textup{ODE}}_{\bm{\theta}}(\bm{x}, T)}_2 + \displaystyle\int_{0}^{T}\norm{\bm{\epsilon}[{\bm{s_{\theta}}}](\bm{x}, \tau)}_2 d\tau  + C_1(L, T, g) (1+\norm{\bm{x}}_2) 
    \\&+  \displaystyle\int_{t}^{T}\big(\frac{L}{2}g^2(\tau) + L \big)\norm{ \bm{s}_{\bm{\theta}} (\bm{x}, \tau)- \bm{s}^{\textup{ODE}}_{\bm{\theta}}(\bm{x}, \tau) }_2d\tau. \nonumber
\end{align*}
Consider the following functions in Lemma~\ref{th:gronwall}
\begin{align*}
    u(t)&:= \norm{\bm{s}_{\bm{\theta}} (\bm{x}, t)- \bm{s}^{\textup{ODE}}_{\bm{\theta}}(\bm{x}, t)}_2
    \\\alpha(t)
    &:=\norm{\bm{s}_{\bm{\theta}} (\bm{x}, T)- \bm{s}^{\textup{ODE}}_{\bm{\theta}}(\bm{x}, T)}_2 + \displaystyle\int_{0}^{T}\norm{\bm{\epsilon}[{\bm{s_{\theta}}}](\bm{x}, \tau)}_2 d\tau  + C_1(L, T, g) (1+\norm{\bm{x}}_2) 
    \\\beta(t)&:= \frac{L}{2}g^2(t) + L.
\end{align*}
We remark that $\alpha\equiv\alpha(t)$ is actually independent of $t$. Then the lemma implies
\begin{align*}
    u(t) \leq&~ \alpha \exp\big(\int_{t}^{T}\beta(\tau)d\tau \big)
    \\ \lesssim&~ \Big[ \norm{\bm{s}_{\bm{\theta}} (\bm{x}, T)- \bm{s}^{\textup{ODE}}_{\bm{\theta}}(\bm{x}, T)}_2 + \displaystyle\int_{0}^{T}\norm{\bm{\epsilon}[{\bm{s_{\theta}}}](\bm{x}, \tau)}_2 d\tau  +  (1+\norm{\bm{x}}_2) \Big],
\end{align*}
where we bound $\exp\big(\int_{t}^{T}\beta(\tau)d\tau \big)$ by $\exp\big(\int_{0}^{T}\beta(\tau)d\tau \big)$ which is a constant, and we absorb all constant terms.

We are going to square both sides of the above estimation and take the expectation over $q_t(\bm{x})$. For the sake of simplicity, we denote $e_{\bm{\theta}}(\bm{x}):=\int_{0}^{T}\norm{\bm{\epsilon}[{\bm{s_{\theta}}}](\bm{x}, \tau)}_2 d\tau$ and $\delta_{\bm{\theta}}(\bm{x}):=\norm{\bm{s}_{\bm{\theta}} (\bm{x}, T)- \bm{s}^{\textup{ODE}}_{\bm{\theta}}(\bm{x}, T)}_2$, and hence, we obtain
\begin{equation}\label{eq:exp_fisher_s_sode}
\begin{aligned}
    \mathbb{E}_{q_{t}(\bm{x})}\big[u^2 (t)\big] 
    \lesssim &~ \mathbb{E}_{q_{t}(\bm{x})}\Bigg( \delta_{\bm{\theta}}(\bm{x}) + e_{\bm{\theta}}(\bm{x})  +  (1+\norm{\bm{x}}_2) \Bigg)^2 
    \\ \lesssim &~  \Bigg\{ \mathbb{E}_{q_{t}(\bm{x})}\big[\delta_{\bm{\theta}}^2(\bm{x})\big] + \mathbb{E}_{q_{t}(\bm{x})}\big[e_{\bm{\theta}}^2(\bm{x})\big] + \mathbb{E}_{q_{t}(\bm{x})}\Big[(1+\norm{\bm{x}}_2)^2\Big]  
    \\&+~\mathbb{E}_{q_{t}(\bm{x})}\big[\delta_{\bm{\theta}}(\bm{x}) e_{\bm{\theta}}(\bm{x}) \big]+\mathbb{E}_{q_{t}(\bm{x})}\big[\delta_{\bm{\theta}}(\bm{x}) (1+\norm{\bm{x}}_2) \big]+\mathbb{E}_{q_{t}(\bm{x})}\big[e_{\bm{\theta}}(\bm{x}) (1+\norm{\bm{x}}_2)\big] \Bigg\}. 
\end{aligned}
\end{equation}
The last three terms of the above inequality can be further bounded via Cauchy–Schwarz inequality
\begin{align*}
    \mathbb{E}_{q_{t}(\bm{x})}\big[\delta_{\bm{\theta}}(\bm{x}) e_{\bm{\theta}}(\bm{x}) \big] &\leq \sqrt{\mathbb{E}_{q_{t}(\bm{x})}\big[\delta_{\bm{\theta}}^2(\bm{x}) \big]} \sqrt{\mathbb{E}_{q_{t}(\bm{x})}\big[ e_{\bm{\theta}}^2(\bm{x}) \big]}
    \\ \mathbb{E}_{q_{t}(\bm{x})}\big[\delta_{\bm{\theta}}(\bm{x}) (1+\norm{\bm{x}}_2) \big] &\leq \sqrt{\mathbb{E}_{q_{t}(\bm{x})}\big[\delta_{\bm{\theta}}^2(\bm{x})\big]} \sqrt{\mathbb{E}_{q_{t}(\bm{x})}\big[ (1+\norm{\bm{x}}_2)^2 \big]}
    \\ \mathbb{E}_{q_{t}(\bm{x})}\big[e_{\bm{\theta}}(\bm{x}) (1+\norm{\bm{x}}_2)\big] &\leq 
    \sqrt{\mathbb{E}_{q_{t}(\bm{x})}\big[e_{\bm{\theta}}^2(\bm{x})\big]} \sqrt{\mathbb{E}_{q_{t}(\bm{x})}\big[(1+\norm{\bm{x}}_2)^2\big]}.
\end{align*}
It is noticed that Assumption~\ref{cond:A}.\ref{cond:expect_x} indeed implies the following estimation which bounds $1^\text{st}$- and $2^\text{nd}$- central moments for all $t\in[0,T]$
\begin{equation}\label{eq:expect_x_all_t}
    \sup_{t\in[0,T]}\Big\{ \mathbb{E}_{\bm{x} \sim q_{t}(\bm{x})}[\norm{\bm{x}}_2]\Big\} \leq L \quad\text{and}\quad \sup_{t\in[0,T]}\Big\{\mathbb{E}_{\bm{x} \sim q_{t}(\bm{x})}[\norm{\bm{x}}_2^2] \Big\} \leq L
\end{equation}
as by Cauchy Schwartz inequality that $ \mathbb{E}_{\bm{x} \sim q_{0}(\bm{x})}[\norm{\bm{x}}_2] \leq \big(\mathbb{E}_{\bm{x} \sim q_{0}(\bm{x})}[\norm{\bm{x}}_2^2] \big)^{1/2}\leq L^{1/2}$ and the transition density $q_{0t}(\bm{x}(t)|\bm{x}(0))$ has bounded covariance matrices as a function in $t\in[0,T]$.
With Ineq.~\eqref{eq:expect_x_all_t} and Assumption~\ref{cond:A}.(j), Ineq.~\eqref{eq:exp_fisher_s_sode} becomes
\begin{align*}
    &\mathbb{E}_{q_{t}(\bm{x})}\Big[\norm{\bm{s}_{\bm{\theta}} (\bm{x}, t)- \bm{s}^{\textup{ODE}}_{\bm{\theta}}(\bm{x}, t)}_2^2\Big] \nonumber
       \\ \lesssim&~  \Bigg\{ \delta_T^2 + \mathbb{E}_{q_{t}(\bm{x})}\big[e_{\bm{\theta}}^2(\bm{x})\big] + (1+3L) \nonumber 
    \\&+~ \delta_T \sqrt{\mathbb{E}_{q_{t}(\bm{x})}\big[ e_{\bm{\theta}}^2(\bm{x}) \big]} + \delta_T \sqrt{1+3L} + \sqrt{1+3L} \sqrt{\mathbb{E}_{q_{t}(\bm{x})}\big[ e_{\bm{\theta}}^2(\bm{x}) \big]} \Bigg\}
    \\ \lesssim&~  \Big(\mathbb{E}_{q_{t}(\bm{x})}\big[e_{\bm{\theta}}^2(\bm{x}) \big] + \sqrt{\mathbb{E}_{q_{t}(\bm{x})}\big[e_{\bm{\theta}}^2(\bm{x})\big]} + C_1(L, T, g, \delta_T) \Big)\nonumber
    \\ \lesssim&~ \Big(M({\bm{\theta}}) + \sqrt{M({\bm{\theta}})} + C_1(L, T, g, \delta_T) \Big)\nonumber.
\end{align*}
Here we use that $e^2_{\bm{\theta}}(\bm{x})=\big(\int_{0}^{T}\norm{\bm{\epsilon}[{\bm{s_{\theta}}}](\bm{x}, \tau)}_2 d\tau\big)^2\leq T \int_{0}^{T}\norm{\bm{\epsilon}[{\bm{s_{\theta}}}](\bm{x}, \tau)}_2^2 d\tau $.
 Again, we we abuse of the notation and summarize constants into $C_1=C_1(L, T, g, \delta_T)$. Therefore, after combining the Ineq.~\eqref{eq:diff_sm_fisher_add} and the estimation above, we obtain (with a fusion of constant term)
\begin{align*}
    \Big(\mathcal{J}_{\text{Diff}}(\bm{\theta})\Big)^2 &\lesssim \mathcal{J}_{\text{SM}}(\bm{\theta})\cdot \mathcal{J}_{\text{Fisher}}(\bm{\theta})
    \\ & \lesssim  \mathcal{J}_{\text{SM}}(\bm{\theta}) \cdot \Big(M({\bm{\theta}}) + \sqrt{M({\bm{\theta}})} +  C_1(L, T, g, \delta_T)  \Big)
\end{align*}

\end{proof}

\subsection{Discussions on Theorem~\ref{th:min_ode}}\label{subsec:disc-th-min-ode}
\textbf{Tighter bounds of Theorem~\ref{th:min_ode}. } We remark that one can easily extend Theorem~\ref{th:min_ode} and obtain a sharper bound by checking carefully the tightness of each estimation. We provide an approach as an instance. Let us assume there is a constant $\delta_{\text{ODE}}>0$ to control the distance between $\grad{\bm{x}} \bm{s}^{\textup{ODE}}_{\bm{\theta}}$ and $\grad{\bm{x}}\bm{s}_{\bm{\theta}}$ instead (in this case, we do not require Assumption~\ref{cond:A}.\ref{cond:ode_lip}). That is,
\begin{align}\label{eq:bdd_grad_ode}
\sup_{\mathbb{R}^D \times [0, T]}\norm{\grad{\bm{x}}\big( \bm{s}_{\bm{\theta}} - \bm{s}^{\textup{ODE}}_{\bm{\theta}} \big)}_2 \leq \delta_{\text{ODE}}.
\end{align}

Notice that Eq.~\eqref{eq:last_rearragne} can be rewritten as 
\begin{align*}
\frac{1}{2}g^2(\tau)\Big(  \grad{\bm{x}}\big( \bm{s}_{\bm{\theta}} - \bm{s}^{\textup{ODE}}_{\bm{\theta}} \big) \cdot\bm{s}_{\bm{\theta}}\Big)
    &+\frac{1}{2}g^2(\tau)\Big( \grad{\bm{x}}\bm{s}_{\bm{\theta}}\cdot \big(\bm{s}_{\bm{\theta}} - \bm{s}^{\textup{ODE}}_{\bm{\theta}}\big) \Big)
    \\&~-\grad{\bm{x}}\bm{f}\cdot\big(\bm{s}_{\bm{\theta}} - \bm{s}^{\textup{ODE}}_{\bm{\theta}}\big) - \grad{\bm{x}}\big( \bm{s}_{\bm{\theta}} - \bm{s}^{\textup{ODE}}_{\bm{\theta}} \big) \cdot\bm{f}.
\end{align*}
Following the same argument as the proof of Theorem.~\ref{th:min_ode} together with the help of Ineq.~\eqref{eq:bdd_grad_ode}, $\mathcal{J}_{\text{Fisher}}(\bm{\theta})$ can be upper bounded by a constant which depends monotonically increasingly on $\delta_{\text{ODE}}$. Therefore, we can get a sharper estimation if $\delta_{\text{ODE}}$ is smaller. 

\textbf{Variants of Theorem~\ref{th:min_ode}. } $M({\bm{\theta}})$ describes one of the worst case scenarios by searching the largest time-averaged residuals among all time slices. Theoretically it is a legit quantity. While we can obtain similar results as Theorems~\ref{th:min_ode} and \ref{th:min_ode_add} (however, need additional assumptions) with more interpretable bounds. Here we just focus on the discussion on Theorem~\ref{th:min_ode}, and a similar argument can be adopted to Theorem~\ref{th:min_ode_add}.

First, we introduce a variant of $M({\bm{\theta}})$ which is defined as
 \begin{align}\label{eq:new_M}
      \widetilde{M}({\bm{\theta}}):=\sup_{t\in[t_0,T]} \mathbb{E}_{\bm{x} \sim q_{t}(\bm{x})}\left[\int_{t_0}^{T} {q_{\tau}(\bm{x})}\norm{\bm{\epsilon}[{\bm{s_{\theta}}}](\bm{x}, \tau)}_2^2 d\tau\right].
 \end{align}

$\widetilde{M}({\bm{\theta}})$ is more interpretable in the sense that for any time slice $t$ and sample $\bm{x}_t\sim q_t(\bm{x})$, if $\tau$
 is deviating from $t$ when taking the time average, the marginal density $q_{\tau}(\bm{x}_t)$ can reduce the effect of the score FPE residual at that by providing a lighter weight.

Notice that if we assume that there is positive constants $Q^*$ and $q^*$ so that $q^* \leq  q_{t}(\bm{x})\leq Q^*$ for all $\bm{x}\in\textup{supp}(q_t)$ and $t\in[t_0,T]$, then $q^* M({\bm{\theta}}) \leq \widetilde{M}({\bm{\theta}}) \leq Q^* M({\bm{\theta}})$.

\begin{theorem}[A variant of Theorem~\ref{th:min_ode}]\label{th:variant_thm4.2} Let $t_0>0$ be a constant indicating the terminal time of the backward diffusion process (or equivalently the initial time of the forward process). Suppose in addition to Assumption~\ref{cond:A} that we have bounded $4^\text{th}$ non-central moment: $\mathbb{E}_{q_{0}(\bm{x})}[\norm{\bm{x}}_2^4] \leq L$.  If there is positive constants $Q^*$ and $q^*$ so that $Q^* \geq  q_{t}(\bm{x})\geq q^*$ for all $\bm{x}\in\textup{supp}(q_t)$ and $t\in[t_0,T]$. Then we can obtain an upper bound for $\mathcal{J}_{\text{Fisher}}(\bm{\theta})$ in terms of $\widetilde{M}({\bm{\theta}})$ as
\begin{equation*}
     \mathcal{J}_{\text{Fisher}}(\bm{\theta}) \lesssim  \widetilde{M}({\bm{\theta}}) + \sqrt{\widetilde{M}({\bm{\theta}})} + C_3,
\end{equation*}
where $C_3>0$ is a constant independent of $\bm{\theta}$ and different from $C_1$ and $C_2$.
\end{theorem}

\begin{proof}{Theorem}~{\ref{th:variant_thm4.2}} 

We start from Eq.~\eqref{eq:diff_ode_nnscore} in the proof of Theorem~\ref{th:min_ode}. 
\begin{align*}
    \partial_t \big(\bm{s}_{\bm{\theta}} (\bm{x}, t)- \bm{s}^{\textup{ODE}}_{\bm{\theta}}(\bm{x}, t)\big)
    = \bm{\epsilon}[{\bm{s_{\theta}}}](\bm{x}, t) +  \grad{\bm{x}}\Big[ \frac{1}{2}g^2 (t)\inner{\bm{s}_{\bm{\theta}} - \bm{s}^{\textup{ODE}}_{\bm{\theta}}}{\bm{s}_{\bm{\theta}}}-\inner{\bm{f}}{\bm{s}_{\bm{\theta}} - \bm{s}^{\textup{ODE}}_{\bm{\theta}}}   \Big].
\end{align*}

Multiplying $q_t(\bm{x})$ from the both sides and integrating from $\tau=T$ to $\tau=t$, we get the following equation with integration by part

\begin{align*}
    \text{LHS}&:=\displaystyle\int_{T}^{t}\partial_{\tau} \big(\bm{s}_{\bm{\theta}} (\bm{x}, \tau)- \bm{s}^{\textup{ODE}}_{\bm{\theta}}(\bm{x}, \tau)\big) q_{\tau}(\bm{x})~d\tau \\
    &= \big(\bm{s}_{\bm{\theta}} (\bm{x}, \tau)- \bm{s}^{\textup{ODE}}_{\bm{\theta}}(\bm{x}, \tau)\big) q_{\tau}(\bm{x})\Big|_{\tau=T}^{\tau=t} - \displaystyle\int_{T}^{t} \big(\bm{s}_{\bm{\theta}} (\bm{x}, \tau)- \bm{s}^{\textup{ODE}}_{\bm{\theta}}(\bm{x}, \tau)\big) \partial_{\tau} q_{\tau}(\bm{x})~d\tau \\
     &= \big(\bm{s}_{\bm{\theta}} (\bm{x}, t)- \bm{s}^{\textup{ODE}}_{\bm{\theta}}(\bm{x}, t)\big) q_{t}(\bm{x}) - \Big[\big(\bm{s}_{\bm{\theta}} (\bm{x}, T)- \bm{s}^{\textup{ODE}}_{\bm{\theta}}(\bm{x}, T)\big) q_{T}(\bm{x}) +
     \displaystyle\int_{T}^{t} \big(\bm{s}_{\bm{\theta}} (\bm{x}, \tau)- \bm{s}^{\textup{ODE}}_{\bm{\theta}}(\bm{x}, \tau)\big) \partial_{\tau} q_{\tau}(\bm{x})~d\tau\Big]
\end{align*}
and
\begin{align*}
    \text{RHS}&:=\displaystyle\int_{T}^{t}\bm{\epsilon}[{\bm{s_{\theta}}}](\bm{x}, \tau)q_{\tau}(\bm{x})~d\tau + \displaystyle\int_{T}^{t}  \grad{\bm{x}}\Big[ \frac{1}{2}g^2 (t)\inner{\bm{s}_{\bm{\theta}} - \bm{s}^{\textup{ODE}}_{\bm{\theta}}}{\bm{s}_{\bm{\theta}}}-\inner{\bm{f}}{\bm{s}_{\bm{\theta}} - \bm{s}^{\textup{ODE}}_{\bm{\theta}}}   \Big]q_{\tau}(\bm{x})~d\tau.
\end{align*}

Hence, we have

\begin{align*}
    \big(\bm{s}_{\bm{\theta}} (\bm{x}, t)- \bm{s}^{\textup{ODE}}_{\bm{\theta}}(\bm{x}, t)\big) q_{t}(\bm{x})
    &=
    \Big[\big(\bm{s}_{\bm{\theta}} (\bm{x}, T)- \bm{s}^{\textup{ODE}}_{\bm{\theta}}(\bm{x}, T)\big) q_{T}(\bm{x}) +
     \displaystyle\int_{T}^{t} \big(\bm{s}_{\bm{\theta}} (\bm{x}, \tau)- \bm{s}^{\textup{ODE}}_{\bm{\theta}}(\bm{x}, \tau)\big) \partial_{\tau} q_{\tau}(\bm{x})d\tau\Big]
    \\&+\displaystyle\int_{T}^{t}\bm{\epsilon}[{\bm{s_{\theta}}}](\bm{x}, \tau)q_{\tau}(\bm{x}) d\tau + \displaystyle\int_{T}^{t} \grad{\bm{x}}\Big[ \frac{1}{2}g^2 (t)\inner{\bm{s}_{\bm{\theta}} - \bm{s}^{\textup{ODE}}_{\bm{\theta}}}{\bm{s}_{\bm{\theta}}}-\inner{\bm{f}}{\bm{s}_{\bm{\theta}} - \bm{s}^{\textup{ODE}}_{\bm{\theta}}}   \Big] q_{\tau}(\bm{x})d\tau.
\end{align*}

Applying the $\ell_2$-norm and using the fact that $Q^* \geq  q_{t}(\bm{x})\geq q^*$ for all $\bm{x}\in\textup{supp}(q_t)$ and $t\in[t_0,T]$ (or alternatively, we may choose sufficiently large $L>0$ so that $L \geq  q_{t}(\bm{x})\geq 1/L$),
\begin{align}\label{eq:fisher_s_sode_var}
    q_{t}(\bm{x})\norm{\bm{s}_{\bm{\theta}} (\bm{x}, t)- \bm{s}^{\textup{ODE}}_{\bm{\theta}}(\bm{x}, t)}_2
    &\lesssim \nonumber
    \norm{\bm{s}_{\bm{\theta}} (\bm{x}, T)- \bm{s}^{\textup{ODE}}_{\bm{\theta}}(\bm{x}, T)}_2+ 
     \displaystyle\int_{T}^{t} \norm{\bm{s}_{\bm{\theta}} (\bm{x}, \tau)- \bm{s}^{\textup{ODE}}_{\bm{\theta}}(\bm{x}, \tau)} \abs{ \partial_{\tau} q_{\tau}(\bm{x})}d\tau
    \\&+ \displaystyle\int_{t}^{T}q_{\tau}(\bm{x})\norm{\bm{\epsilon}[{\bm{s_{\theta}}}](\bm{x}, \tau)}_2 d\tau   
    \\&+ \displaystyle\int_{t}^{T} \norm{\grad{\bm{x}}\Big[ \frac{1}{2}g^2 (t)\inner{\bm{s}_{\bm{\theta}} - \bm{s}^{\textup{ODE}}_{\bm{\theta}}}{\bm{s}_{\bm{\theta}}}-\inner{\bm{f}}{\bm{s}_{\bm{\theta}} - \bm{s}^{\textup{ODE}}_{\bm{\theta}}}   \Big] }_2 d\tau . \nonumber
\end{align}

With the classic Fokker-Planck equation and $\bm{s}(\bm{x}, \tau) =\grad{\bm{x}}q_{\tau}(\bm{x})/q_{\tau}(\bm{x})$
\begin{align*}
    \partial_{\tau} q_{\tau}(\bm{x}) = -\big[(\div{\bm{x}}(\bm{f})-\frac{1}{2}g^2(\tau)\div{\bm{x}}(\bm{s}))q_{\tau} + \inner{\bm{f} - \frac{1}{2}g^2(\tau) \bm{s}}{\bm{s}}q_{\tau} \big],
\end{align*}
we can bound the term $\displaystyle\int_{T}^{t} \norm{\bm{s}_{\bm{\theta}} (\bm{x}, \tau)- \bm{s}^{\textup{ODE}}_{\bm{\theta}}(\bm{x}, \tau)} \abs{ \partial_{\tau} q_{\tau}(\bm{x})}d\tau \leq C_3 (1+\norm{\bm{x}}_2)^2$,
where $C_3:=C_3(L, T, g,  Q^*, q^*)>0$ is a constant depending on $L$, $T$,  $Q^*$, $q^*$, and the function $g$.

In the last term of Ineq.~\eqref{eq:fisher_s_sode}, we may compute $\grad{\bm{x}}\Big[ \frac{1}{2}g^2 (t)\inner{\bm{s}_{\bm{\theta}} - \bm{s}^{\textup{ODE}}_{\bm{\theta}}}{\bm{s}_{\bm{\theta}}}-\inner{\bm{f}}{\bm{s}_{\bm{\theta}} - \bm{s}^{\textup{ODE}}_{\bm{\theta}}}   \Big]$ as 
\begin{equation*}
\begin{aligned}
\frac{1}{2}g^2(\tau)\Big( \grad{\bm{x}}\bm{s}_{\bm{\theta}}\cdot\bm{s}_{\bm{\theta}} - \grad{\bm{x}}\bm{s}^{\textup{ODE}}_{\bm{\theta}}\cdot\bm{s}_{\bm{\theta}}\Big)
    &+\frac{1}{2}g^2(\tau)\Big( \grad{\bm{x}}\bm{s}_{\bm{\theta}}\cdot \big(\bm{s}_{\bm{\theta}} - \bm{s}^{\textup{ODE}}_{\bm{\theta}}\big) \Big)
    \\&~-\grad{\bm{x}}\bm{f}\cdot\big(\bm{s}_{\bm{\theta}} - \bm{s}^{\textup{ODE}}_{\bm{\theta}}\big) - \grad{\bm{x}}\bm{s}_{\bm{\theta}}\cdot\bm{f} + \grad{\bm{x}}\bm{s}^{\textup{ODE}}_{\bm{\theta}}\cdot\bm{f} 
\end{aligned}
\end{equation*}

Hence, we can further estimate the last term of Ineq.~\eqref{eq:fisher_s_sode_var} as 
\begin{align*}
    &~\displaystyle\int_{t}^{T} \norm{\grad{\bm{x}}\Big[ \frac{1}{2}g^2 (t)\inner{\bm{s}_{\bm{\theta}} - \bm{s}^{\textup{ODE}}_{\bm{\theta}}}{\bm{s}_{\bm{\theta}}}-\inner{\bm{f}}{\bm{s}_{\bm{\theta}} - \bm{s}^{\textup{ODE}}_{\bm{\theta}}}   \Big] }_2 d\tau
    \\ \leq&~
    \displaystyle\int_{t}^{T}\frac{1}{2}g^2(\tau)\norm{\grad{\bm{x}}\bm{s}_{\bm{\theta}}\cdot\bm{s}_{\bm{\theta}}  }_2d\tau + \displaystyle\int_{t}^{T}\frac{1}{2}g^2(\tau)\norm{\grad{\bm{x}}\bm{s}^{\textup{ODE}}_{\bm{\theta}}\cdot\bm{s}_{\bm{\theta}}  }_2d\tau 
    \\+&~
    \displaystyle\int_{t}^{T}\frac{1}{2}g^2(\tau)\norm{ \grad{\bm{x}}\bm{s}_{\bm{\theta}}\cdot \big(\bm{s}_{\bm{\theta}} - \bm{s}^{\textup{ODE}}_{\bm{\theta}}\big) }_2d\tau 
    +\displaystyle\int_{t}^{T}\norm{ \grad{\bm{x}}\bm{f}\cdot\big(\bm{s}_{\bm{\theta}} - \bm{s}^{\textup{ODE}}_{\bm{\theta}}\big) }_2d\tau 
    \\+&~
    \displaystyle\int_{t}^{T}\norm{ \grad{\bm{x}}\bm{s}_{\bm{\theta}}\cdot\bm{f} }_2d\tau + 
    \displaystyle\int_{t}^{T}\norm{ \grad{\bm{x}}\bm{s}^{\textup{ODE}}_{\bm{\theta}}\cdot\bm{f} }_2d\tau
    \\   \leq&~ 
     \displaystyle\int_{t}^{T}\frac{1}{2}g^2(\tau)\norm{\grad{\bm{x}}\bm{s}_{\bm{\theta}}}_{\text{op}}\norm{\bm{s}_{\bm{\theta}}  }_2d\tau +  \displaystyle\int_{t}^{T}\frac{1}{2}g^2(\tau)\norm{\grad{\bm{x}}\bm{s}^{\textup{ODE}}_{\bm{\theta}}}_{\text{op}}\norm{\bm{s}_{\bm{\theta}}  }_2d\tau
    \\+&~
    \displaystyle\int_{t}^{T}\frac{1}{2}g^2(\tau)\norm{ \grad{\bm{x}}\bm{s}_{\bm{\theta}}}_{\text{op}}\norm{ \bm{s}_{\bm{\theta}} - \bm{s}^{\textup{ODE}}_{\bm{\theta}} }_2d\tau 
    +\displaystyle\int_{t}^{T}\norm{ \grad{\bm{x}}\bm{f}}_{\text{op}}\norm{\bm{s}_{\bm{\theta}} - \bm{s}^{\textup{ODE}}_{\bm{\theta}} }_2d\tau 
    \\+&~
        \displaystyle\int_{t}^{T}\norm{ \grad{\bm{x}}\bm{s}_{\bm{\theta}}}_{\text{op}}\norm{\bm{f} }_2d\tau
        + \displaystyle\int_{t}^{T}\norm{ \grad{\bm{x}}\bm{s}^{\textup{ODE}}_{\bm{\theta}}}_{\text{op}}\norm{\bm{f} }_2d\tau
     \\ \leq&~ \Bigg[L^2  (\int_{0}^{T} g^2(\tau)  d\tau) (1+\norm{\bm{x}}_2)\Bigg] + \Bigg[\displaystyle\int_{t}^{T}\big(\frac{L}{2}g^2(\tau) + L \big)\norm{ \bm{s}_{\bm{\theta}} - \bm{s}^{\textup{ODE}}_{\bm{\theta}} }_2d\tau \Bigg] 
     + \Bigg[2L^2 T (1+\norm{\bm{x}}_2) \Bigg]
     \\ \leq&~ C_3(L, T, g) (1+\norm{\bm{x}}_2) + \displaystyle\int_{t}^{T}\big(\frac{L}{2}g^2(\tau) + L \big)\norm{ \bm{s}_{\bm{\theta}} - \bm{s}^{\textup{ODE}}_{\bm{\theta}} }_2d\tau
\end{align*}

where $\norm{\bm{A}}_{\textup{op}}:=\max_{\bm{x}\neq \bm{0} }\frac{\norm{\bm{A}\bm{x}}_2}{\norm{\bm{x}}_2}$ denotes the operator norm of the matrix $\bm{A}$. In the second-to-last inequality, we apply Assumption~\ref{cond:A} together with the Rademacher’s theorem which bounds the total differentiations of $\bm{s}_{\bm{\theta}}$, $\bm{s}^{\textup{ODE}}_{\bm{\theta}}$, and $\bm{f}$ by their Lipschitz constants. Moreover, we summarize constant terms into $C_3:=C_3(L, T, g, Q^*, q^*)$. 

By dividing both side of Ineq.~\eqref{eq:fisher_s_sode_var} with $q_t(\bm{x})$ and using that $\frac{1}{Q^*}\leq\frac{1}{q_t(\bm{x})}\leq \frac{1}{q^*}$ for all $t\in[t_0,T]$ and $\bm{x}\in\textup{supp}(q_t)$, we combine above estimations which leads to 
\begin{align*}
    \norm{\bm{s}_{\bm{\theta}} (\bm{x}, t)- \bm{s}^{\textup{ODE}}_{\bm{\theta}}(\bm{x}, t)}_2
    &\leq \nonumber
    \frac{1}{q^*} \norm{\bm{s}_{\bm{\theta}} (\bm{x}, T)- \bm{s}^{\textup{ODE}}_{\bm{\theta}}(\bm{x}, T)}_2 + \frac{1}{q^*}\displaystyle\int_{t_0}^{T}\norm{\bm{\epsilon}[{\bm{s_{\theta}}}](\bm{x}, \tau)}_2 d\tau  
    \\&+ C_3(L, T, g, q^*) [(1+\norm{\bm{x}}_2) + (1+\norm{\bm{x}}_2)^2]  \\&+  \displaystyle\int_{t}^{T}\frac{1}{q^*}\big(\frac{L}{2}g^2(\tau) + L \big)\norm{ \bm{s}_{\bm{\theta}} (\bm{x}, \tau)- \bm{s}^{\textup{ODE}}_{\bm{\theta}}(\bm{x}, \tau) }_2d\tau. \nonumber
\end{align*}
Consider the following functions in the Lemma~G.2 (Grönwall's inequality)
\begin{align*}
    u(t)&:= \norm{\bm{s}_{\bm{\theta}} (\bm{x}, t)- \bm{s}^{\textup{ODE}}_{\bm{\theta}}(\bm{x}, t)}_2
    \\
    \alpha(t)
    &:= \frac{1}{q^*} \norm{\bm{s}_{\bm{\theta}} (\bm{x}, T)- \bm{s}^{\textup{ODE}}_{\bm{\theta}}(\bm{x}, T)}_2 + \frac{1}{q^*} \displaystyle\int_{t_0}^{T}q_{\tau}(\bm{x})\norm{\bm{\epsilon}[{\bm{s_{\theta}}}](\bm{x}, \tau)}_2 d\tau  
    \\& +C_3(L, T, g, q^*) [(1+\norm{\bm{x}}_2) + (1+\norm{\bm{x}}_2)^2] 
    \\\beta(t)&:= \frac{1}{q^*} \big( \frac{L}{2}g^2(t) + L \big).
\end{align*}
We remark that $\alpha\equiv\alpha(t)$ is actually independent of $t$. Then the lemma implies
\begin{align*}
    u(t) \leq&~ \alpha \exp\big(\int_{t}^{T}\beta(\tau)d\tau \big)
    \\ \lesssim&~ \Big[ \norm{\bm{s}_{\bm{\theta}} (\bm{x}, T)- \bm{s}^{\textup{ODE}}_{\bm{\theta}}(\bm{x}, T)}_2 + \displaystyle\int_{t_0}^{T}q_{\tau}(\bm{x})\norm{\bm{\epsilon}[{\bm{s_{\theta}}}](\bm{x}, \tau)}_2 d\tau  +  (1+\norm{\bm{x}}_2) +  (1+\norm{\bm{x}}_2)^2 \Big],
\end{align*}
where we bound $\exp\big(\int_{t}^{T}\beta(\tau)d\tau \big)$ by $\exp\big(\int_{t_0}^{T}\beta(\tau)d\tau \big)$ which is a constant, and we absorb all constant terms.

We are going to square both sides of the above estimation and take the expectation over $q_t(\bm{x})$. For the sake of simplicity, we denote $e_{\bm{\theta}}(\bm{x}):=\int_{t_0}^{T}q_{\tau}(\bm{x})\norm{\bm{\epsilon}[{\bm{s_{\theta}}}](\bm{x}, \tau)}_2 d\tau$ and $\delta_{\bm{\theta}}(\bm{x}):=\norm{\bm{s}_{\bm{\theta}} (\bm{x}, T)- \bm{s}^{\textup{ODE}}_{\bm{\theta}}(\bm{x}, T)}_2$, and hence, we obtain
\begin{equation}\label{eq:exp_fisher_s_sode_var}
\begin{aligned}
    \mathbb{E}_{q_{t}(\bm{x})}\big[u^2 (t)\big] 
    \lesssim &~ \mathbb{E}_{q_{t}(\bm{x})}\Bigg( \delta_{\bm{\theta}}(\bm{x}) + e_{\bm{\theta}}(\bm{x})  +  \big(2+3\norm{\bm{x}}_2+\norm{\bm{x}}_2^2\big)  \Bigg)^2 
    \\ \lesssim &~  \Bigg\{ \mathbb{E}_{q_{t}(\bm{x})}\big[\delta_{\bm{\theta}}^2(\bm{x})\big] + \mathbb{E}_{q_{t}(\bm{x})}\big[e_{\bm{\theta}}^2(\bm{x})\big] + \mathbb{E}_{q_{t}(\bm{x})}\Big[\big(2+3\norm{\bm{x}}_2+\norm{\bm{x}}_2^2\big)^2\Big]  
    \\&+~\mathbb{E}_{q_{t}(\bm{x})}\big[\delta_{\bm{\theta}}(\bm{x}) e_{\bm{\theta}}(\bm{x}) \big]+\mathbb{E}_{q_{t}(\bm{x})}\big[\delta_{\bm{\theta}}(\bm{x}) \big(2+3\norm{\bm{x}}_2+\norm{\bm{x}}_2^2\big) \big]+\mathbb{E}_{q_{t}(\bm{x})}\big[e_{\bm{\theta}}(\bm{x}) \big(2+3\norm{\bm{x}}_2+\norm{\bm{x}}_2^2\big)\big] \Bigg\}. 
\end{aligned}
\end{equation}
The last three terms of the above inequality can be further bounded via Cauchy–Schwarz inequality
\begin{align*}
    \mathbb{E}_{q_{t}(\bm{x})}\big[\delta_{\bm{\theta}}(\bm{x}) e_{\bm{\theta}}(\bm{x}) \big] &\leq \sqrt{\mathbb{E}_{q_{t}(\bm{x})}\big[\delta_{\bm{\theta}}^2(\bm{x}) \big]} \sqrt{\mathbb{E}_{q_{t}(\bm{x})}\big[ e_{\bm{\theta}}^2(\bm{x}) \big]}
    \\ \mathbb{E}_{q_{t}(\bm{x})}\big[\delta_{\bm{\theta}}(\bm{x}) \big(2+3\norm{\bm{x}}_2+\norm{\bm{x}}_2^2\big) \big] &\leq \sqrt{\mathbb{E}_{q_{t}(\bm{x})}\big[\delta_{\bm{\theta}}^2(\bm{x})\big]} \sqrt{\mathbb{E}_{q_{t}(\bm{x})}\big[ \big(2+3\norm{\bm{x}}_2+\norm{\bm{x}}_2^2\big)^2 \big]}
    \\ \mathbb{E}_{q_{t}(\bm{x})}\big[e_{\bm{\theta}}(\bm{x}) \big(2+3\norm{\bm{x}}_2+\norm{\bm{x}}_2^2\big)\big] &\leq 
    \sqrt{\mathbb{E}_{q_{t}(\bm{x})}\big[e_{\bm{\theta}}^2(\bm{x})\big]} \sqrt{\mathbb{E}_{q_{t}(\bm{x})}\big[\big(2+3\norm{\bm{x}}_2+\norm{\bm{x}}_2^2\big)^2\big]}.
\end{align*}
It is noticed that bounded $4^\text{th}$ non-central moment $\mathbb{E}_{q_{0}(\bm{x})}[\norm{\bm{x}}_2^4] \leq L$ indeed implies the following estimation which bounds all lower non-central moments for all $t\in[t_0,T]$
\begin{equation}\label{eq:expect_x_all_t_var}
    \sup_{t\in[t_0,T]}\Big\{ \mathbb{E}_{\bm{x} \sim q_{t}(\bm{x})}[\norm{\bm{x}}_2]\Big\},\quad \sup_{t\in[t_0,T]}\Big\{\mathbb{E}_{\bm{x} \sim q_{t}(\bm{x})}[\norm{\bm{x}}_2^2] \Big\},\quad \sup_{t\in[t_0,T]}\Big\{\mathbb{E}_{\bm{x} \sim q_{t}(\bm{x})}[\norm{\bm{x}}_2^3] \Big\},\quad \sup_{t\in[t_0,T]}\Big\{\mathbb{E}_{\bm{x} \sim q_{t}(\bm{x})}[\norm{\bm{x}}_2^4] \Big\} \leq L
\end{equation}
as by Cauchy Schwartz inequality and the transition density $q_{0t}(\bm{x}(t)|\bm{x}(0))$ has bounded covariance matrices as a function in $t\in[t_0,T]$.
With Ineq.~\eqref{eq:expect_x_all_t_var} and Assumption~\ref{cond:A}.(j), Ineq.~\eqref{eq:exp_fisher_s_sode_var} becomes
\begin{equation}\label{eq:fisher_end_var}
    \begin{aligned}
    &\mathbb{E}_{q_{t}(\bm{x})}\Big[\norm{\bm{s}_{\bm{\theta}} (\bm{x}, t)- \bm{s}^{\textup{ODE}}_{\bm{\theta}}(\bm{x}, t)}_2^2\Big] \nonumber
       \\ \lesssim&~  \Bigg\{ \delta_T^2 + \mathbb{E}_{q_{t}(\bm{x})}\big[e_{\bm{\theta}}^2(\bm{x})\big] + (4+32L) \nonumber 
    \\&+~ \delta_T \sqrt{\mathbb{E}_{q_{t}(\bm{x})}\big[ e_{\bm{\theta}}^2(\bm{x}) \big]} + \delta_T \sqrt{4+32L} + \sqrt{4+32L} \sqrt{\mathbb{E}_{q_{t}(\bm{x})}\big[ e_{\bm{\theta}}^2(\bm{x}) \big]} \Bigg\}
    \\ \lesssim&~  \Big(\mathbb{E}_{q_{t}(\bm{x})}\big[e_{\bm{\theta}}^2(\bm{x}) \big] + \sqrt{\mathbb{E}_{q_{t}(\bm{x})}\big[e_{\bm{\theta}}^2(\bm{x})\big]} + C_3(L, T, g, \delta_T) \Big)\nonumber
    \\ \lesssim&~ \Big(\widetilde{M}({\bm{\theta}})) + \sqrt{\widetilde{M}({\bm{\theta}})} + C_3(L, T, g, q^*, \delta_T) \Big),
\end{aligned}
\end{equation}

 Again, we we abuse of the notation and summarize constants into $C_3=C_3(L, T, g, Q^*, q^*, \delta_T)$. Therefore, after combining the Ineq.~\eqref{eq:diff_sm_fisher_add} and the estimation above, we obtain (with a fusion of constant term)
\begin{align*}
    \Big(\mathcal{J}_{\text{Diff}}(\bm{\theta})\Big)^2 &\lesssim \mathcal{J}_{\text{SM}}(\bm{\theta})\cdot \mathcal{J}_{\text{Fisher}}(\bm{\theta})
    \\ & \lesssim  \mathcal{J}_{\text{SM}}(\bm{\theta}) \cdot \Big(\widetilde{M}({\bm{\theta}})) + \sqrt{\widetilde{M}({\bm{\theta}})} +  C_3(L, T, g, Q^*, q^*, \delta_T)  \Big)
\end{align*}
\end{proof}

\textbf{Tighter bounds of Theorem~\ref{th:variant_thm4.2}. }Indeed,  we can obtain tighter upper bound from Ineq.~\eqref{eq:fisher_end_var}.  Let us revisit the argument in Ineq.~\eqref{eq:fisher_end_var}:
\begin{equation*}
    \mathbb{E}_{q_{t}(\bm{x})}\Big[\norm{\bm{s}_{\bm{\theta}} (\bm{x}, t)- \bm{s}^{\textup{ODE}}_{\bm{\theta}}(\bm{x}, t)}_2^2\Big] \nonumber
        \lesssim~  \Big(\mathbb{E}_{q_{t}(\bm{x})}\big[e_{\bm{\theta}}^2(\bm{x}) \big] + \sqrt{\mathbb{E}_{q_{t}(\bm{x})}\big[e_{\bm{\theta}}^2(\bm{x})\big]} + C_3(L, T, g, Q^*, q^*, \delta_T) \Big).\nonumber
\end{equation*}
This implies

\begin{align*}
    \mathcal{J}_{\text{Fisher}}(\bm{\theta})&=\frac{1}{2}\int_{t_0}^T g^2 (t) \mathbb{E}_{q_{t}(\bm{x})}\Big[\norm{\bm{s}_{\bm{\theta}} (\bm{x}, t)- \bm{s}^{\textup{ODE}}_{\bm{\theta}}(\bm{x}, t)}_2^2\Big]dt \nonumber
      \\&  \lesssim~  \int_{t_0}^T g^2 (t) \mathbb{E}_{q_{t}(\bm{x})}\big[e_{\bm{\theta}}^2(\bm{x}) \big]dt + \int_{t_0}^T g^2 (t) \sqrt{\mathbb{E}_{q_{t}(\bm{x})}\big[e_{\bm{\theta}}^2(\bm{x})\big]} dt + C_3(L, T, g, Q^*, q^*, \delta_T) .\nonumber
\end{align*}

By applying Cauchy-Schwartz inequality,
\begin{align*}
    \int_{t_0}^T g^2 (t) \sqrt{\mathbb{E}_{q_{t}(\bm{x})}\big[e_{\bm{\theta}}^2(\bm{x})\big]} dt 
    &=     \int_{t_0}^T \Big( g (t) \sqrt{\mathbb{E}_{q_{t}(\bm{x})}\big[e_{\bm{\theta}}^2(\bm{x})\big]} \Big) \cdot g (t) dt 
    \\&\leq \Big(\int_{t_0}^T  g^2 (t) \mathbb{E}_{q_{t}(\bm{x})}\big[e_{\bm{\theta}}^2(\bm{x})\big]   dt\Big)^{1/2} \Big(\int_{t_0}^T  g^2 (t)   dt\Big)^{1/2}.
\end{align*}
Therefore, we obtain
   \begin{align}\label{eq:ext_fisher_end_var}
    \mathcal{J}_{\text{Fisher}}(\bm{\theta})&:=\frac{1}{2}\int_{t_0}^T g^2 (t) \mathbb{E}_{q_{t}(\bm{x})}\Big[\norm{\bm{s}_{\bm{\theta}} (\bm{x}, t)- \bm{s}^{\textup{ODE}}_{\bm{\theta}}(\bm{x}, t)}_2^2\Big]dt \nonumber
      \\&  \lesssim~  \int_{t_0}^T g^2 (t) \mathbb{E}_{q_{t}(\bm{x})}\big[e_{\bm{\theta}}^2(\bm{x}) \big]dt + \norm{g}_{L^2 ([0, T])} \sqrt{\int_{t_0}^T g^2 (t) \mathbb{E}_{q_{t}(\bm{x})}\big[e_{\bm{\theta}}^2(\bm{x}) \big]dt}  + C_3(L, T, g, Q^*, q^*, \delta_T).
\end{align}
We further define 
 \begin{align*}
      \widetilde{\widetilde{M}}({\bm{\theta}}):=\int_{t_0}^T g^2 (t) \mathbb{E}_{\bm{x} \sim q_{t}(\bm{x})}\left[\int_{t_0}^{T} {q_{\tau}(\bm{x})}\norm{\bm{\epsilon}[{\bm{s_{\theta}}}](\bm{x}, \tau)}_2^2 d\tau\right]dt.
 \end{align*}
Then Ineq.~\eqref{eq:ext_fisher_end_var} can be rewritten as
   \begin{align*}
    \mathcal{J}_{\text{Fisher}}(\bm{\theta})  \lesssim  \widetilde{\widetilde{M}}({\bm{\theta}}) + \norm{g}_{L^2 ([0, T])} \sqrt{\widetilde{\widetilde{M}}({\bm{\theta}})}  + C_3(L, T, g, Q^*, q^*, \delta_T) .\nonumber
\end{align*}
Indeed, $g^2(\cdot)$ in $\widetilde{\widetilde{M}}$ motivates the choice of time-weighting function $\lambda_{\text{FP}}(\cdot)$ as $g^2(\cdot)$ in the score FPE-regularizer for the training of more complicated datasets such as CIFAR-10 and ImageNet32.

Let $g_{\textup{max}}:=\max_{t\in[t_0,T]}g^2 (t)$.
At last, we summarize a relation between all bounds $\widetilde{\widetilde{M}}({\bm{\theta}})$, $\widetilde{M}({\bm{\theta}})$, and ${M}({\bm{\theta}})$:
\begin{align*}
    \widetilde{\widetilde{M}}({\bm{\theta}}) \leq Tg_{\textup{max}} \cdot\widetilde{M}({\bm{\theta}}) \quad \textup{and} \quad  q^* M({\bm{\theta}}) \leq \widetilde{M}({\bm{\theta}}) \leq Q^* M({\bm{\theta}}).
\end{align*}

\subsection{Proof of Theorem~\ref{th:min_ode_add}}\label{subsec:pf-thm4.3}
\begin{proof}
Now we prove Ineq.~\eqref{eq:diff_sm_M_add}. As the argument of Theorem~\ref{th:min_ode}, we also start with Ineq.~\eqref{eq:diff_sm_fisher_add} and attempt to seek for its upper bound.

By rearranging Eq.~\eqref{eq:fp_error} and combining with Eq.~\eqref{eq:s_ode}, it results in 
\begin{align*}
    \bm{\epsilon}[{\bm{s_{\theta}}}](\bm{x}, t)
    &=\partial_t \bm{s}_{\bm{\theta}} - \grad{\bm{x}}\Big[\frac{1}{2}g^2(t) \div{\bm{x}}(\bm{s}_{\bm{\theta}}) - \div{\bm{x}}(\bm{f})  \Big] - \grad{\bm{x}}\Big[ \frac{1}{2}g^2(t)\norm{\bm{s}_{\bm{\theta}}}^2_2 -\inner{\bm{f}}{\bm{s}_{\bm{\theta}}}   \Big]
    \\&= \partial_t \bm{s}_{\bm{\theta}} - \partial_t \bm{s}^{\textup{ODE}}_{\bm{\theta}}
    - \grad{\bm{x}}\Big[ \frac{1}{2}g^2 (t)\inner{\bm{s}_{\bm{\theta}} - \bm{s}^{\textup{ODE}}_{\bm{\theta}}}{\bm{s}_{\bm{\theta}}}-\inner{\bm{f}}{\bm{s}_{\bm{\theta}} - \bm{s}^{\textup{ODE}}_{\bm{\theta}}}   \Big]
\end{align*}
That is, 
\begin{align*}
    \partial_t \big(\bm{s}_{\bm{\theta}} (\bm{x}, t)- \bm{s}^{\textup{ODE}}_{\bm{\theta}}(\bm{x}, t)\big)
    = \bm{\epsilon}[{\bm{s_{\theta}}}](\bm{x}, t) +  \grad{\bm{x}}\Big[ \frac{1}{2}g^2 (t)\inner{\bm{s}_{\bm{\theta}} - \bm{s}^{\textup{ODE}}_{\bm{\theta}}}{\bm{s}_{\bm{\theta}}}-\inner{\bm{f}}{\bm{s}_{\bm{\theta}} - \bm{s}^{\textup{ODE}}_{\bm{\theta}}}   \Big]
\end{align*}
Fix a $t\in[0,T]$, we integrate both sides of the above equation from $\tau=T$ to $\tau=t$  
\begin{equation}\label{eq:before_simp}
\begin{aligned}
    \bm{s}_{\bm{\theta}} (\bm{x}, t)- \bm{s}^{\textup{ODE}}_{\bm{\theta}}(\bm{x}, t)
    &=
    \bm{s}_{\bm{\theta}} (\bm{x}, T)- \bm{s}^{\textup{ODE}}_{\bm{\theta}}(\bm{x}, T)
    \\&+\displaystyle\int_{T}^{t}\bm{\epsilon}[{\bm{s_{\theta}}}](\bm{x}, \tau) d\tau + \displaystyle\int_{T}^{t} \grad{\bm{x}}\Big[ \frac{1}{2}g^2 (t)\inner{\bm{s}_{\bm{\theta}} - \bm{s}^{\textup{ODE}}_{\bm{\theta}}}{\bm{s}_{\bm{\theta}}}-\inner{\bm{f}}{\bm{s}_{\bm{\theta}} - \bm{s}^{\textup{ODE}}_{\bm{\theta}}}   \Big] d\tau.
\end{aligned}
\end{equation}

In the last term, we may compute $\grad{\bm{x}}\Big[ \frac{1}{2}g^2 (t)\inner{\bm{s}_{\bm{\theta}} - \bm{s}^{\textup{ODE}}_{\bm{\theta}}}{\bm{s}_{\bm{\theta}}}-\inner{\bm{f}}{\bm{s}_{\bm{\theta}} - \bm{s}^{\textup{ODE}}_{\bm{\theta}}}   \Big]$ as 
\begin{equation}
\begin{aligned}\label{eq:last_rearragne_add}
&\grad{\bm{x}}\Big[ \frac{1}{2}g^2 (t)\norm{\bm{s}_{\bm{\theta}}}_2-\inner{\bm{f}}{\bm{s}_{\bm{\theta}}} - \frac{1}{2}g^2 (t)\inner{\bm{s}^{\textup{ODE}}_{\bm{\theta}}}{\bm{s}_{\bm{\theta}}} + \inner{\bm{f}}{\bm{s}^{\textup{ODE}}_{\bm{\theta}}}  \Big]
\\=&\grad{\bm{x}}\Big[ \mathcal{L}[\bm{s}_{\bm{\theta}}]  + \div{\bm{x}}(\bm{f}) - \frac{1}{2}g^2 (t) \div{\bm{x}}(\bm{s}_{\bm{\theta}}) - \frac{1}{2}g^2 (t)\inner{\bm{s}^{\textup{ODE}}_{\bm{\theta}}}{\bm{s}_{\bm{\theta}}} + \inner{\bm{f}}{\bm{s}^{\textup{ODE}}_{\bm{\theta}}}  \Big]
\\=&\grad{\bm{x} } \mathcal{L}[\bm{s}_{\bm{\theta}}] + \grad{\bm{x}}\Big[ \div{\bm{x}}(\bm{f}) - \frac{1}{2}g^2 (t) \div{\bm{x}}(\bm{s}_{\bm{\theta}}) - \frac{1}{2}g^2 (t)\inner{\bm{s}^{\textup{ODE}}_{\bm{\theta}}}{\bm{s}_{\bm{\theta}}} + \inner{\bm{f}}{\bm{s}^{\textup{ODE}}_{\bm{\theta}}}  \Big]
,
\end{aligned}
\end{equation}
where $\mathcal{L}[\bm{s}_{\bm{\theta}}](\bm{x}, t):= \frac{1}{2}g^2 (t)\norm{\bm{s}_{\bm{\theta}}(\bm{x}, t)}_2-\inner{\bm{f}(\bm{x}, t)}{\bm{s}_{\bm{\theta}}(\bm{x}, t)} + \frac{1}{2}g^2 (t) \div{\bm{x}}(\bm{s}_{\bm{\theta}}(\bm{x}, t)) - \div{\bm{x}}(\bm{f}(\bm{x}, t))$. We apply the Taylor expansion at any fixed point $\bm{x}_0$ to $\grad{\bm{x}} \mathcal{L}[\bm{s}_{\bm{\theta}}]$ and get 
\begin{align}\label{eq:taylor}
    \mathcal{L}[\bm{s}_{\bm{\theta}}](\bm{x}_0, t) -  \mathcal{L}[\bm{s}_{\bm{\theta}}](\bm{x}, t) = \grad{\bm{x}} \mathcal{L}[\bm{s}_{\bm{\theta}}](\bm{x}, t) \cdot (\bm{x}_0- \bm{x}) + \mathcal{O}(\norm{\bm{x}- \bm{x}_0}_2^2). 
\end{align}
Now set $\bm{x}_0 := \bm{x} + \bm{s}_{\bm{\theta}}(\bm{x}, t) - \bm{s}^{\textup{ODE}}_{\bm{\theta}}(\bm{x}, t)$ and re-denote it as $\bm{x}_{\bm{\theta}}$. Combining Eq.~\eqref{eq:before_simp}, Eq.~\eqref{eq:last_rearragne_add}, and Eq.~\eqref{eq:taylor}, and taking the dot product with $\bm{x}_{\bm{\theta}} - \bm{x}$ from the both side of Eq.~\eqref{eq:before_simp}, we obtain

\begin{equation}
\begin{aligned}
\norm{\bm{s}_{\bm{\theta}}(\bm{x}, t) - \bm{s}^{\textup{ODE}}_{\bm{\theta}}(\bm{x}, t)}_2^2 
&\leq \abs{\inner{\bm{s}_{\bm{\theta}} (\bm{x}, T)- \bm{s}^{\textup{ODE}}_{\bm{\theta}}(\bm{x}, T)}{\bm{x}_{\bm{\theta}} - \bm{x}}} + \abs{\inner{\displaystyle\int_{T}^{t}\bm{\epsilon}[{\bm{s_{\theta}}}](\bm{x}, \tau) d\tau}{\bm{x}_{\bm{\theta}} - \bm{x}}}
\\& + \displaystyle\int_{T}^{t} \abs{\mathcal{L}[\bm{s}_{\bm{\theta}}](\bm{x}_0, \tau) -  \mathcal{L}[\bm{s}_{\bm{\theta}}](\bm{x}_{\bm{\theta}}, \tau)} d\tau + \mathcal{O}(\norm{\bm{x}_{\bm{\theta}} - \bm{x}}_2^2)
\\& + \abs{\inner{\displaystyle\int_{T}^{t} \grad{\bm{x}}\Big[ \div{\bm{x}}(\bm{f}) - \frac{1}{2}g^2 (t) \div{\bm{x}}(\bm{s}_{\bm{\theta}}) - \frac{1}{2}g^2 (t)\inner{\bm{s}^{\textup{ODE}}_{\bm{\theta}}}{\bm{s}_{\bm{\theta}}} + \inner{\bm{f}}{\bm{s}^{\textup{ODE}}_{\bm{\theta}}}  \Big]}{\bm{x}_{\bm{\theta}} - \bm{x}}}
\end{aligned}
\end{equation}

With Assumption~\ref{cond:A}~\ref{cond:s_linear}-\ref{cond:ode_linear},

\begin{equation}
\begin{aligned}
 \norm{\bm{s}_{\bm{\theta}}(\bm{x}, t) - \bm{s}^{\textup{ODE}}_{\bm{\theta}}(\bm{x}, t)}_2^2 
&\leq\norm{\bm{s}_{\bm{\theta}} (\bm{x}, T)- \bm{s}^{\textup{ODE}}_{\bm{\theta}}(\bm{x}, T)}\norm{{\bm{x}_{\bm{\theta}} - \bm{x}}} + \displaystyle\int_{0}^{T}\norm{\bm{\epsilon}[{\bm{s_{\theta}}}](\bm{x}, \tau)} d\tau \norm{\bm{x}_{\bm{\theta}} - \bm{x}}
\\& + 2 \sup_{\bm{x}} \displaystyle\int_{0}^{T} \abs{\mathcal{L}[\bm{s}_{\bm{\theta}}](\bm{x}, \tau)} d\tau + \mathcal{O}(\norm{\bm{x}_{\bm{\theta}} - \bm{x}}_2^2) + \big(1+\norm{\bm{x}}\big) \norm{\bm{x}_{\bm{\theta}} - \bm{x}}
\\& \lesssim \displaystyle\int_{0}^{T}\norm{\bm{\epsilon}[{\bm{s_{\theta}}}](\bm{x}, \tau)} d\tau \cdot \big(1+\norm{\bm{x}}\big) + \sup_{\bm{x}} \displaystyle\int_{0}^{T} \abs{\mathcal{L}[\bm{s}_{\bm{\theta}}](\bm{x}, \tau)} d\tau 
\\ &+  \big(1+\norm{\bm{x}}\big) + \big(1+\norm{\bm{x}}\big)^2
\end{aligned}
\end{equation}

Taking the expectation over $q_{t}(\bm{x})$ and applying Cauchy-Schwartz inequality, we obtain
\begin{equation}
\begin{aligned}
 \mathbb{E}_{q_{t}(\bm{x})}\big[\norm{\bm{s}_{\bm{\theta}}(\bm{x}, t) - \bm{s}^{\textup{ODE}}_{\bm{\theta}}(\bm{x}, t)}_2^2 \big] 
 &\lesssim \delta_T \mathbb{E}_{q_{t}(\bm{x})}\big[\big(1+\norm{\bm{x}}\big)\big] +
  \mathbb{E}_{q_{t}(\bm{x})}\Big[\displaystyle\int_{0}^{T}\norm{\bm{\epsilon}[{\bm{s_{\theta}}}](\bm{x}, \tau)} d\tau \Big] \cdot \mathbb{E}_{q_{t}(\bm{x})}\Big[\big(1+\norm{\bm{x}}\big)\Big] 
  \\&+ 2\sup_{\bm{x}} \displaystyle\int_{0}^{T} \abs{\mathcal{L}[\bm{s}_{\bm{\theta}}](\bm{x}, \tau)} d\tau +   \mathbb{E}_{q_{t}(\bm{x})}\big[\big(1+\norm{\bm{x}}_2\big) \big]+  \mathbb{E}_{q_{t}(\bm{x})}\big[\big(1+\norm{\bm{x}}_2\big)^2\big]
\\&\lesssim  \mathbb{E}_{q_{t}(\bm{x})}\Big[\displaystyle\int_{0}^{T}\norm{\bm{\epsilon}[{\bm{s_{\theta}}}](\bm{x}, \tau)} d\tau \Big] + \sup_{\bm{x}} \displaystyle\int_{0}^{T} \abs{\mathcal{L}[\bm{s}_{\bm{\theta}}](\bm{x}, \tau)} d\tau  + C_2(L, T, \delta_T, g).
\end{aligned}
\end{equation}
\end{proof}

\subsection{Proof and discussion of Proposition~\ref{th:conservativity}}\label{subsec:proof-conservativity}

\begin{proof}
Integrating the following equation w.r.t. time from $\tau=t_{\bm{\theta}}$ to $\tau=t$ with $t\in[0,T]$ fixed,
\begin{equation*}
    \partial_t \bm{s}_{\bm{\theta}}  = \grad{\bm{x}}\Big[\frac{1}{2}g^2(t) \div{\bm{x}}(\bm{s}_{\bm{\theta}}) + \frac{1}{2}g^2(t)\norm{\bm{s}_{\bm{\theta}}}^2_2 -\inner{\bm{f}}{\bm{s}_{\bm{\theta}}} - \div{\bm{x}}(\bm{f})  \Big] + \bm{\epsilon}[{\bm{s_{\theta}}}](\bm{x}, t),
\end{equation*}
leads to 
\begin{align*}
    \bm{s}_{\bm{\theta}}(\bm{x}, t) -  \bm{s}_{\bm{\theta}}(\bm{x}, t_{\bm{\theta}})  = \grad{\bm{x}}\Big\{\displaystyle\int_{t_{\bm{\theta}}}^{t} \Big[\frac{1}{2}g^2(t) \div{\bm{x}}(\bm{s}_{\bm{\theta}}) + \frac{1}{2}g^2(t)\norm{\bm{s}_{\bm{\theta}}}^2_2 -\inner{\bm{f}}{\bm{s}_{\bm{\theta}}} - &\div{\bm{x}}(\bm{f})  \Big] d\tau\Big\} \\
    &+ \displaystyle\int_{t_{\bm{\theta}}}^{t} \bm{\epsilon}[{\bm{s_{\theta}}}](\bm{x}, t)d\tau,
\end{align*}
where the swap of integration and differentiation is valid if the integrand is sufficiently smooth.

With the assumption, we obtain that for all $t\in[0,T]$
\begin{align*}
    \bm{s}_{\bm{\theta}}(\bm{x}, t) - \grad{\bm{x}}\Big\{ \log q_{t_{\bm{\theta}}}(\bm{x}) + \displaystyle\int_{t_{\bm{\theta}}}^{t} \Big[\frac{1}{2}g^2(t) \div{\bm{x}}(\bm{s}_{\bm{\theta}}) + \frac{1}{2}g^2(t)\norm{\bm{s}_{\bm{\theta}}}^2_2 -\inner{\bm{f}}{\bm{s}_{\bm{\theta}}} - &\div{\bm{x}}(\bm{f})  \Big] d\tau \Big\} \\ 
    &= \displaystyle\int_{t_{\bm{\theta}}}^{t}  \bm{\epsilon}[{\bm{s_{\theta}}}](\bm{x}, \tau)d\tau.
\end{align*}
We let $\Psi_{\bm{\theta}}(\bm{x}, t)= \log q_{t_{\bm{\theta}}}(\bm{x})+ \int_{t_{\bm{\theta}}}^{t} \Big[\frac{1}{2}g^2(\tau) \div{\bm{x}}(\bm{s}_{\bm{\theta}}) + \frac{1}{2}g^2(\tau)\norm{\bm{s}_{\bm{\theta}}}^2_2 -\inner{\bm{f}}{\bm{s}_{\bm{\theta}}} - \div{\bm{x}}(\bm{f})  \Big]d\tau$. By taking the norm of the above equation, one can obtain
\begin{align*}
    \norm{\bm{s}_{\bm{\theta}}(\bm{x}, t) - \grad{\bm{x}} \Psi_{\bm{\theta}}(\bm{x}, t)}_2 = \norm{\displaystyle\int_{t_{\bm{\theta}}}^{t}  \bm{\epsilon}[{\bm{s_{\theta}}}](\bm{x}, \tau)d\tau}_2.
\end{align*}
From which we obtain
\begin{align*}
    \norm{\bm{s}_{\bm{\theta}}(\bm{x}, t) - \grad{\bm{x}} \Psi_{\bm{\theta}}(\bm{x}, t)}_2 = \norm{\displaystyle\int_{t_{\bm{\theta}}}^{t}  \bm{\epsilon}[{\bm{s_{\theta}}}](\bm{x}, \tau)d\tau}_2
    \leq \left|\displaystyle\int_{t_{\bm{\theta}}}^{t}  \norm{\bm{\epsilon}[{\bm{s_{\theta}}}](\bm{x}, \tau)}_2d\tau\right|.
\end{align*}
Hence, the proposition is proved.
\end{proof}

Proposition~\ref{th:conservativity} requests a perfect match of scores at some single timestep $t_{\bm{\theta}}\in[0,T]$: $\bm{s}_{\bm{\theta}}(\bm{x}, t_{\bm{\theta}}) =\nabla_{\bm{x}} \log q_{t_{\bm{\theta}}}(\bm{x})$ for all $\bm{x}$. However, we can involve an error term when the scores are not matched exactly and formulate an extended version of Proposition~\ref{th:conservativity} as the following.
\begin{proposition}\label{th:revised_conservativity}
Suppose that there is a constant $\delta>0$ so that for any $\bm{\theta}$, there is a single timestep $t_{\bm{\theta}}\in[0,T]$ such that 
$\norm{\bm{s}_{\bm{\theta}}(\bm{x}, t_{\bm{\theta}}) -\nabla_{\bm{x}} \log q_{t_{\bm{\theta}}}(\bm{x})}_2\leq \delta$, then there exists a real-valued function $\Psi_{\bm{\theta}}\colon \mathbb{R}^D \times [0, T] \rightarrow \mathbb{R}$ (with an explicit expression) that satisfies 
\begin{equation*}
    \norm{\bm{s}_{\bm{\theta}}(\bm{x}, t) - \grad{\bm{x}} \Psi_{\bm{\theta}}(\bm{x}, t)}_2 \leq \delta + \abs{\int_{t}^{t_{\bm{\theta}}} \norm{\bm{\epsilon}[{\bm{s_{\theta}}}](\bm{x}, \tau)}_2 d\tau}.
\end{equation*}
\end{proposition}

\begin{proof} The proof is almost identical to the original one. We start with
\begin{align*}
    \bm{s}_{\bm{\theta}}(\bm{x}, t) -  \bm{s}_{\bm{\theta}}(\bm{x}, t_{\bm{\theta}})  = \grad{\bm{x}}\Big\{\displaystyle\int_{t_{\bm{\theta}}}^{t} \Big[\frac{1}{2}g^2(t) \div{\bm{x}}(\bm{s}_{\bm{\theta}}) + \frac{1}{2}g^2(t)\norm{\bm{s}_{\bm{\theta}}}^2_2 -\inner{\bm{f}}{\bm{s}_{\bm{\theta}}} - &\div{\bm{x}}(\bm{f})  \Big] d\tau\Big\} \\
    &+ \displaystyle\int_{t_{\bm{\theta}}}^{t} \bm{\epsilon}[{\bm{s_{\theta}}}](\bm{x}, t)d\tau.
\end{align*}
By inserting the term $\grad{\bm{x}} \log q_{t_{\bm{\theta}}}(\bm{x})$, we have
\begin{align*}
    \bm{s}_{\bm{\theta}}(\bm{x}, t) - \grad{\bm{x}}\Big\{ \log q_{t_{\bm{\theta}}}(\bm{x}) + \displaystyle\int_{t_{\bm{\theta}}}^{t} \Big[\frac{1}{2}g^2(t) \div{\bm{x}}(\bm{s}_{\bm{\theta}}) &+ \frac{1}{2}g^2(t)\norm{\bm{s}_{\bm{\theta}}}^2_2 -\inner{\bm{f}}{\bm{s}_{\bm{\theta}}} - \div{\bm{x}}(\bm{f})  \Big] d\tau \Big\} \\ 
    &= \big( \bm{s}_{\bm{\theta}}(\bm{x}, t_{\bm{\theta}}) - \grad{\bm{x}} \log q_{t_{\bm{\theta}}}(\bm{x})  \big) + \displaystyle\int_{t_{\bm{\theta}}}^{t}  \bm{\epsilon}[{\bm{s_{\theta}}}](\bm{x}, \tau)d\tau.
\end{align*}
We now let $\Psi_{\bm{\theta}}(\bm{x}, t)= \log q_{t_{\bm{\theta}}}(\bm{x})+ \int_{t_{\bm{\theta}}}^{t} \Big[\frac{1}{2}g^2(\tau) \div{\bm{x}}(\bm{s}_{\bm{\theta}}) + \frac{1}{2}g^2(\tau)\norm{\bm{s}_{\bm{\theta}}}^2_2 -\inner{\bm{f}}{\bm{s}_{\bm{\theta}}} - \div{\bm{x}}(\bm{f})  \Big]d\tau$. By taking the norm of the above equation, we establish the claim.

\end{proof}

\subsection{Proof of Proposition~\ref{th:conti_strong}}
\label{subsec:pf-prop4.5}
\begin{lemma}\label{th:lemma_psde}
Let $\bm{s}_{\bm{\theta}}$ be a score obtained from denoising score matching (Eq.~\eqref{eq:dsm}) and write $\bm{s}^{\textup{SDE}}_{\bm{\theta}}(\cdot, t):=\grad{\bm{x}}\log p_{t,\bm{\theta}}^{\text{SDE}}$. Then
\begin{enumerate}
    \item\citep{lu2022maximum} Eq.~\eqref{eq:sde_backward_sub} associates with the following forward SDE whose marginal density is $\bm{s}^{\textup{SDE}}_{\bm{\theta}}$:
    \begin{equation*}
        d\bm{x}_{\bm{\theta}}(t) = \Big[\bm{f}(\bm{x}_{\bm{\theta}}(t), t) + g^2(t)\big(\bm{s}^{\textup{SDE}}_{\bm{\theta}}(\bm{x}_{\bm{\theta}}(t), t) - \bm{s}_{\bm{\theta}}(\bm{x}_{\bm{\theta}}(t), t)  \big)\Big] dt + g(t) \bm{w}_t
    \end{equation*}
    \item $\bm{s}^{\textup{SDE}}_{\bm{\theta}}$ satisfies the following score FPE:
    \begin{equation}\label{eq:psde_fp}
        \partial_t \bm{s}^{\textup{SDE}}_{\bm{\theta}} -\grad{\bm{x}} \Big[\frac{1}{2}g^2(t) \div{\bm{x}}\big(2\bm{s}_{\bm{\theta}} - \bm{s}^{\textup{SDE}}_{\bm{\theta}}\big) + \frac{1}{2}g^2(t) \big( 2\inner{\bm{s}_{\bm{\theta}}}{\bm{s}^{\textup{SDE}}_{\bm{\theta}}} -\norm{\bm{s}^{\textup{SDE}}_{\bm{\theta}}}_2^2\big) - \inner{\bm{f}}{\bm{s}^{\textup{SDE}}_{\bm{\theta}}} - \div{\bm{x}}(\bm{f})\Big] = 0.
    \end{equation}
\end{enumerate}
\end{lemma}

\begin{proof}{Lemma}~{\ref{th:lemma_psde}}

The proof of the first statement can be found in \citep{lu2022maximum}. We now prove the second statement.

Consider 
    \begin{equation*}
        \bm{F}(\bm{x}, t):= \bm{f}(\bm{x}, t) + g^2(t)(\bm{s}^{\textup{SDE}}_{\bm{\theta}} - \bm{s}_{\bm{\theta}}) \quad \textup{ and } \quad \bm{G}(\bm{x}, t):= g(t)\bm{I}    
    \end{equation*}
in Eq.~\eqref{eq:general_sde}, and apply Proposition~\ref{th:fp}, the lemma is then established.   
\end{proof}

\begin{proof}{Proposition}~{\ref{th:conti_strong}}

We recall Eq.~\eqref{eq:fp_error}, which indicates

\begin{equation}\label{eq:dsm_fp}
    \partial_t \bm{s}_{\bm{\theta}} - \grad{\bm{x}}\Big[\frac{1}{2}g^2(t) \div{\bm{x}}(\bm{s}_{\bm{\theta}}) + \frac{1}{2}g^2(t)\norm{\bm{s}_{\bm{\theta}}}^2_2 -\inner{\bm{f}}{\bm{s}_{\bm{\theta}}} - \div{\bm{x}}(\bm{f})  \Big] - \bm{\epsilon}[{\bm{s_{\theta}}}]=0.
\end{equation}
First, we subtract Eq.~\eqref{eq:psde_fp} by the above equation and get
\begin{equation}\label{eq:psde_diff}
    \partial_t (\bm{s}^{\textup{SDE}}_{\bm{\theta}} - \bm{s}_{\bm{\theta}}) -\grad{\bm{x}} \Big[\frac{1}{2}g^2(t) \div{\bm{x}}(\bm{s}_{\bm{\theta}} - \bm{s}^{\textup{SDE}}_{\bm{\theta}} )  - \frac{1}{2}g^2(t) \norm{\bm{s}_{\bm{\theta}} - \bm{s}^{\textup{SDE}}_{\bm{\theta}} }_2^2 - \inner{\bm{f}}{\bm{s}_{\bm{\theta}} - \bm{s}^{\textup{SDE}}_{\bm{\theta}} } \Big] + \bm{\epsilon}[{\bm{s_{\theta}}}]= 0.
\end{equation}
Consider when $\bm{\theta} = \bm{\theta}_0$ and let $\bm{u}_{\bm{\theta}_0}:=\bm{s}^{\textup{SDE}}_{\bm{\theta}_0} - \bm{s}_{\bm{\theta}_0}$. Then the PDEs become 
\begin{equation*}
    \partial_t \bm{u}_{\bm{\theta}_0} + \grad{\bm{x}} \Big[\frac{1}{2}g^2(t) \div{\bm{x}}(\bm{u}_{\bm{\theta}_0} )  + \frac{1}{2}g^2(t) \norm{\bm{u}_{\bm{\theta}_0} }_2^2 + \inner{\bm{f}}{\bm{u}_{\bm{\theta}_0} } \Big]  = 0.
\end{equation*}
Here, $\bm{u}_{\bm{\theta}_0}$ is a solution to the PDEs. It is noticed that this system of PDEs has a zero initial condition and zero boundary condition as both $\bm{s}_{\bm{\theta}_0}$ and $ \bm{s}^{\textup{SDE}}_{\bm{\theta}_0}$ share the same initial/boundary condition. Thus, from the assumption of the uniqueness of solution, we know that $\bm{u}_{\bm{\theta}_0}\equiv \bm{0}$, and hence, $\bm{s}^{\textup{SDE}}_{\bm{\theta}_0} \equiv \bm{s}_{\bm{\theta}_0}$.

We repeat the same trick to subtract Eq.~\eqref{eq:fp_gt} by Eq.~\eqref{eq:dsm_fp} from which we can obtain $\bm{s}_{\bm{\theta}_0}\equiv\bm{s}$. Similarly, the same argument can be applied to Eq.~\eqref{eq:diff_ode_nnscore} to prove $\bm{s}^{\textup{ODE}}_{\bm{\theta}_0}\equiv \bm{s}_{\bm{\theta}_0}$.

\end{proof}

\subsection{Proof of Proposition~\ref{th:higher}}\label{subsec:pf-prop4.6}
\begin{proof}
By subtracting the following two equations 
\begin{align*}
    \partial_t \bm{s}_{\bm{\theta}}  &= \grad{\bm{x}}\Big[\frac{1}{2}g^2(t) \div{\bm{x}}(\bm{s}_{\bm{\theta}}) + \frac{1}{2}g^2(t)\norm{\bm{s}_{\bm{\theta}}}^2_2 -\inner{\bm{f}}{\bm{s}_{\bm{\theta}}} - \div{\bm{x}}(\bm{f})  \Big] + \bm{\epsilon}[{\bm{s_{\theta}}}]
    \\
    \partial_t \bm{s} &= \grad{\bm{x}}\Big[\frac{1}{2}g^2(t) \div{\bm{x}}(\bm{s}) + \frac{1}{2}g^2(t)\norm{\bm{s}}^2_2 -\inner{\bm{f}}{\bm{s}} - \div{\bm{x}}(\bm{f})  \Big],
\end{align*}
we obtain 
\begin{align*}
    \partial_t (\bm{s}_{\bm{\theta}} - \bm{s}  )&= \grad{\bm{x}}\Big[\frac{1}{2}g^2(t) \div{\bm{x}}(\bm{s}_{\bm{\theta}} - \bm{s}  ) + \frac{1}{2}g^2(t) \big( \norm{\bm{s}_{\bm{\theta}}}^2_2 - \norm{\bm{s}}^2_2 \big) -\inner{\bm{f}}{\bm{s}_{\bm{\theta}} - \bm{s} }   \Big] + \bm{\epsilon}[{\bm{s_{\theta}}}]
\end{align*}
Notice that $\norm{\bm{s}_{\bm{\theta}}}^2_2 - \norm{\bm{s}}^2_2 =\norm{\bm{s}_{\bm{\theta}}-\bm{s}}^2_2 +2\inner{\bm{s}_{\bm{\theta}}-\bm{s}}{\bm{s}}$. Integrating over time from $\tau=0$ to $\tau=t$, we obtain

\begin{align*}
\displaystyle\int_{0}^{t}  \bm{\epsilon}[{\bm{s_{\theta}}}](\bm{x}, \tau)d\tau &= \big(\bm{s}_{\bm{\theta}}(\bm{x}, t) - \bm{s}(\bm{x}, t)\big) - \big(\bm{s}_{\bm{\theta}}(\bm{x}, 0) - \bm{s}(\bm{x}, 0) \big) \\
&- \displaystyle\int_{0}^{t} \frac{1}{2}g^2(\tau) \grad{\bm{x}}\div{\bm{x}}(\bm{s}_{\bm{\theta}}-\bm{s}) d\tau\\
&- \displaystyle\int_{0}^{t} g^2(\tau)\Big[\inner{\grad{\bm{x}}(\bm{s}_{\bm{\theta}}-\bm{s})}{\bm{s}_{\bm{\theta}}-\bm{s}} + \inner{\grad{\bm{x}}(\bm{s}_{\bm{\theta}}-\bm{s})}{\bm{s}} +\inner{\bm{s}_{\bm{\theta}}-\bm{s}}{\grad{\bm{x}}\bm{s}}\Big]d\tau\\
&+\displaystyle\int_{0}^{t}\Big[\inner{\grad{\bm{x}}\bm{f}}{\bm{s}_{\bm{\theta}}-\bm{s} } +\inner{\bm{f}}{\grad{\bm{x}}(\bm{s}_{\bm{\theta}}-\bm{s}) }\Big]d\tau
\end{align*}


By applying the $\ell_2$-norm and Cauchy-Schwartz inequality while noting the relation $\norm{A}_2 \leq \norm{A}_F$ for a general square matrix $A$, the statement is proved.

\end{proof}

\subsection{Proof of Proposition~\ref{th:error_analysis}}\label{subsec:proof-error}
\begin{proof} For any $\delta$ which is small enough, by \citep{skorski2021modern}, we have

\begin{align}\label{eq:bd_hut}
\mathbb{P}\big(\norm{\mathscr{H}[\bm{s}]}_D \geq \delta \big) \leq \exp^{-\Big(\frac{\frac{4M}{g^*} \delta^2}{2(1-\frac{16}{3g^*})\delta}  \Big)}. 
\end{align}
On the other hand, Lemma~\ref{th:finite_diff} implies that there is a $C>0$ so that $\norm{\mathscr{F}[\bm{s}]}_D < CD h$, where $h=\frac{h_s^2 h_d + h_d h_s^2}{h_s + h_d}$ and $\norm{\cdot}_D$ indicates the $\ell_D$-norm. Hence, we have
\begin{align}\label{eq:bd_fd}
\mathbb{P}\big(\norm{\mathscr{F}[\bm{s}]}_D \geq CDh \big) \leq 0. 
\end{align}
Now rearranging, we have 
\begin{align*}
\mathscr{E}[\bm{s}]= \big(\partial_t\bm{s}-\textup{FD}(\bm{s}) \big) - \big(\frac{1}{2}g^2\textup{tr}(\nabla\bm{s}) - \frac{1}{2}g^2\textup{tr}_{H^{(M)}}(\nabla\bm{s}) \big)=\mathscr{T}[\bm{s}] - \mathscr{H}[\bm{s}].
\end{align*}
With the statistical bounds \eqref{eq:bd_hut} and \eqref{eq:bd_fd}, we obtain
\begin{align*}
    \mathbb{P}\big(\norm{\mathscr{E}[\bm{s}]}_D \geq \delta + CDh \big) \leq \mathbb{P}\big(\norm{\mathscr{F}[\bm{s}]}_D \geq \delta \big) + \mathbb{P}\big(\norm{\mathscr{H}[\bm{s}]}_D \geq CDh \big)  \leq \exp^{-\Big(\frac{\frac{4M}{g^*} \delta^2}{2(1-\frac{16}{3g^*})\delta}  \Big)}
\end{align*}
If $h$ and $\epsilon$ is selected small enough (as the assumption), by taking $\delta = \epsilon - CDh$, we observe that $\epsilon > \delta >\frac{\epsilon}{2}$ and that 
\begin{align*}
    \exp^{-\Big(\frac{4M \epsilon^2}{2(g^* - \frac{16}{3}\epsilon)}  \Big)}\leq  \exp^{-\Big(\frac{\frac{4M}{g^*} \delta^2}{2(1-\frac{16}{3g^*})\delta}  \Big)}   \leq \exp^{-\Big(\frac{M \epsilon^2}{2(g^* - \frac{8}{3}\epsilon)}  \Big)}.
\end{align*}
Thus, the claimed error bound is established.

\end{proof}

\end{document}

%% file: math_commands_icml.tex
\usepackage{graphicx}
\usepackage{wrapfig}
\usepackage[para,online,flushleft]{threeparttable}
\usepackage{algorithm,algorithmic}
\usepackage{hyperref}
\usepackage{xurl}
\usepackage{verbatim}
\usepackage{varwidth}
\usepackage{threeparttable}
\usepackage{tabularx}
\usepackage{multirow}
\usepackage{amsmath,amsfonts,bm,amssymb,amsthm}
\usepackage{mathrsfs}  
\usepackage{array,arydshln}
\usepackage{threeparttable}
\usepackage{graphicx}
\usepackage[super]{nth}
\usepackage{booktabs}
\usepackage{xcolor}
\usepackage[normalem]{ulem}
\usepackage{caption}
\usepackage{lipsum}

\newcommand{\inner}[2]{\langle#1,#2\rangle}

\newcommand{\grad}[1]{{\nabla_{\bm{#1}}}}
\renewcommand{\div}[1]{{\textup{div}_{\bm{#1}}}}

\def\eqref#1{(\ref{#1})}

\def\1{\bm{1}}

\DeclareMathAlphabet{\mathsfit}{\encodingdefault}{\sfdefault}{m}{sl}
\SetMathAlphabet{\mathsfit}{bold}{\encodingdefault}{\sfdefault}{bx}{n}

\newcommand{\norm}[1]{\left\lVert#1\right\rVert}
\newcommand{\abs}[1]{\left|#1\right|}

\definecolor{ube}{rgb}{0.53, 0.47, 0.76}
\definecolor{ballblue}{rgb}{0.13, 0.67, 0.8}

\definecolor{spink}{rgb}{0.831,0.184,0.494}

%% file: main_icml23.bbl
\begin{thebibliography}{46}
\providecommand{\natexlab}[1]{#1}
\providecommand{\url}[1]{\texttt{#1}}
\expandafter\ifx\csname urlstyle\endcsname\relax
  \providecommand{\doi}[1]{doi: #1}\else
  \providecommand{\doi}{doi: \begingroup \urlstyle{rm}\Url}\fi

\bibitem[Anderson(1982)]{anderson1982reverse}
Anderson, B.~D.
\newblock Reverse-time diffusion equation models.
\newblock \emph{Stochastic Processes and their Applications}, 12\penalty0
  (3):\penalty0 313--326, 1982.

\bibitem[Artstein(1975)]{artstein1975continuous}
Artstein, Z.
\newblock Continuous dependence on parameters: On the best possible results.
\newblock \emph{Journal of Differential Equations}, 19\penalty0 (2):\penalty0
  214--225, 1975.

\bibitem[Blechschmidt \& Ernst(2021)Blechschmidt and
  Ernst]{blechschmidt2021three}
Blechschmidt, J. and Ernst, O.~G.
\newblock Three ways to solve partial differential equations with neural
  networks—a review.
\newblock \emph{GAMM-Mitteilungen}, 44\penalty0 (2):\penalty0 e202100006, 2021.

\bibitem[Chao et~al.(2022)Chao, Sun, Cheng, and Lee]{chao2022quasi}
Chao, C.-H., Sun, W.-F., Cheng, B.-W., and Lee, C.-Y.
\newblock Quasi-conservative score-based generative models.
\newblock \emph{arXiv preprint arXiv:2209.12753}, 2022.

\bibitem[Chen et~al.(2022)Chen, Lee, and Lu]{chen2022improved}
Chen, H., Lee, H., and Lu, J.
\newblock Improved analysis of score-based generative modeling: User-friendly
  bounds under minimal smoothness assumptions.
\newblock \emph{arXiv preprint arXiv:2211.01916}, 2022.

\bibitem[Chen et~al.(2018)Chen, Rubanova, Bettencourt, and
  Duvenaud]{chen2018neural}
Chen, R.~T., Rubanova, Y., Bettencourt, J., and Duvenaud, D.~K.
\newblock Neural ordinary differential equations.
\newblock \emph{Advances in neural information processing systems}, 31, 2018.

\bibitem[Cheuk et~al.(2022)Cheuk, Sawata, Uesaka, Murata, Takahashi, Takahashi,
  Herremans, and Mitsufuji]{cheuk2022diffroll}
Cheuk, K.~W., Sawata, R., Uesaka, T., Murata, N., Takahashi, N., Takahashi, S.,
  Herremans, D., and Mitsufuji, Y.
\newblock Diffroll: Diffusion-based generative music transcription with
  unsupervised pretraining capability.
\newblock \emph{arXiv preprint arXiv:2210.05148}, 2022.

\bibitem[Chrabaszcz et~al.(2017)Chrabaszcz, Loshchilov, and
  Hutter]{chrabaszcz2017downsampled}
Chrabaszcz, P., Loshchilov, I., and Hutter, F.
\newblock A downsampled variant of imagenet as an alternative to the cifar
  datasets.
\newblock \emph{arXiv preprint arXiv:1707.08819}, 2017.

\bibitem[De~Bortoli et~al.(2021)De~Bortoli, Thornton, Heng, and
  Doucet]{de2021diffusion}
De~Bortoli, V., Thornton, J., Heng, J., and Doucet, A.
\newblock Diffusion schr{\"o}dinger bridge with applications to score-based
  generative modeling.
\newblock \emph{Advances in Neural Information Processing Systems},
  34:\penalty0 17695--17709, 2021.

\bibitem[Dhariwal \& Nichol(2021)Dhariwal and Nichol]{dhariwal2021diffusion}
Dhariwal, P. and Nichol, A.
\newblock Diffusion models beat gans on image synthesis.
\newblock \emph{Advances in Neural Information Processing Systems},
  34:\penalty0 8780--8794, 2021.

\bibitem[Evans \& Garzepy(2018)Evans and Garzepy]{evans2018measure}
Evans, L.~C. and Garzepy, R.~F.
\newblock \emph{Measure theory and fine properties of functions}.
\newblock Routledge, 2018.

\bibitem[Fokker(1914)]{fokker1914mittlere}
Fokker, A.~D.
\newblock Die mittlere energie rotierender elektrischer dipole im
  strahlungsfeld.
\newblock \emph{Annalen der Physik}, 348\penalty0 (5):\penalty0 810--820, 1914.

\bibitem[Fornberg(1988)]{fornberg1988generation}
Fornberg, B.
\newblock Generation of finite difference formulas on arbitrarily spaced grids.
\newblock \emph{Mathematics of computation}, 51\penalty0 (184):\penalty0
  699--706, 1988.

\bibitem[Gronwall(1919)]{gronwall1919note}
Gronwall, T.~H.
\newblock Note on the derivatives with respect to a parameter of the solutions
  of a system of differential equations.
\newblock \emph{Annals of Mathematics}, pp.\  292--296, 1919.

\bibitem[Ho et~al.(2019)Ho, Chen, Srinivas, Duan, and Abbeel]{ho2019flow++}
Ho, J., Chen, X., Srinivas, A., Duan, Y., and Abbeel, P.
\newblock Flow++: Improving flow-based generative models with variational
  dequantization and architecture design.
\newblock In \emph{International Conference on Machine Learning}, pp.\
  2722--2730. PMLR, 2019.

\bibitem[Ho et~al.(2020)Ho, Jain, and Abbeel]{ho2020denoising}
Ho, J., Jain, A., and Abbeel, P.
\newblock Denoising diffusion probabilistic models.
\newblock \emph{Advances in Neural Information Processing Systems},
  33:\penalty0 6840--6851, 2020.

\bibitem[Hutchinson(1989)]{hutchinson1989stochastic}
Hutchinson, M.~F.
\newblock A stochastic estimator of the trace of the influence matrix for
  laplacian smoothing splines.
\newblock \emph{Communications in Statistics-Simulation and Computation},
  18\penalty0 (3):\penalty0 1059--1076, 1989.

\bibitem[Hyv{\"a}rinen \& Dayan(2005)Hyv{\"a}rinen and
  Dayan]{hyvarinen2005estimation}
Hyv{\"a}rinen, A. and Dayan, P.
\newblock Estimation of non-normalized statistical models by score matching.
\newblock \emph{Journal of Machine Learning Research}, 6\penalty0 (4), 2005.

\bibitem[Kawar et~al.(2022)Kawar, Elad, Ermon, and Song]{kawar2022denoising}
Kawar, B., Elad, M., Ermon, S., and Song, J.
\newblock Denoising diffusion restoration models.
\newblock \emph{arXiv preprint arXiv:2201.11793}, 2022.

\bibitem[Kim et~al.(2022)Kim, Kim, Kang, and Moon]{kim2022refining}
Kim, D., Kim, Y., Kang, W., and Moon, I.-C.
\newblock Refining generative process with discriminator guidance in
  score-based diffusion models.
\newblock \emph{arXiv preprint arXiv:2211.17091}, 2022.

\bibitem[Kong et~al.(2020)Kong, Ping, Huang, Zhao, and
  Catanzaro]{kong2020diffwave}
Kong, Z., Ping, W., Huang, J., Zhao, K., and Catanzaro, B.
\newblock Diffwave: A versatile diffusion model for audio synthesis.
\newblock \emph{arXiv preprint arXiv:2009.09761}, 2020.

\bibitem[Kwon et~al.(2022)Kwon, Fan, and Lee]{kwon2022score}
Kwon, D., Fan, Y., and Lee, K.
\newblock Score-based generative modeling secretly minimizes the wasserstein
  distance.
\newblock \emph{arXiv preprint arXiv:2212.06359}, 2022.

\bibitem[Lu et~al.(2022)Lu, Zheng, Bao, Chen, Li, and Zhu]{lu2022maximum}
Lu, C., Zheng, K., Bao, F., Chen, J., Li, C., and Zhu, J.
\newblock Maximum likelihood training for score-based diffusion odes by high
  order denoising score matching.
\newblock In \emph{International Conference on Machine Learning}, pp.\
  14429--14460. PMLR, 2022.

\bibitem[Lunardi(2012)]{lunardi2012analytic}
Lunardi, A.
\newblock \emph{Analytic semigroups and optimal regularity in parabolic
  problems}.
\newblock Springer Science \& Business Media, 2012.

\bibitem[Masry \& Rice(1992)Masry and Rice]{masry1992gaussian}
Masry, E. and Rice, J.~A.
\newblock Gaussian deconvolution via differentiation.
\newblock \emph{Canadian Journal of Statistics}, 20\penalty0 (1):\penalty0
  9--21, 1992.

\bibitem[Meng et~al.(2021{\natexlab{a}})Meng, Song, Li, and
  Ermon]{meng2021estimating}
Meng, C., Song, Y., Li, W., and Ermon, S.
\newblock Estimating high order gradients of the data distribution by
  denoising.
\newblock \emph{Advances in Neural Information Processing Systems},
  34:\penalty0 25359--25369, 2021{\natexlab{a}}.

\bibitem[Meng et~al.(2021{\natexlab{b}})Meng, Song, Song, Wu, Zhu, and
  Ermon]{meng2021sdedit}
Meng, C., Song, Y., Song, J., Wu, J., Zhu, J.-Y., and Ermon, S.
\newblock Sdedit: Image synthesis and editing with stochastic differential
  equations.
\newblock \emph{arXiv preprint arXiv:2108.01073}, 2021{\natexlab{b}}.

\bibitem[Murata et~al.(2023)Murata, Saito, Lai, Takida, Uesaka, Mitsufuji, and
  Ermon]{murata2023gibbsddrm}
Murata, N., Saito, K., Lai, C.-H., Takida, Y., Uesaka, T., Mitsufuji, Y., and
  Ermon, S.
\newblock Gibbsddrm: A partially collapsed gibbs sampler for solving blind
  inverse problems with denoising diffusion restoration, 2023.

\bibitem[{\O}ksendal(2003)]{oksendal2003stochastic}
{\O}ksendal, B.
\newblock Stochastic differential equations.
\newblock In \emph{Stochastic differential equations}, pp.\  65--84. Springer,
  2003.

\bibitem[Papageorgiou(1994)]{papageorgiou1994solution}
Papageorgiou, N.
\newblock On the solution set of nonlinear evolution inclusions depending on a
  parameter.
\newblock 1994.

\bibitem[Pidstrigach(2022)]{pidstrigach2022score}
Pidstrigach, J.
\newblock Score-based generative models detect manifolds.
\newblock \emph{arXiv preprint arXiv:2206.01018}, 2022.

\bibitem[Planck(1917)]{planck1917satz}
Planck, V.
\newblock {\"U}ber einen satz der statistischen dynamik und seine erweiterung
  in der quantentheorie.
\newblock \emph{Sitzungberichte der}, 1917.

\bibitem[Raissi et~al.(2019)Raissi, Perdikaris, and
  Karniadakis]{raissi2019physics}
Raissi, M., Perdikaris, P., and Karniadakis, G.~E.
\newblock Physics-informed neural networks: A deep learning framework for
  solving forward and inverse problems involving nonlinear partial differential
  equations.
\newblock \emph{Journal of Computational physics}, 378:\penalty0 686--707,
  2019.

\bibitem[Rombach et~al.(2022)Rombach, Blattmann, Lorenz, Esser, and
  Ommer]{rombach2022high}
Rombach, R., Blattmann, A., Lorenz, D., Esser, P., and Ommer, B.
\newblock High-resolution image synthesis with latent diffusion models.
\newblock In \emph{Proceedings of the IEEE/CVF Conference on Computer Vision
  and Pattern Recognition}, pp.\  10684--10695, 2022.

\bibitem[Roosta-Khorasani \& Ascher(2015)Roosta-Khorasani and
  Ascher]{roosta2015improved}
Roosta-Khorasani, F. and Ascher, U.
\newblock Improved bounds on sample size for implicit matrix trace estimators.
\newblock \emph{Foundations of Computational Mathematics}, 15\penalty0
  (5):\penalty0 1187--1212, 2015.

\bibitem[Saharia et~al.(2022)Saharia, Chan, Saxena, Li, Whang, Denton,
  Ghasemipour, Ayan, Mahdavi, Lopes, et~al.]{saharia2022photorealistic}
Saharia, C., Chan, W., Saxena, S., Li, L., Whang, J., Denton, E., Ghasemipour,
  S. K.~S., Ayan, B.~K., Mahdavi, S.~S., Lopes, R.~G., et~al.
\newblock Photorealistic text-to-image diffusion models with deep language
  understanding.
\newblock \emph{arXiv preprint arXiv:2205.11487}, 2022.

\bibitem[Saito et~al.(2022)Saito, Murata, Uesaka, Lai, Takida, Fukui, and
  Mitsufuji]{saito2022unsupervised}
Saito, K., Murata, N., Uesaka, T., Lai, C.-H., Takida, Y., Fukui, T., and
  Mitsufuji, Y.
\newblock Unsupervised vocal dereverberation with diffusion-based generative
  models.
\newblock \emph{arXiv preprint arXiv:2211.04124}, 2022.

\bibitem[Salimans \& Ho(2021)Salimans and Ho]{salimans2021should}
Salimans, T. and Ho, J.
\newblock Should ebms model the energy or the score?
\newblock In \emph{Energy Based Models Workshop-ICLR 2021}, 2021.

\bibitem[Shen et~al.(2022)Shen, Wang, Kale, Ribeiro, Karbasi, and
  Hassani]{shen2022self}
Shen, Z., Wang, Z., Kale, S., Ribeiro, A., Karbasi, A., and Hassani, H.
\newblock Self-consistency of the fokker planck equation.
\newblock In \emph{Conference on Learning Theory}, pp.\  817--841. PMLR, 2022.

\bibitem[Skorski(2021)]{skorski2021modern}
Skorski, M.
\newblock Modern analysis of hutchinson's trace estimator.
\newblock In \emph{2021 55th Annual Conference on Information Sciences and
  Systems (CISS)}, pp.\  1--5. IEEE, 2021.

\bibitem[Sohl-Dickstein et~al.(2015)Sohl-Dickstein, Weiss, Maheswaranathan, and
  Ganguli]{sohl2015deep}
Sohl-Dickstein, J., Weiss, E., Maheswaranathan, N., and Ganguli, S.
\newblock Deep unsupervised learning using nonequilibrium thermodynamics.
\newblock In \emph{International Conference on Machine Learning}, pp.\
  2256--2265. PMLR, 2015.

\bibitem[Song \& Ermon(2019)Song and Ermon]{song2019generative}
Song, Y. and Ermon, S.
\newblock Generative modeling by estimating gradients of the data distribution.
\newblock \emph{Advances in Neural Information Processing Systems}, 32, 2019.

\bibitem[Song et~al.(2020{\natexlab{a}})Song, Garg, Shi, and
  Ermon]{song2020sliced}
Song, Y., Garg, S., Shi, J., and Ermon, S.
\newblock Sliced score matching: A scalable approach to density and score
  estimation.
\newblock In \emph{Uncertainty in Artificial Intelligence}, pp.\  574--584.
  PMLR, 2020{\natexlab{a}}.

\bibitem[Song et~al.(2020{\natexlab{b}})Song, Sohl-Dickstein, Kingma, Kumar,
  Ermon, and Poole]{song2020score}
Song, Y., Sohl-Dickstein, J., Kingma, D.~P., Kumar, A., Ermon, S., and Poole,
  B.
\newblock Score-based generative modeling through stochastic differential
  equations.
\newblock \emph{arXiv preprint arXiv:2011.13456}, 2020{\natexlab{b}}.

\bibitem[Song et~al.(2021)Song, Durkan, Murray, and Ermon]{song2021maximum}
Song, Y., Durkan, C., Murray, I., and Ermon, S.
\newblock Maximum likelihood training of score-based diffusion models.
\newblock \emph{Advances in Neural Information Processing Systems},
  34:\penalty0 1415--1428, 2021.

\bibitem[Vincent(2011)]{vincent2011connection}
Vincent, P.
\newblock A connection between score matching and denoising autoencoders.
\newblock \emph{Neural computation}, 23\penalty0 (7):\penalty0 1661--1674,
  2011.

\end{thebibliography}
